%% file: arxiv_v2.tex
\documentclass{article}
\usepackage[final,nonatbib]{neurips_2022}
\usepackage[utf8]{inputenc} 
\usepackage[T1]{fontenc}    
\usepackage{hyperref}       
\usepackage{url}            
\usepackage{booktabs}       
\usepackage{amsfonts}       
\usepackage{nicefrac}       
\usepackage{microtype}      
\usepackage{xcolor}         
\usepackage[numbers]{natbib}%
\input{./mymacro.tex}

\usepackage{enumitem}
\usepackage{xspace}
\usepackage{hyperref,url,dsfont,nicefrac}
\hypersetup{
    colorlinks,
    breaklinks,
    urlcolor  = blue,
    linkcolor = blue,
    citecolor = blue,
}

\def \h {\bm{h}}
\newcommand{\ols}{\mbox{OLaS}\xspace}

\title{Adapting to Online Label Shift with \\Provable Guarantees}

\author{%
  Yong Bai$ ^{1*}$,~ Yu-Jie Zhang$ ^{2,1}$\thanks{Equal contribution.}~~,~ Peng Zhao$ ^1$,~ Masashi Sugiyama$ ^{3,2}$,~ Zhi-Hua Zhou$ ^1$\thanks{Correspondence to: Zhi-Hua Zhou <zhouzh@lamda.nju.edu.cn>} \\
  $^1$ National Key Laboratory for Novel Software Technology, Nanjing University, Nanjing, China\\
  $^2$ The University of Tokyo, Chiba, Japan\\
  $^3$ RIKEN AIP, Tokyo, Japan
}

\begin{document}

\maketitle

\begin{abstract}
The standard supervised learning paradigm works effectively when training data shares the same distribution as the upcoming testing samples. However, this stationary assumption is often violated in real-world applications, especially when testing data appear in an online fashion. In this paper, we formulate and investigate the problem of \emph{online label shift} (\ols): the learner trains an initial model from the labeled offline data and then deploys it to an unlabeled online environment where the underlying label distribution changes over time but the label-conditional density does not. The non-stationarity nature and the lack of supervision make the problem challenging to be tackled. To address the difficulty, we construct a new unbiased risk estimator that utilizes the unlabeled data, which exhibits many benign properties albeit with potential non-convexity. Building upon that, we propose novel online ensemble algorithms to deal with the non-stationarity of the environments. Our approach enjoys optimal \emph{dynamic regret}, indicating that the performance is competitive with a clairvoyant who knows the online environments in hindsight and then chooses the best decision for each round. The obtained dynamic regret bound scales with the intensity and pattern of label distribution shift, hence exhibiting the adaptivity in the \ols problem. Extensive experiments are conducted to validate the effectiveness and support our theoretical findings.
\end{abstract}

\input{sections/introduction}
\input{sections/problem_formulation}
\input{sections/method}
\input{sections/experiments}

\input{sections/conclusion}

\section*{Acknowledgments}
  Yong Bai, Peng Zhao, and Zhi-Hua Zhou were supported by NSFC~(61921006, 62206125), \mbox{Jiangsu}SF (BK20220776), and Collaborative Innovation Center of Novel Software Technology and Industrialization. Yu-Jie Zhang was supported by Todai Fellowship and the Institute for AI and Beyond, UTokyo. Masashi Sugiyama was supported by JST AIP Acceleration Research Grant Number JPMJCR20U3 and the Institute for AI and Beyond, UTokyo. Peng Zhao thanks Chen-Yu Wei for discussions on the conversation between function variation and gradient variation. The authors also thank Ruihan Wu, Yu-Yang Qian, and Dheeraj Baby for their helpful discussions.

\bibliography{onlineTS,onlineBy}
\bibliographystyle{unsrtnat}

\newpage
\appendix
\input{sections/appendices/catlog}
\input{sections/appendices/experiments}
\input{sections/appendices/preliminary}

\input{sections/appendices/related_work}

\input{sections/appendices/comparison}
\input{sections/appendices/proofs}

\input{sections/appendices/technique_lemma}

\end{document}

%% file: mymacro.tex
\usepackage{amsmath}
\usepackage{amssymb}
\usepackage{mathtools}
\usepackage{amsthm}
\usepackage{dsfont}
\usepackage{lipsum}
\usepackage{bm,bbm}
\usepackage{bbding}

\usepackage{paralist}
\usepackage{multirow}
\usepackage{graphicx}
\usepackage{subfigure}
\usepackage[font=small,labelfont=bf]{caption}
\usepackage{subfloat}
\usepackage{float}
\usepackage{algorithm,algorithmic}
\usepackage{wrapfig}
\usepackage{booktabs}
\usepackage{verbatim}
\usepackage[shortlabels]{enumitem}
\usepackage{setspace}
\usepackage{tcolorbox}

\def \E {\mathbb{E}}
\def \x {\mathbf{x}}

\def \D {\mathcal{D}}

\def \u {\mathbf{u}}
\def \H {\mathcal{H}}
\def \w {\mathbf{w}}
\def \O {\mathcal{O}}
\def \R {\mathbb{R}}
\def \indicator {\mathbbm{1}}

\def \SReg {\mathop{\bm{\mathrm{Reg}}}\nolimits_{T}^{\bm{\mathsf{s}}}}
\def \DReg {\mathop{\bm{\mathrm{Reg}}}\nolimits_{T}^{\bm{\mathsf{d}}}}

\def \Y {\mathcal{Y}}

\def \G {G}

\def \W {\mathcal{W}}

\def \p {\mathbf{p}}

\def \h {\mathbf{h}}

\def \Rh {\hat{R}}

\def \V {\mathcal{V}}

\def \m {\mathbf{m}}

\def \V {V}

\def \wh {\widehat{\w}}

\def \X {\mathcal{X}}

\DeclareMathOperator*{\argmax}{arg\,max}
\DeclareMathOperator*{\argmin}{arg\,min}
\renewcommand{\tilde}{\widetilde}
\renewcommand{\hat}{\widehat}

\newcommand\given[1][]{\:#1\vert\:}
\newcommand\givenn[1][]{\:#1\middle\vert\:}

\usepackage{mathtools}
\usepackage{appendix}
\let\norm\undefined 
\DeclarePairedDelimiter\norm{\lVert}{\rVert}

\newcommand\inner[2]{\langle #1, #2 \rangle}
\newtheorem{myThm}{Theorem}

\newtheorem{myLemma}{Lemma}

\theoremstyle{definition}
\newtheorem{myAssum}{Assumption}

\newtheorem{myRemark}{Remark}

\renewcommand{\tilde}{\widetilde}
\renewcommand{\hat}{\widehat}

\def \E {\mathbb{E}}
\def \D {\mathcal{D}}

\def \H {\mathcal{H}}

\def \R {\mathbb{R}}

\def \W {\mathcal{W}}
\def \X {\mathcal{X}}
\def \Y {\mathcal{Y}}

\def \x {\mathbf{x}}

\def \Cf {C_{f_0}}
\def \Chf {\hat{C}_{f_0}}
\def \Jf {J_{f_0}}
\def \Jhf {\hat{J}_{f_0}}
\def \Wf {W_{f_0}}
\def \Whf {\hat{W}_{f_0}}
\def \Rhb {\hat{\bm{R}}}
\def \epsilon {\varepsilon}

\definecolor{DSgray}{cmyk}{0,1,0,0}

\usepackage{prettyref}
\newcommand{\pref}[1]{\prettyref{#1}}

\newcommand{\savehyperref}[2]{\texorpdfstring{\hyperref[#1]{#2}}{#2}}
\newrefformat{eq}{\savehyperref{#1}{Eq.~(\ref*{#1})}}
\newrefformat{eqn}{\savehyperref{#1}{Equation~\ref*{#1}}}
\newrefformat{lem}{\savehyperref{#1}{Lemma~\ref*{#1}}}
\newrefformat{lemma}{\savehyperref{#1}{Lemma~\ref*{#1}}}
\newrefformat{def}{\savehyperref{#1}{Definition~\ref*{#1}}}
\newrefformat{line}{\savehyperref{#1}{Line~\ref*{#1}}}
\newrefformat{thm}{\savehyperref{#1}{Theorem~\ref*{#1}}}
\newrefformat{corr}{\savehyperref{#1}{Corollary~\ref*{#1}}}
\newrefformat{cor}{\savehyperref{#1}{Corollary~\ref*{#1}}}
\newrefformat{sec}{\savehyperref{#1}{Section~\ref*{#1}}}
\newrefformat{app}{\savehyperref{#1}{Appendix~\ref*{#1}}}
\newrefformat{appendix}{\savehyperref{#1}{Appendix~\ref*{#1}}}
\newrefformat{assum}{\savehyperref{#1}{Assumption~\ref*{#1}}}
\newrefformat{ex}{\savehyperref{#1}{Example~\ref*{#1}}}
\newrefformat{fig}{\savehyperref{#1}{Figure~\ref*{#1}}}
\newrefformat{alg}{\savehyperref{#1}{Algorithm~\ref*{#1}}}
\newrefformat{rem}{\savehyperref{#1}{Remark~\ref*{#1}}}
\newrefformat{conj}{\savehyperref{#1}{Conjecture~\ref*{#1}}}
\newrefformat{prop}{\savehyperref{#1}{Proposition~\ref*{#1}}}
\newrefformat{proto}{\savehyperref{#1}{Protocol~\ref*{#1}}}
\newrefformat{prob}{\savehyperref{#1}{Problem~\ref*{#1}}}
\newrefformat{claim}{\savehyperref{#1}{Claim~\ref*{#1}}}
\newrefformat{que}{\savehyperref{#1}{Question~\ref*{#1}}}
\newrefformat{op}{\savehyperref{#1}{Open Problem~\ref*{#1}}}
\newrefformat{fn}{\savehyperref{#1}{Footnote~\ref*{#1}}}
\allowdisplaybreaks[3]

%% file: sections/introduction.tex

\section{Introduction}
\label{sec:introduction}
One of the fundamental challenges for modern machine learning is the distribution shift~\citep{edit:Quinonero-Candela+etal:2009, book/mit/sugiyama2012machine,TGD:robust-AI,CACM'21:DL-AI}. The learned model's testing performance would significantly drop when the distribution is different from the initial training distribution. More severely, in many real-world applications, testing data often come in an \emph{online} fashion after deploying the trained model such that the underlying distribution might continuously change over time. Hence, it is necessary to develop learning methods to handle distribution shift in online and open environments~\citep{nsr'22:Open-Survey}. Another practical concern is the \emph{label scarcity} issue in real tasks, particularly those tasks emerging in online scenarios. For example, in the species monitoring task~\citep{ECOI'17:wild, PNAS'18:Species}, a learned model is deployed to detect species of wild animals. The data consist of received signals from sensors and hence are naturally in the streaming form. The data distribution of upcoming animals will change due to the variety of species across different geographic locations and seasons, and moreover, it is hard to gather the labels of streaming data in time.

Motivated by the above real demand, this paper is concerned with the following problem: how to design an algorithm that can adapt to non-stationary environments with a few labeled data or even unlabeled data observed at every time? In addition to getting empirical performance gain, the overall method is desired to have clear and strong theoretical guarantees. The problem is generally hard due to the non-stationarity of online environments and the lack of supervision. As such, we investigate a simplified case with a focus on the specific change on the \emph{label distribution}. We formalize the \emph{online label shift} (OLaS) problem,\footnote{We use \ols instead of OLS, since OLS often refers to ``ordinary least squares''.} which consists of two stages, including offline initialization and online adaptation. 
Specifically, the learner collects labeled samples drawn independently from the initial distribution $\D_0(\x,y)$ where $\x$ and $y$ denote the feature vector and its associated label, and trains an initial model following standard supervised learning methods. Subsequently, she needs to adapt it to an unknown non-stationary environment where the underlying label distributions change over time. Specifically, at time $t$, she receives a few \emph{unlabeled} samples drawn from the current distribution $\D_t(\x)$ and uses them to update the model. In \ols problems, the essential environment change happens on the label distribution $\D_t(y)$ with the conditional $\D(\x \given y)$ always remaining the same.

The label shift problem has been widely studied in the offline setting~\citep{DBLP:journals/neco/SaerensLD02, conf/icml/ZhangSMW13, DBLP:journals/nn/PlessisS14, DBLP:conf/acml/NguyenPS15,conf/icml/LiptonWS18, conf/iclr/Azizzadenesheli19, DBLP:conf/nips/GargWBL20}, but this is less explored in the more challenging online setup. One natural impulse is to handle \ols by online learning techniques~\citep{book'16:Hazan-OCO}, but it is generally non-trivial due to the lack of supervision in the adaptation stage and also the non-stationarity issue.~\citet{NIPS'21:Online-LS} made the first such attempt for \ols, where they constructed an unbiased risk estimator with the unlabeled data for model assessment and used online gradient descent for model updating. Let $T$ denote the number of rounds. They proved an $\O(\sqrt{T})$ regret bound, which measures the gap between the learner's decision and the best \emph{fixed} decision in hindsight. However, in non-stationary environments, a single decision can hardly perform well all the time, which makes the guarantee less attractive for \ols problems. Another technical caveat is that their theory relies on a vital assumption of the convexity of risk functions, which was not verified strictly. In fact, this assumption can hardly be satisfied as an operation to take the argument of the maximum is involved in its formulation of the risk estimator. 

In this paper, we aim to develop an algorithm for adapting to online label shift with \emph{provable} guarantees. To this end, we first reframe the construction of the unbiased risk estimator via risk rewriting techniques and prove that the estimator still enjoys benign theoretical properties, albeit with a potential \emph{non-convex} behavior. Second, to handle the non-stationarity of the online stream, instead of using traditional regret as the performance measure, we employ \emph{dynamic regret} to guide the algorithm design, which ensures the online algorithm is competitive with a clairvoyant who knows the online functions in hindsight and hence chooses the best decision of each round. To optimize such a strengthened measure, we propose a novel online ensemble algorithm building upon the risk estimator, consisting of a meta-algorithm running over a group of base-learners, each associated with a customized configuration. Our algorithm enjoys an $\O(V_T^{1/3}T^{2/3})$ dynamic regret, where $V_T = \sum_{t=2}^T \norm{\bm{\mu}_{y_t} - \bm{\mu}_{y_{t-1}}}_1$ measures the variation of label distributions, with $\bm{\mu}_{y_t}$ denoting the vector consisting of the class-prior probabilities at time $t$. Notably, the regret guarantee achieved by our algorithm is \emph{minimax optimal} in terms of both number of rounds and non-stationarity measure, and importantly, our algorithm does not require the unknown class-prior variation $V_T$ as the input. 

Furthermore, for many situations where online label shift contains some patterns such as periodicity or gradual change behavior, we present an improved algorithm to exploit such structures and achieve provably better guarantees. The key idea is to leverage historical information to serve as a hint for online updates.
We prove an $\O(V_T^{1/3}G_T^{1/3}T^{1/3})$ dynamic regret, where $G_T$ measures the reusability of historical information that is at most $\O(T)$ while could be much smaller in benign environments. As a benefit, the improved algorithm safeguards the $\O(V_T^{1/3} T^{2/3})$ worst-case bound and meanwhile achieves great improvement in easier environments. 
Extensive experiments are conducted to evaluate our approach, which show the usefulness of meta-base structure in tackling non-stationarity and validate the effectiveness of other adaptive components. 

\vspace{1mm}
\textbf{Technical Contribution.~~} Our method is not a direct application of existing non-stationary online learning methods~\citep{OR'15:dynamic-function-VT,NIPS'18:Zhang-Ader,NIPS'20:sword}, but rather requires in-depth technical innovations. First, the potentially \emph{non-convex} risk estimator makes it hard to apply existing techniques of online convex optimization (OCO)~\citep{book'16:Hazan-OCO}, but fortunately, we prove the convexity of its expectation such that the OCO framework can still be used (see Remark~\ref{remark:non-convexity}). Second, to optimize the dynamic regret, we employ a meta-base structure to hedge the uncertainty of the \emph{unknown} minimizer of the expected risk function at each round and convert the variation of expected risk minimizers to the intensity of label distribution drifts, a natural non-stationarity measure for the \ols problem (see Remark~\ref{remark:ns-measure-OLS}). Third, earlier study showed adaptive dynamic regret bounds for convex and smooth functions~\citep{NIPS'20:sword}, while the smoothness assumption can hardly be satisfied in our case. We remove such a constraint by introducing an \emph{implicit update}, which could be of independent interest for general OCO purposes (see Remark~\ref{remark:implicit-update}).

%% file: sections/problem_formulation.tex
\section{Problem Formulation}
\label{sec:problem-formulation}

 We focus on \ols of multi-class classification with feature space $\X \subseteq \R^d$ and label space $\Y = [K] \triangleq\{1,\ldots,K\}$, where $d$ is the dimension and $K$ is the number of classes. Below we formulate setups of two stages of \ols (offline initialization and online adaptation).

 \vspace{1mm}
\textbf{Problem Setup.~~} In the \emph{offline initialization stage}, the learner collects a number of labeled samples denoted by $S_0 = \{(\x_n,y_n)\}_{n=1}^{N_0}$ drawn from the distribution $\D_0(\x,y)$ and then obtains a well-performed initial model $f_0:\X\mapsto\Y$. 
In the \emph{online adaptation stage}, data come in the streaming form \emph{without} labels. The current model is deployed to predict the labels of online data and also to evolve adaptively. Specifically, at each round $t \in [T]$, the learner receives a small number of \emph{unlabeled} data $S_t = \{\x_n\}_{n=1}^{N_t}$ drawn from the distribution $\D_t(\x)$. The non-stationary nature indicates $\D_t \neq \D_{t'}$ in general for different $t,t' \in [T]$. \ols considers the simplified case that essential changes come from label distributions and there are no new classes, formally described in the following condition.
\begin{myAssum}[Online Label Shift]
\label{assum:label-shift}
In the online label shift problem, the label distribution $\D_t(y)$ changes over time while the class-conditional distribution $\D_t(\x \given y)$ is identical throughout the process for $t\in\{0,1,\ldots,T\}$. Moreover, it holds that $\D_0(y)> 0$ for any $y\in\Y$.
\end{myAssum}

\vspace{1mm}
\textbf{Performance Measure.~~} At round $t \in [T]$, the learner uses the information observed so far to make the prediction and also update the model $\w_t\in\W$, where $\W$ is a convex decision set with diameter $\Gamma \triangleq \sup_{\w,\w' \in \W} \norm{\w - \w'}_2$. The goal is to ensure that the $t$-round model $\w_t$ generalizes well on the underlying distribution $\D_t$. Thus, the model's quality is evaluated by its risk defined as $R_t(\w) = \E_{(\x,y)\sim \D_t}[\ell(f(\w,\x),y)]$,
where $f:\W\times\X\mapsto\R^K$ is the predictive function  and $\ell:\R^K\times\Y\mapsto\R$ is any convex surrogate loss for classification such that $\ell(f(\w,\x),y)$ is convex in $\w$.
We introduce two constants, $G \triangleq \sup_{(\x,y)\in\X\times\Y,\w\in\W} \norm{\nabla_\w \ell(f(\w,\x),y)}_2$ and $B \triangleq \sup_{(\x,y)\in\X\times\Y,\w\in\W} \vert \ell(f(\w,\x),y)\vert$, as upper bounds of gradient norm and loss function value. 

We use regret to examine the performance of online algorithms. In particular, \emph{dynamic regret}~\citep{ICML'03:zinkvich,OR'15:dynamic-function-VT} is employed to compete the algorithm's performance with the best response at each round, defined as  
\begin{equation}
	\label{eq:dynamic-regret-def}
	\DReg \triangleq \sum_{t=1}^T R_t(\w_t) - \sum_{t=1}^T \min_{\w \in \W} R_t(\w) = \sum_{t=1}^T R_t(\w_t) - \sum_{t=1}^T R_t(\w_t^*),
\end{equation}
where $\w_t^* \in \argmin_{\w\in\W} R_t(\w)$ is the model (or one of the models) with the best generalization ability on the distribution $\D_t$. Notably, it is known that a sublinear dynamic regret is impossible in the worst case~\citep{OR'15:dynamic-function-VT}, so an upper bound of dynamic regret is desired to scale with a certain non-stationarity measure. A natural measure for \ols would be the variation intensity of label distributions.

\begin{myRemark}[Static regret vs.~dynamic regret]
\label{remark:dynamic-regret}
The classic measure for online learning is \emph{static} regret, defined as $\SReg = \sum_{t=1}^T R_t(\w_t)- \sum_{t=1}^T R_t(\w^*)$, where $\w^* \in \argmin_{\w\in\W}\sum_{t=1}^T R_t(\w)$ is the best \emph{fixed} model in hindsight. The measure was adopted in the prior work of \ols~\citep{NIPS'21:Online-LS}. However, the measure is not suitable for \ols, because it is too optimistic to expect a single fixed model to behave well over the whole process in changing environments. By contrast, minimizing dynamic regret facilitates the online algorithm with more adaptivity and robustness to non-stationary environments.
\end{myRemark}

%% file: sections/method.tex

\section{Proposed Approach}
\label{sec:approach}
This section presents our approach for the \ols problem, including the algorithms and theoretical guarantees. In the following, we respectively address the two central challenges of \ols: the lack of supervision and the non-stationarity of online environments.

\subsection{Unbiased Risk Estimator for Online Convex Optimization}
\label{sec:unbiased-risk-estimator}
OCO is a powerful and versatile framework for online learning problems, which enjoys both practical and theoretical appeals~\citep{book/Cambridge/cesa2006prediction,book'16:Hazan-OCO}. Online Gradient Descent (OGD)~\citep{ICML'03:zinkvich} is one of the most fundamental and powerful algorithms due to its light computational cost and sound regret guarantees. In the \ols problem, recall that the learner's goal is to obtain a model sequence  $\{\w_t\}_{t=1}^T$ enjoying low cumulative risk $\sum_{t=1}^T R_t(\w_t)$. Thus, suppose the model's risk $R_t(\w_t)$ is known at each round; then OGD simply updates the model by $\w_{t+1} = \Pi_{\W}\left[\w_t - \eta \nabla R_t(\w_t)\right]$, where $\Pi_{\W}[\cdot]$ denotes the projection onto $\W$ and $\eta>0$ is the step size. It is well-known that OGD guarantees the regret bound $\sum_{t=1}^T R_t(\w_t) - \min_{\w \in \W}\sum_{t=1}^T R_t(\w)\leq \O(\sqrt{T})$ when risk function $R_t(\w)$ is convex and step size is set as $\eta = \Theta(T^{-1/2})$~\citep{book'16:Hazan-OCO} (see Appendix~\ref{sec-appendix:preliminary-oco} for more details).

However, the expected risk function $R_t(\w)$ is unknown in the current \ols setup as it is defined over the underlying joint distribution $\D_t(\x, y)$. More severely, the online environments in the \ols problem are fully \emph{unlabeled}, which poses great challenges to apply the OCO framework. Indeed, the lack of supervision makes it hard to empirically assess the expected risk, not to mention ensuring the convexity. In the following, we establish an unbiased estimator $\hat{R}_t(\w)$ with unlabeled data $S_t$, which exhibits nice properties such that the OCO framework is still applicable for our purpose.

\vspace{1mm}
\textbf{Unbiased Estimator under Label Shift.~~} Inspired by the progress in offline label shift~\citep{conf/icml/ZhangSMW13,conf/icml/LiptonWS18,conf/iclr/Azizzadenesheli19}, we establish an unbiased risk estimator in \ols for $R_t(\w)$ with unlabeled data $S_t$ and offline data $S_0$ by the risk rewriting technique. To this end, let $\bm{\mu}_{y_t}\in\Delta_{K}$ denote the label distribution vector with the $k$-th entry $[\bm{\mu}_{y_t}]_k \triangleq \D_t(y = k)$, then we have the following decomposition for the true risk:
\begin{align}
\label{eq:risk-decomposition}
  R_t(\w) \triangleq \E_{(\x,y)\sim\D_t} [\ell(f(\w,\x),y)]
  =\sum_{k=1}^K [\bm{\mu}_{y_t}]_k\cdot R_t^k(\w)
  =\sum_{k=1}^K [\bm{\mu}_{y_t}]_k\cdot R_0^k(\w),
\end{align}
where $R_t^k(\w) \triangleq \E_{\x\sim\D_t\left(\x \givenn y=k\right)}[\ell(f(\w,\x),k)]$ is the risk of the model over the $k$-th label at round $t$. The second equality holds due to the law of total probability, and the third equality is by the label shift assumption that $\D_t(\x \given y) = \D_0(\x \given y)$ for any $t\in[T]$. Since the labeled offline data $S_0$ is always available, one can approximate $R_0^k(\w)$ by its empirical version $\hat{R}_0^k(\w)$ with  offline data $S_0$, where $\hat{R}_0^k(\w) \triangleq \frac{1}{\vert S_0^k\vert}\sum_{\x_n\in S_0^k} \ell(f(\w,\x_n),k)$ and $S_0^k$ is a subset of $S_0$ containing all samples with label $k$. 

Therefore, the task is now to estimate the label distribution vector $\bm{\mu}_{y_t}$. To this end, we employ the Black Box Shift Estimation (BBSE) method~\citep{conf/icml/LiptonWS18} to construct an estimator via only offline data $S_0$ and unlabeled data $S_t$. Specifically, we first use the initial offline model $f_0$ to predict over the unlabeled data $S_t$ and get predictive labels $\hat{y}_t$, and next estimate the label distribution via solving the crucial equation $\bm{\mu}_{y_t} = C_{f_0}^{-1}\bm{\mu}_{\hat{y}_t}$, where $\bm{\mu}_{\hat{y}_t}\in\Delta_{K}$ is the distribution vector of the predictive labels $\hat{y}_t$ and $C_{f_0}\in \R^{K\times K}$ is the confusion matrix with $[C_{f_0}]_{ij} = \E_{\x\sim\D_0\left(\x \given y=j\right)}[\indicator\{f_0(\x) = i\}]$ being the classification rate that the initial model $f_0$ predicts samples from class $i$ as class $j$. We defer more details to Appendix~\ref{sec-appendix:preliminary-bbse}. As a result, through risk rewriting and prior estimation, we can construct the following estimator for the true risk $R_t(\w)$:
\begin{align}
\label{eq:unbiased-risk-estimator}
\hat{R}_t(\w) 
= \sum_{k=1}^K [\hat{C}_{f_0}^{-1}\hat{\bm{\mu}}_{\hat{y}_t}]_k\cdot \hat{R}_0^k(\w), 
\end{align}
where $\hat{C}_{f_0}$ and $\hat{\bm{\mu}}_{\hat{y}_t}$ are empirical estimators of the confusion matrix and predictive label distribution vector using offline data $S_0$ and unlabeled data $S_t$ only. Our constructed risk estimator enjoys the unbiasedness property, which plays a crucial role in the later algorithm design and theoretical analysis. 
\begin{myLemma}
\label{lemma:unbiased-estimator}
The estimator $\hat{R}_t(\w)$ in~\pref{eq:unbiased-risk-estimator} is unbiased to $R_t(\w) = \E_{(\x,y)\sim \D_t}[\ell(f(\w,\x),y)]$, i.e., $\E_{S_t\sim\D_t}[\hat{R}_t(\w)] = R_t(\w)$, for any $\w\in\W$ independent of the dataset $S_t$, provided $C_{f_0}$ is invertible and the offline dataset $S_0$ has sufficient samples such that $\hat{C}_{f_0} = C_{f_0}$ and $\hat{R}_0^k(\w) = R_0^k(\w)$, $\forall k\in\Y$. 
\end{myLemma}

The proof of Lemma~\ref{lemma:unbiased-estimator} is in Appendix~\ref{sec-appendix:proof-lemma1}. Note that the sufficient sample assumption is introduced on offline data $S_0$ to simplify the presentation. Indeed, we can further show $\vert\E_{S_t\sim \D_t}[\hat{R}_t(\w)] - R_t(\w)\vert\leq \O(\sqrt{1/\vert S_0\vert})$ with high probability (details in Appendix~\ref{sec-appendix:risk-estimator-hp}). Such an additional dependence on $S_0$ also appears in the classical offline label shift~\citep{conf/icml/LiptonWS18,conf/iclr/Azizzadenesheli19} and is negligible when a large number of offline data is collected at the initial stage. The requirement is easy to realize and will not trivialize the online adaptation. Another caveat is that storing all the offline data can be burdensome in resource-constrained learning scenarios, then one may use data sketching techniques like corsets~\citep{arXiv'17:coreset,ICML'20:coreset} or reduced kernel mean embedding~\citep{TKDE'21:RKME,AAAI'21:UnseenJob,FTML'17:KME} to further improve the storage complexity.

\begin{myRemark}[Non-convexity issue]
\label{remark:non-convexity}
Our risk estimator $\hat{R}_t(\w)$ can be \emph{non-convex} as the estimated label distribution $[\hat{C}_{f_0}^{-1}\hat{\bm{\mu}}_{\hat{y}_t}]_k$ might be negative in order to ensure the unbiasedness. Such a non-convex behavior introduces a great challenge for applying the OCO framework. Fortunately, owing to its unbiasedness and the fact that the \emph{expected} risk $R_t(\w)$ is indeed convex, we can continue the following algorithm design and theoretical analysis building upon the constructed unbiased estimator.
\end{myRemark}

\vspace{1mm}
\textbf{OGD with Unbiased Estimator.~~} Building upon the risk estimator in~\pref{eq:unbiased-risk-estimator}, we then deploy OGD  and obtain our \textsc{UOGD} algorithm (abbreviated for ``OGD with unbiased risk estimator''), namely,
\begin{equation}
    \label{eq:UOGD}
    \w_{t+1} = \Pi_{\W}[\w_t - \eta \nabla \hat{R}_t(\w_t)].
\end{equation}
Despite the potential non-convexity of the risk estimator itself, we can still establish solid regret guarantees via the OCO framework due to the benign property that the risk estimator is unbiased and the expected risk is indeed convex. For example, UOGD provably enjoys an $\O(\sqrt{T})$ static regret.  
See the formal statement in Appendix~\ref{sec-appendix:static-regret}. We remark that our attained static regret already achieves the state-of-the-art theoretical understanding of \ols, in the sense that previously the same bound can be only achieved with an additional unrealistic convexity assumption imposed over the algorithm~\citep{NIPS'21:Online-LS}, which UOGD does not require.
Concretely,~\citet{NIPS'21:Online-LS} assumed that \emph{the risk estimator is convex} (in expectation), which is hard to theoretically verify since the estimator approximates the $0/1$-loss and involves an indicator function and an argmax operation due to the employed reweighting mechanism. By contrast, our estimator directly approximates the surrogate loss without reweighting and thus does not suffer from such limitations. Even if modifying their estimator to optimize a surrogate loss, the reweighting mechanism makes their method still hardly suitable for the OCO framework. More details are in Appendix~\ref{sec-appendix:difference}. In a nutshell, our constructed risk estimator enjoys nice properties, which are indispensable for the algorithm design and theoretical analysis. 

\subsection{Adapting to Non-stationarity of Online Label Shift}
\label{sec:online-algorithm}
So far, an $\O(\sqrt{T})$ static regret has been established for \ols; however, the guarantee is not appealing because static regret is not suitable for non-stationary online problems as discussed in Remark~\ref{remark:dynamic-regret}. We now introduce our method adapting to the non-stationarity with provable dynamic regret guarantees.

First, benefiting from the unbiasedness and expected convexity of our risk estimator, we prove that UOGD achieves a dynamic regret scaling with the label distribution drift $V_T$.
\begin{myThm}
\label{thm:UOGD}
Under the same assumptions as Lemma~\ref{lemma:unbiased-estimator}, UOGD in~\pref{eq:UOGD} with step size $\eta$ satisfies
\begin{equation}
\label{eq:dynamic-UOGD}
\E\big[\DReg\big] \leq 2(\nicefrac{KG^2}{\sigma^2}+B^2)\eta T + \nicefrac{\Gamma^2}{\eta} + 4(\Gamma+1)\sqrt{\nicefrac{B V_T T}{\eta}} = \O\Big( \eta T + 1/\eta + \sqrt{(V_T T)/\eta} \Big),
{}\end{equation}
where the constant $\sigma > 0 $ denotes the minimum singular value of the invertible confusion matrix $C_{f_0}$. Moreover, $V_T = \sum_{t=2}^T \norm{\bm{\mu}_{y_t}-\bm{\mu}_{y_{t-1}}}_1$ measures the intensity of the label distribution shift. 
\end{myThm}

The proof of Theorem~\ref{thm:UOGD} can be found in Appendix~\ref{appendix:sec-proof-UOGD}. The dynamic regret guarantee is obtained in a non-trivial way, and below we expand the technical innovations. 
\begin{myRemark}[Non-stationarity measure for \ols]
  \label{remark:ns-measure-OLS}
For readers who are familiar with the literature, our result is reminiscent of the existing dynamic regret bound in the  OCO studies~\citep{OR'15:dynamic-function-VT,AISTATS'15:dynamic-optimistic} on the surface; however, our result exhibits fundamental differences. The key caveat is that in our case the comparator $\w_t^*$ at each round is \emph{not} the minimizer of the online function. Specifically, as the expected risk $R_t$ is inaccessible, one has to work on the unbiased risk estimator $\Rh_t$ and requires optimizing the \emph{empirical} dynamic regret $\sum_{t=1}^T \Rh_t(\w_t) - \sum_{t=1}^T \Rh_t(\w_t^*)$. Importantly, $\w_t^* \in \argmin_{\w \in \W} R_t(\w)$ but $\w_t^* \notin \argmin_{\w \in \W} \Rh_t(\w)$ in general. Although the empirical dynamic regret can be trivially bounded by $\sum_{t=1}^T \Rh_t(\w_t) - \sum_{t=1}^T \min_{\w \in \W} \Rh_t(\w)$, the bound will then be loose and related to temporal variation of risk estimators~\citep{OR'15:dynamic-function-VT,AISTATS'15:dynamic-optimistic}, making it hard to establish relationship to the natural non-stationarity measure of \ols: $V_T$.\footnote{\citet{OR'15:dynamic-function-VT} also considered a  more general setting with noisy function value feedback. In such cases, the comparator is not the exact minimizer of the online function at each round, and their algorithm will require a \emph{periodical restart} to deal with the non-stationarity. By contrast, ours does not require the restart in the algorithm.} 
On the other hand, there exist studies benchmarking dynamic regret with other choices of comparators~\citep{ICML'03:zinkvich,NIPS'18:Zhang-Ader}, but the bounds scale with the consecutive variation of comparators $\sum_{t=2}^T \norm{\w_t^* - \w_{t-1}^*}_2$, whose relation to $V_T$ is also unclear. To address the difficulty, drawing inspiration from~\citep{UAI'20:Simple}, we decompose the expected dynamic regret into two parts: (i) the analysis of the first part relies on an in-depth analysis of the empirical dynamic regret of UOGD to track a specific sequence of piecewise-stationary comparators to avoid undesired variation of comparators; and (ii) the analysis of the second part is directly conducted on the \emph{original} risk functions to attain a non-stationarity measure only related to the underlying label distribution shift.
\end{myRemark}

\begin{figure}[!t]
\begin{minipage}{0.49\textwidth}
\begin{algorithm}[H]
   \caption{\textsc{Atlas}: base-algorithm}
   \label{alg:atlas-base-hedge}
\begin{algorithmic}[1]
  \REQUIRE{step size $\eta_i \in \H$}
  \STATE{let $\w_{1,i}$ be any point in $\W$}
    \FOR{$t=2$ {\bfseries to} $T$}
      \STATE construct the risk estimator $\hat{R}_{t-1}$ as~\eqref{eq:unbiased-risk-estimator}
      \STATE update the model of base-learner by \\$\w_{t,i} = \Pi_{\W}[\w_{t-1,i} - \eta_i \nabla \Rh_{t-1}(\w_{t-1,i})]$      
      \label{alg:atlas-base-hedge-update}
      \STATE send $\w_{t,i}$ to the meta-algorithm
    \ENDFOR
\end{algorithmic}
\end{algorithm}
\end{minipage}
\hfill
\begin{minipage}{0.49\textwidth}
\begin{algorithm}[H]
   \caption{\textsc{Atlas}: meta-algorithm}
   \label{alg:atlas-meta-hedge}
   \begin{spacing}{0.957}
    \begin{algorithmic}[1]
      \REQUIRE{step size pool $\H$; learning rate $\varepsilon$}
      \STATE{initialization: $\forall i\in [N], p_{1,i} = 1/N$}
        \FOR{$t=2$ {\bfseries to} $T$}
          \STATE  receive $\{\w_{t,i}\}_{i=1}^N$ from base-learners
          \STATE update weight $\p_t \in \Delta_N$ according to $p_{t,i} \propto \exp(-\varepsilon\sum_{s=1}^{t-1}\Rh_s(\w_{s,i}))$, $i \in[N]$ \label{alg:atlas-meta-hedge-update}
          \STATE predict final output $\w_{t} = \sum_{i=1}^{N} p_{t,i} \w_{t,i}$
        \ENDFOR
    \end{algorithmic}
   \end{spacing}
\end{algorithm}
\end{minipage}
\vspace{-5mm}
\end{figure}

From the upper bound in~\pref{eq:dynamic-UOGD}, we can observe that a proper step size tuning is crucial. Specifically, it can be verified that when the environment is near-stationary (more precisely, $V_T \leq \Theta(T^{-\frac{1}{2}})$), simply choosing $\eta = \Theta(1/\sqrt{T})$ ensures an $\O(\sqrt{T})$ dynamic regret, which is known to be minimax optimal even for the weaker measure of static regret~\citep{COLT'08:lower-bound}. Thus, in the following, we focus on the non-degenerated \ols situation where $V_T \geq \Theta(T^{-\frac{1}{2}})$, and then the dynamic regret upper bound can be further simplified as $\O ( \eta T + 1/\eta + \sqrt{(V_T T)/\eta} )$. As a result, UOGD can attain an $\O (V_T^{\frac{1}{3}} T^{\frac{2}{3}})$ dynamic regret by setting the step size optimally as $\eta =  \Theta(T^{-\frac{1}{3}}V_T^{\frac{1}{3}})$. According to the lower bound results of~\citet{OR'15:dynamic-function-VT}, we know that UOGD with an optimal step size tuning ensures a minimax optimal dynamic regret guarantee, see more discussions on the minimax optimality in Appendix~\ref{appendix:discuss-optimal}.

However, the optimal step size $\eta^* = \Theta(T^{-\frac{1}{3}}V_T^{\frac{1}{3}})$ crucially depends on the label distribution drift $V_T = \sum_{t=2}^T \norm{\bm{\mu}_{y_t}-\bm{\mu}_{y_{t-1}}}_1$, which measures the non-stationarity and is unfortunately \emph{unknown} to the learner. It is worth emphasizing that the problem cannot be addressed by standard adaptive step size tuning mechanisms in online learning literature, such as the doubling trick~\citep{JACM'97:doubling-trick} or self-confident tuning~\citep{JCSS'02:Auer-self-confident}, in that the variation quantity $V_T$ cannot be empirically evaluated as it is defined over the underlying label distribution $\D_t(y)$ inaccessible to the learner. Even diving into the analysis of Theorem~\ref{thm:UOGD}, the adaptive tuning requires the knowledge of \emph{original} risk functions $\{R_t\}_{t=1}^T$, which are also unavailable. Intuitively, the hardness of such optimal step size tuning essentially comes from the uncertainty of the non-stationary label shifts. 

To overcome the difficulty, inspired by recent advances in non-stationary online learning~\citep{NIPS'18:Zhang-Ader,NIPS'20:sword}, we propose an \emph{online ensemble} algorithm for the \ols problem called \textsc{Atlas} (\underline{A}dapting \underline{T}o \underline{LA}bel \underline{S}hift). Specifically, to cope with the uncertainty in the optimal step size tuning, we first design a pool of candidate step sizes denoted by $\H = \{\eta_1,\ldots,\eta_N\}$ such that $\eta_*$ can be well approximated by at least one of those candidates, which will be configured later and here $N$ is the number of candidate step sizes. Then, \textsc{Atlas} deploys a two-layer structure by maintaining a group of base-learners, each associated with a candidate step size from the pool $\H$ and then employs a meta-algorithm to track the best base-learner. The main procedures are presented in Algorithm~\ref{alg:atlas-base-hedge} (base-algorithm) and Algorithm~\ref{alg:atlas-meta-hedge} (meta-algorithm), and we describe the details below.

\vspace{1mm}
\textbf{Base-algorithm.~~} We parallelly run a group of instances of UOGD, each one is associated with a candidate step size in the step size pool $\H$. Formally, the $i$-th base-learner yields a sequence of base models $\{\w_{t,i}\}_{t=1}^T$, which is updated by $\w_{t,i} = \Pi_{\W}[\w_{t-1,i} - \eta_i \nabla \Rh_{t-1}(\w_{t-1,i})]$ with $\eta_i \in \H$.

\vspace{1mm}
\textbf{Meta-algorithm.~~} The meta-algorithm aims to combine all the base-learners' decisions such that the final output is competitive with the decisions returned by the (unknown) base-learner associated with the best step size. To achieve this, we employ a weighted combination mechanism with the final model as $\w_t = \sum_{i=1}^N p_{t,i} \w_{t,i}$, where $\mathbf{p}_t \in\Delta_N$ is the weight vector with $p_{t,i}$ denoting the weight of the $i$-th base-learner. We use the classic Hedge algorithm~\citep{JCSS'97:boosting} to update the weight vector, namely, $p_{t,i} \propto \exp\big(-\varepsilon\sum_{s=1}^{t-1}\Rh_s(\w_{s,i})\big)$, where $\varepsilon > 0$ is the learning rate of the meta-algorithm that can be simply set as $\Theta(\sqrt{(\ln N)/T})$ without dependence on label shift quantity. Intuitively, the meta-algorithm puts larger weights on base-learners with a smaller cumulative estimated risk so that the overall models $\{\w_t\}_{t=1}^T$ can be competitive to the base-learner with the best performance. 
 
\textsc{Atlas} enjoys the following dynamic regret guarantee with the proof in Appendix~\ref{appendix:sec-proof-overall-UOGD}.
\begin{myThm}
\label{thm:overall-UOGD} 
Set the step size pool as $\H =\{ \eta_i = \frac{\Gamma\sigma}{2G\sqrt{KT}}\cdot 2^{i-1} \mid i\in[N]\}$, where $N = 1+ \lceil\frac{1}{2} \log_2(1+2T) \rceil$ is the number of base-learners.
\textsc{Atlas} ensures that $\E[\DReg] \leq \O(\max\{\V_T^{\frac{1}{3}}T^{\frac{2}{3}},\sqrt{T}\})$, or simplified as $\O(\V_T^{\frac{1}{3}}T^{\frac{2}{3}})$ for non-degenerated cases of $V_T \geq \Theta(T^{-\frac{1}{2}})$.
\end{myThm}

Theorem~\ref{thm:overall-UOGD} shows that \textsc{Atlas} enjoys the same dynamic regret as UOGD with the (unknown) optimal step size, but unknown non-stationarity $\V_T$ is no more required in advance. 
Algorithmically, our online method maintains $N = \Theta(\log T)$ base-learners to achieve an optimal dynamic regret, which would be computationally acceptable given a logarithmic dependence on $T$.

\subsection{More Adaptive Algorithm by Exploiting Label Shift Structures}
\label{sec:more-adaptive}
\textsc{Atlas} is equipped with a minimax optimal dynamic regret, which indicates that it can safeguard the optimal theoretical property in the worst case. Worst-case optimality serves as the ``stress-testing'' for the robustness to non-stationarity environments~\citep{book'20:beyond-worst-case}. At the same time, more adaptive results beyond the worst-case analysis are also urgently desired, as improved performance is naturally expected in many easier situations when the shift admits specific patterns such as periodicity or gradual evolution.

To this end, we propose an improved algorithm called \textsc{Atlas-ada} with provably more adaptive guarantees. The key idea is to exploit the label shift patterns and reuse historical information to help the current online update~\citep{MLJ'20:Condor}. 
We build on the framework of \emph{optimistic online learning}~\citep{COLT'12:variation-Yang,conf/colt/RakhlinS13} by introducing a \emph{hint} function $H_t:\W\mapsto\R$ to encode shift patterns from historical data, which serves as an estimation of the expected risk $R_t(\w_t)$. Below, we start with a given hint function and describe the usage, and finally elaborate on how to design $H_t(\cdot)$ guided by the theory.

Similar to \textsc{Atlas}, the improved \textsc{Atlas-ada} also deploys a two-layer meta-base structure. The key difference lies in the usage of the hint function at both the base-level and meta-level.

\vspace{1mm}
\textbf{Base-algorithm.~~} Besides the gradient descent step as did in \textsc{Atlas}, another update step related to the hint function $H_t(\cdot)$ is performed. Concretely, the $i$-th base-learner updates the parameters by
\begin{equation}
\label{eq:IOMD}
  \wh_{t,i} = \Pi_{\W}[\wh_{t-1,i} - \eta_i \nabla \Rh_{t-1}(\w_{t-1,i})], \quad \w_{t,i} = \argmin\nolimits_{\w\in\W} \eta_i H_t(\w) + \nicefrac{1}{2}\cdot \norm{\w-\hat{\w}_{t,i}}_2^2,
\end{equation}
where $\hat{\w}_{t,i}$ is an intermediate output and $\w_{t,i}$ is the final returned model. When $H_t(\w) =0$ (i.e., without a hint function), the above two-step update simply degenerates to the same UOGD update in the base-learner of \textsc{Atlas} by noting that now $\w_{t,i} = \hat{\w}_{t,i}$.  In the general $H_t(\cdot)$ case, the second step in~\eqref{eq:IOMD} is crucial and can be regarded as another descent towards the direction specified by the hint function. As a result, this will reduce the regret whenever the hint function is set appropriately to approximate well the next-round risk function, which will be clear in the regret bound presented later.

\vspace{1mm}
\textbf{Meta-algorithm.~~} The meta-algorithm is used to track the best base-learner, and the hint function $H_t(\cdot)$ is also necessary to be considered in the update to achieve the adaptivity. To this end, we inject the hint function as the loss evaluation of the meta-algorithm, and then the weight 
is updated by $p_{t, i} \propto \exp\big({ -\varepsilon\big(\sum_{s=1}^{t-1}\Rh_s(\w_{s,i})+H_{t}(\w_{t,i})\big)}\big)$ for all $i \in [N]$, where 
$\epsilon$ is the learning rate that can be set properly without dependence on $V_T$. 
The key distinction to the meta-algorithm of \textsc{Atlas} is the additional loss $H_{t}(\w_{t,i})$ evaluated over the current local models $\{\w_{t,i}\}_{i=1}^N$ by the hint function.

The main procedures are presented in Algorithm~\ref{alg:atlas-ada-base} (base-algorithm) and Algorithm~\ref{alg:atlas-ada-meta} (meta-algorithm). We have the following dynamic regret bound for \textsc{Atlas-ada} (proof in Appendix~\ref{appendix:sec-proof-Atlas-ada}).
\begin{myThm}
\label{thm:overall-guarantees} 
Suppose the hint function $H_t: \W \mapsto \R$ is convex, satisfies $\max_{\w\in\W}\norm{\nabla H_t(\w)}_2\leq \max_{\w\in\W}\norm{\nabla\Rh_t(\w)}_2$, and is independent of current data $S_t$. Set the step size pool as $\H = \{ \eta_i = \frac{\Gamma\sigma}{\sqrt{\sigma^2+4G^2KT}}\cdot 2^{i-1} \mid i\in[N]\}$ with $N = 2+ \lceil \frac{1}{2}\log_2 (3T(1+4G^2KT/\sigma)) \rceil$. \textsc{Atlas-ada} ensures $\E[\DReg] \leq \O(\V_T^{1/3}\G_T^{1/3}T^{1/3})$,
where $G_T = \sum_{t=1}^T\E[\sup_{\w\in\W}\norm{\nabla \Rh_t(\w)-\nabla H_t(\w)}_2^2]$ measures the reusability of historical information, depending on label shift patterns and hint function designs. 
\end{myThm}

In the worst case, $\G_T$ is at most $\O(T)$ given a bounded gradient norm $\norm{\nabla H_t(\w)}_2$, and thus the bound presented in Theorem~\ref{thm:overall-guarantees} safeguards the same $\O(V_T^{1/3}T^{2/3})$ bound as \textsc{Atlas}. More importantly, when the hint function $H_t(\cdot)$ encodes beneficial information and is close to the risk function, the obtained bound can be substantially better than the minimax rate.

\begin{myRemark}[Implicit update]
\label{remark:implicit-update}
Problem-dependent dynamic regret was first presented in~\citep{NIPS'20:sword} for convex and smooth functions. However, their result critically relied on the \emph{smoothness} condition, which is not satisfied in our \ols case. Our key technical innovation is the \emph{implicit update} in the second step of~\eqref{eq:IOMD}. The previous method required the gradient-descent type update $\w_{t,i} = \Pi_{\w\in\W}[\wh_{t,i} - \eta_i \nabla H_t(\w_{t-1,i})]$, which can be deemed as an approximated optimization over the linearized loss $\inner{\nabla H_t(\w_{t-1,i})}{\w}$. By contrast, we directly updates over the \emph{original function} $H_t(\w)$, hence called the ``implicit'' update~\citep{ICML'10:Implict-OL,NIPS'20:Implict-V_T}. Albeit with slightly larger computational complexity (which will not be a barrier given a proper design of hint functions), our method enjoys the same dynamic regret \emph{without} smoothness, which could be of independent interest for general OCO purposes.
\end{myRemark}

\vspace{-1mm}
\textbf{Design of Hint Functions.~~} 
\label{sec:optimistic}
The hint functions should minimize the reusability measure $\G_T$ to sharpen the dynamic regret as suggested by Theorem~\ref{thm:overall-guarantees}. Recall that $\Rh_t(\w) =\sum_{k=1}^K \left[\hat{\bm{\mu}}_{y_t}\right]_k\cdot \Rh_0^k(\w)$. Thus, a natural construction is 
$H_t(\w) = \sum_{k=1}^K [\h_{y_t}]_k\cdot \Rh_0^k(\w)$  parametrized by \emph{hint priors} $\h_{y_t}\in\R^K$, which is used to estimate the class prior based on the past observed data $\{S_\tau\}_{\tau=1}^{t-1}$. Then, $G_T$ satisfies the \emph{bias-variance decomposition} (with bias term $\E[\norm{\bm{h}_{y_t} - \bm{\mu}_{y_t}}_2^2]$ and variance term $\E[\norm{\bm{\mu}_{y_t} - \hat{\bm{\mu}}_{y_t}}_2^2]$):
\begin{equation*}
\label{eq:upper-bound-optimism}
\G_T \leq KG^2\sum\nolimits_{t=1}^T\E\left[\norm{\h_{y_t} - \hat{\bm{\mu}}_{y_t}}_2^2\right] \leq 2KG^2\sum\nolimits_{t=1}^T\Big(\E\left[\norm{\bm{h}_{y_t} - \bm{\mu}_{y_t}}_2^2\right] + \E\left[\norm{\bm{\mu}_{y_t} - \hat{\bm{\mu}}_{y_t}}_2^2\right]\Big). 
\end{equation*}
Setting $\bm{h}_{y_t}= \bm{\mu}_{y_t}$ will make the upper bound tightest possible, though the underlying class prior $\bm{\mu}_{y_t}$ is not accessible in practice. So the design of the hint function is actually a task of approximating it with different parts of previous data guided by prior patterns. In the experiments, we design four hint functions by encoding different knowledge, including Forward Hint (\emph{Fwd}), Window Hint (\emph{Win}), Periodic Hint (\emph{Peri}), and Online KMeans Hint (\emph{OKM}). More details are in Appendix~\ref{sec-appendix:atala-ada-algorithm}.

%% file: sections/experiments.tex
\section{Experiments}
\label{sec:experiments}
In this section, we conduct extensive experiments to validate the effectiveness of the proposed methods (\textsc{Atlas} and \textsc{Atlas-Ada}) and justify the theoretical findings. We begin this section with a brief introduction to the experimental setups (more details are deferred to Appendix~\ref{sec-appendix:experiments}) and then present empirical results on the synthetic and real-world data, respectively.

\vspace{1mm}
\textbf{Experiments Setup.~~} We compare seven algorithms in various experimental configurations. The contenders include a baseline that predicts with the initial model directly (\emph{{FIX}}), three \ols algorithms proposed by the previous work~\citep{NIPS'21:Online-LS} ({\emph{ROGD}}, {\emph{FTH}}, and {\emph{FTFWH}}), and our proposals ({\emph{UOGD}}, {\textsc{Atlas}}, and {\textsc{Atlas-Ada}}) with the logistic regression model. Besides, we simulate four types of label shift on synthetic and benchmark data to capture different distribution change patterns. Two of them, including \texttt{Sine Shift} (\raisebox{-0.25ex}{\includegraphics[height=1.8ex, width=3ex]{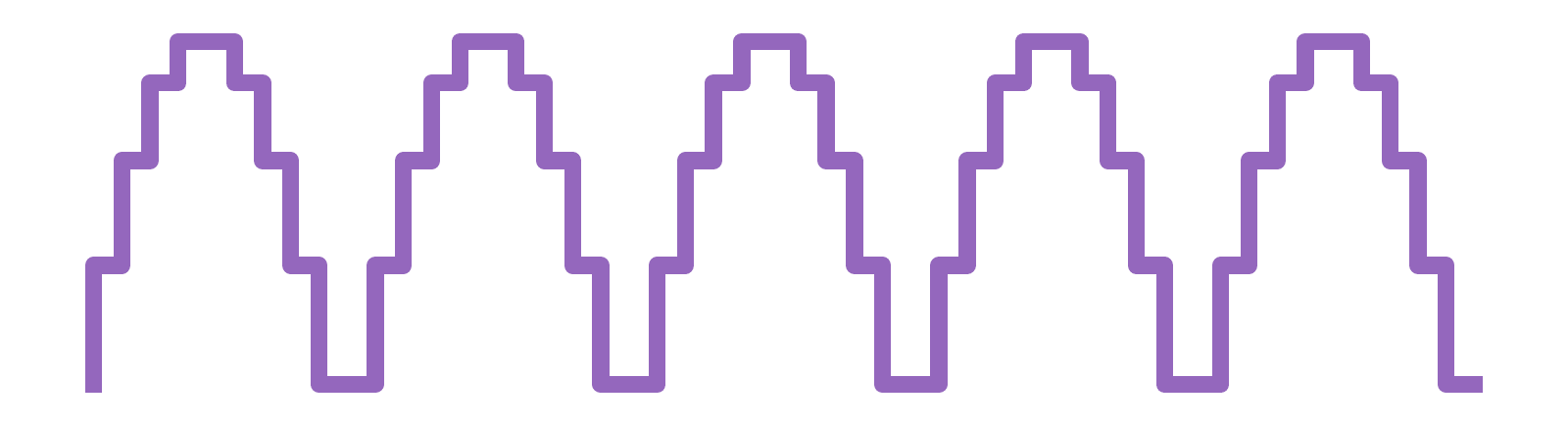}}) and \texttt{Square Shift} (\raisebox{-0.25ex}{\includegraphics[height=1.8ex, width=3ex]{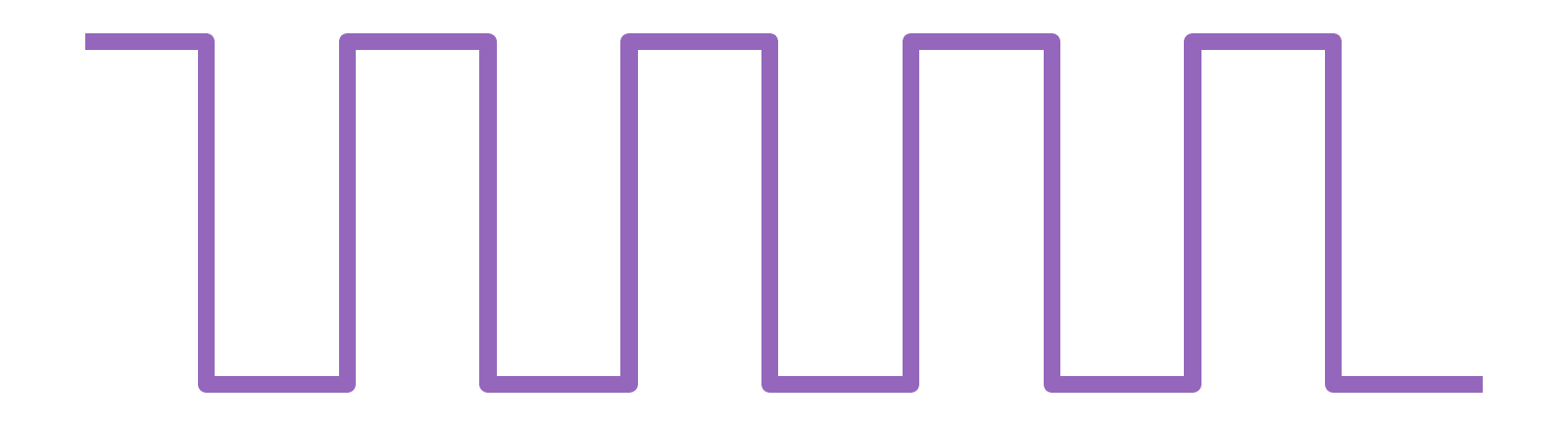}}) change in a periodic pattern. The other two have no periodic structure but are introduced to capture different shift intensities. The underlying distribution changes slowly in the \texttt{Linear Shift} (\raisebox{-0.25ex}{\includegraphics[height=1.8ex, width=3ex]{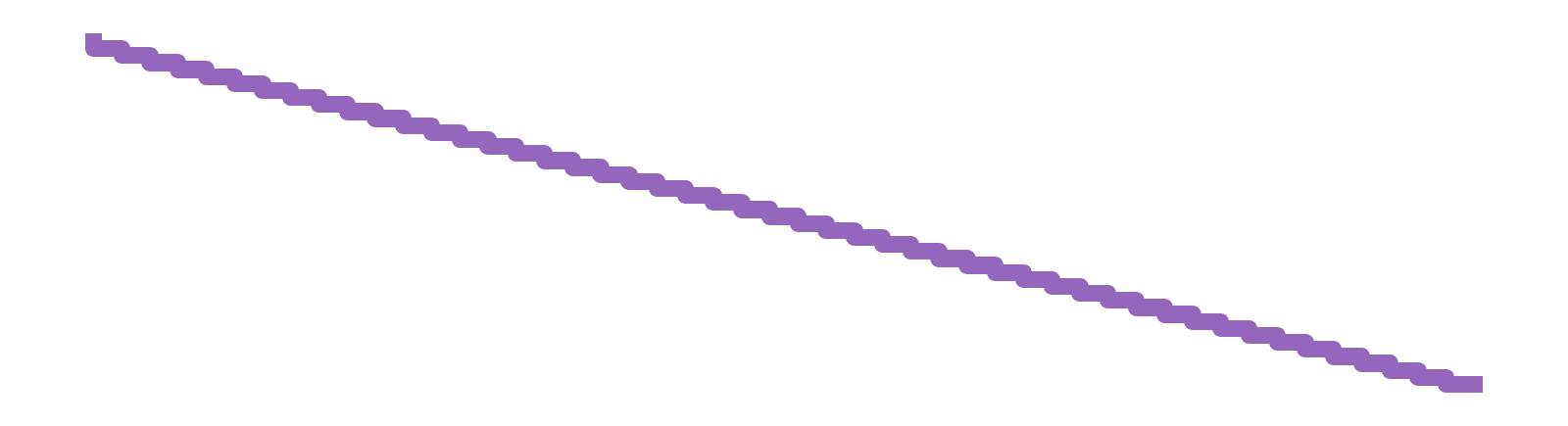}}) while changes fast in \texttt{Bernoulli Shift}~ (\raisebox{-0.25ex}{\includegraphics[height=1.8ex, width=3ex]{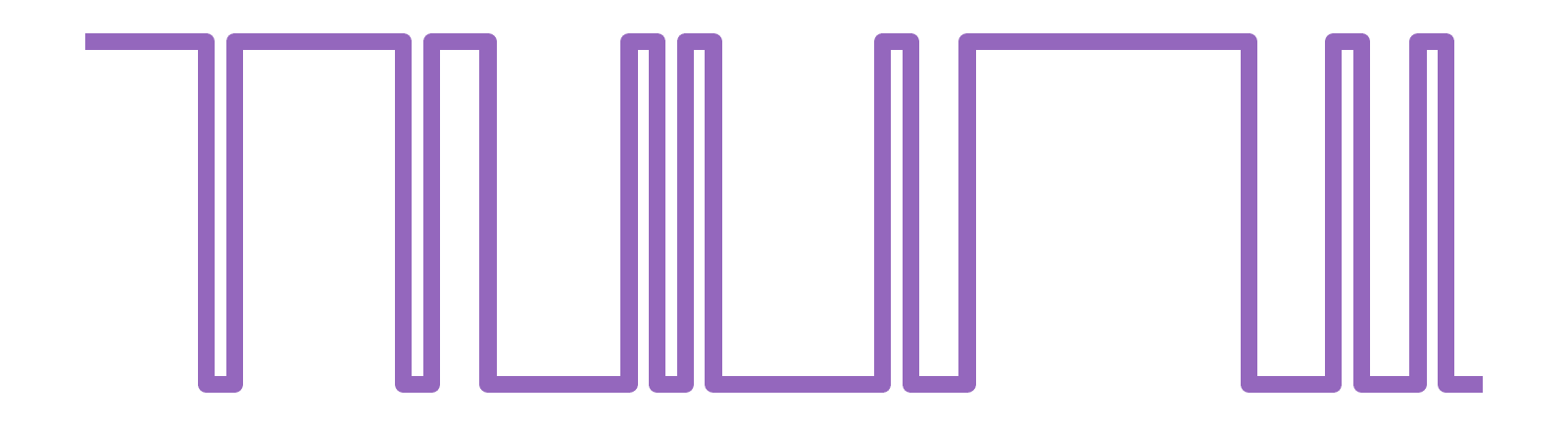}}). We repeat all experiments for five times and evaluate the contenders by the average error $\frac{1}{T} \sum_{t=1}^{T} \frac{1}{\vert S_t \vert} \sum_{n=1}^{\vert S_t \vert} \indicator\left[f\left(\mathbf{w}, \mathbf{x}_{n}\right) \neq y_{n}\right]$ over $T = 10,000$ rounds.

\subsection{Illustrations on Synthetic Data}
This subsection first compares all contenders on the synthetic data. Then, we further illustrate the effectiveness of our proposal by a closer look at the two key components, including a meta-base structure for step size search and a hint function for historical information reuse.

\input{sections/experiments/meta-weights}

\vspace{1mm}
\textbf{Overall Performance Comparison.~~}
 Table~\ref{tab:new-syn-results} compares $\textsc{Atlas}$ with other methods when $N_t = 100$ samples are received at every iteration. Basically, \emph{FIX} is inferior to the online methods, which shows the necessity of designing online algorithms for the \ols problem. Besides, \emph{UOGD} outperforms \emph{ROGD}, the OGD algorithm running with the risk estimator proposed by~\citep{NIPS'21:Online-LS}. The comparison demonstrates the empirical superiority of our estimator besides its benign theoretical properties. Moreover, the \textsc{Atlas} surpasses almost all other methods in the four shift patterns. In particular, it achieves a significant advantage over \emph{UOGD} (with step size $\eta = \Theta(T^{-1/2})$) when the environments change relatively fast (\texttt{Squ}, \texttt{Sin}, and \texttt{Ber}). The results justify our theoretical finding that the small step size $\Theta(T^{-1/2})$ suggested by the static regret analysis is unsuitable for the dynamic environments. Our method can better adapt to the changing environments by enjoying the dynamic regret guarantees.

\vspace{1mm}
\textbf{Effectiveness of Meta-Base Structure.~~} One key component of our method is the meta-base structure to address the non-stationarity. To better illustrate its effectiveness, we visualize the weights $p_{t,i}$ assigned for each base-learner of \textsc{Atlas}. As shown in Figure~\ref{fig-meta-weights-part2}, the meta-algorithm can quickly assign larger weights to appropriate base-learners along the learning process. Specifically, Figure~\ref{tab:main-new-syn-results-a} illustrates the case of slowly changing environments, where more weights are assigned to the base-learners with small step sizes. In the fast-changing case, see Figure~\ref{tab:main-new-syn-results-b}, larger step sizes are preferred. The results show that our algorithm can adaptively track the suitable step sizes according to the shift intensity of environments. Additional results for other shifts can be found in Figure~\ref{fig-meta-weights-part2}.

\input{sections/experiments/new-syn-results-optimism}

\vspace{1mm}
\textbf{Effectiveness of Using Hint Functions.~~} Table~\ref{tab:new-syn-results-optimism} reports the performance of \textsc{Atlas-Ada} with four different hint functions under sample sizes are $N_t = 1, 10, $ and $100$. All hint functions improve over vanilla \textsc{Atlas} in most cases. When $N_t$ is reasonably large, the Fwd hint, performing transductive learning with current unlabeled data, achieves the best performance. While, the OKM hint, which learns previous patterns by online k-means, is the best choice for a small sample size case ($N_t =1$). 

To illustrate how the hint function works, we further vary the buffer size of the Periodic Hint (\emph{Peri}) on the \texttt{Squ} environment (\raisebox{-0.25ex}{\includegraphics[height=1.8ex, width=3ex]{figures/sketch/squ_sketch.pdf}}), which shifts in a periodic length $L=40$. As shown in Figure~\ref{fig:optimism-po}, when the buffer size matches the multiples of length $L$, \emph{Peri} can significantly improve the vanilla \textsc{Atlas}. Besides the improvement, our method is shown to achieve comparable performance with \textsc{Atlas} even if the buffer size is misspecified. The result validates our theoretical guarantee of the safety of using hint functions (the readers can refer to the paragraph above Remark~\ref{remark:implicit-update}).

\subsection{Comparisons on Real-world Data}
\label{sec:expr-benchmark}

We conduct experiments on real-world data, including six benchmark datasets (ArXiv, EuroSAT, MNIST, Fashion, CIFAR10, and CINIC10) and the SHL dataset~\citep{DBLP:journals/access/GjoreskiCWMMVR18} for the real-life locomotion detection task. Table~\ref{tab:benchmark-part1} reports the averaged error of different algorithms, which shows a similar tendency as the results in the synthetic experiments. When the distribution changes rapidly ({\texttt{Ber}}), \textsc{Atlas} and \textsc{Atlas-Ada} outperform other contenders. Similar results are also observed in the Sine and Square shift, see Appendix~\ref{appendix-expr-benchmark} for details. Even in a relatively stationary environment ($\texttt{Lin}$), our algorithms are comparable with the best algorithm (UOGD), which is specifically designed for stationary cases. The above results validate the adaptivity of the proposed algorithms.

\input{sections/experiments/benchmark_part1.tex}

\input{sections/experiments/locomotion}

Further, we highlight the results on the locomotion detection task. The task aims to distinguish six types of locomotion with sensor data from mobile phones. Figure~\ref{fig:result-shl-performance} reports the averaged error of all contenders, which shows the superiority of our proposals \textsc{Atlas} and \textsc{Atlas-ada} (with the \emph{OKM} hint) over the entire time horizons. In addition, to further validate the adaptivity of our algorithm to the underlying environment regardless of the fast or slow changes, we simulate various shift intensities by sampling the original data with different frequencies. Figure~\ref{fig:result-shl-heatmap} shows the weight assignment of \textsc{Atlas-Ada} for each step size in the final round. Our method automatically selects a larger step size for larger $V_T$ while tracking a small step size in a relatively static scenario.

%% file: sections/experiments/meta-weights.tex
\begin{figure}[!t]  
    
        \begin{minipage}[t]{0.47\columnwidth}
            \vspace{0pt}
            \captionof{table}{Average error (\%) for different algorithms under various simulated shifts for the synthetic data. The best algorithms are emphasized in bold (paired $t$-test at a 5\% confidence level).}
            \vspace{1mm}
            \resizebox{\textwidth}{!}{
\begin{tabular}{ccccc}
        \toprule
              & \texttt{Lin}                & \texttt{Squ}       & \texttt{Sin}       & \texttt{Ber}       \\ \midrule
        FIX   & 7.87$\pm$0.03          & 7.87$\pm$0.02 & 7.34$\pm$0.03 & 7.79$\pm$0.02 \\
        FTH   & \textbf{4.70$\pm$0.02} & 6.50$\pm$0.01 & 6.36$\pm$0.03 & 6.60$\pm$0.01 \\
        FTFWH & 5.27$\pm$0.02          & 6.52$\pm$0.01 & 6.36$\pm$0.02 & 6.60$\pm$0.01 \\
        ROGD  & 6.08$\pm$0.01          & 7.11$\pm$0.01 & 6.87$\pm$0.02 & 6.40$\pm$0.01 \\
        UOGD  & 5.35$\pm$0.02          & 6.17$\pm$0.01 & 6.37$\pm$0.01 & 5.46$\pm$0.05 \\
        \multicolumn{1}{l}{ATLAS} &
          \multicolumn{1}{l}{5.44$\pm$0.02} &
          \multicolumn{1}{l}{\textbf{4.27$\pm$0.02}} &
          \multicolumn{1}{l}{\textbf{5.75$\pm$0.01}} &
          \multicolumn{1}{l}{\textbf{4.04$\pm$0.07}} \\ \bottomrule
        \end{tabular}%
    }
    \vspace{-2mm}
   
    \label{tab:new-syn-results}
    \end{minipage} \vspace{-4mm}
        \hfill
        \begin{minipage}[t]{0.5\columnwidth}
            \vspace{0pt}
            \centering 
            \begin{minipage}[b]{.9\textwidth}
                 \includegraphics[width=\textwidth]{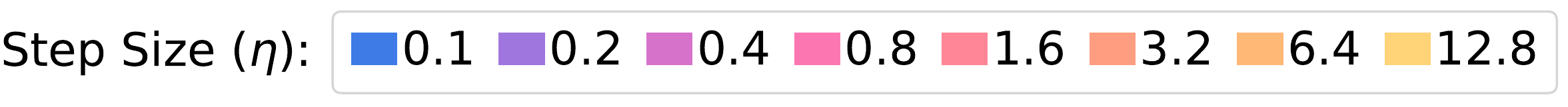} 
            \end{minipage}\vspace{-2mm}
            \vfill
            \subfigure[\texttt{Linear shift}]{
            \begin{minipage}[b]{0.46\textwidth}
               \label{tab:main-new-syn-results-a} \includegraphics[width=\textwidth, height=.575\textwidth]{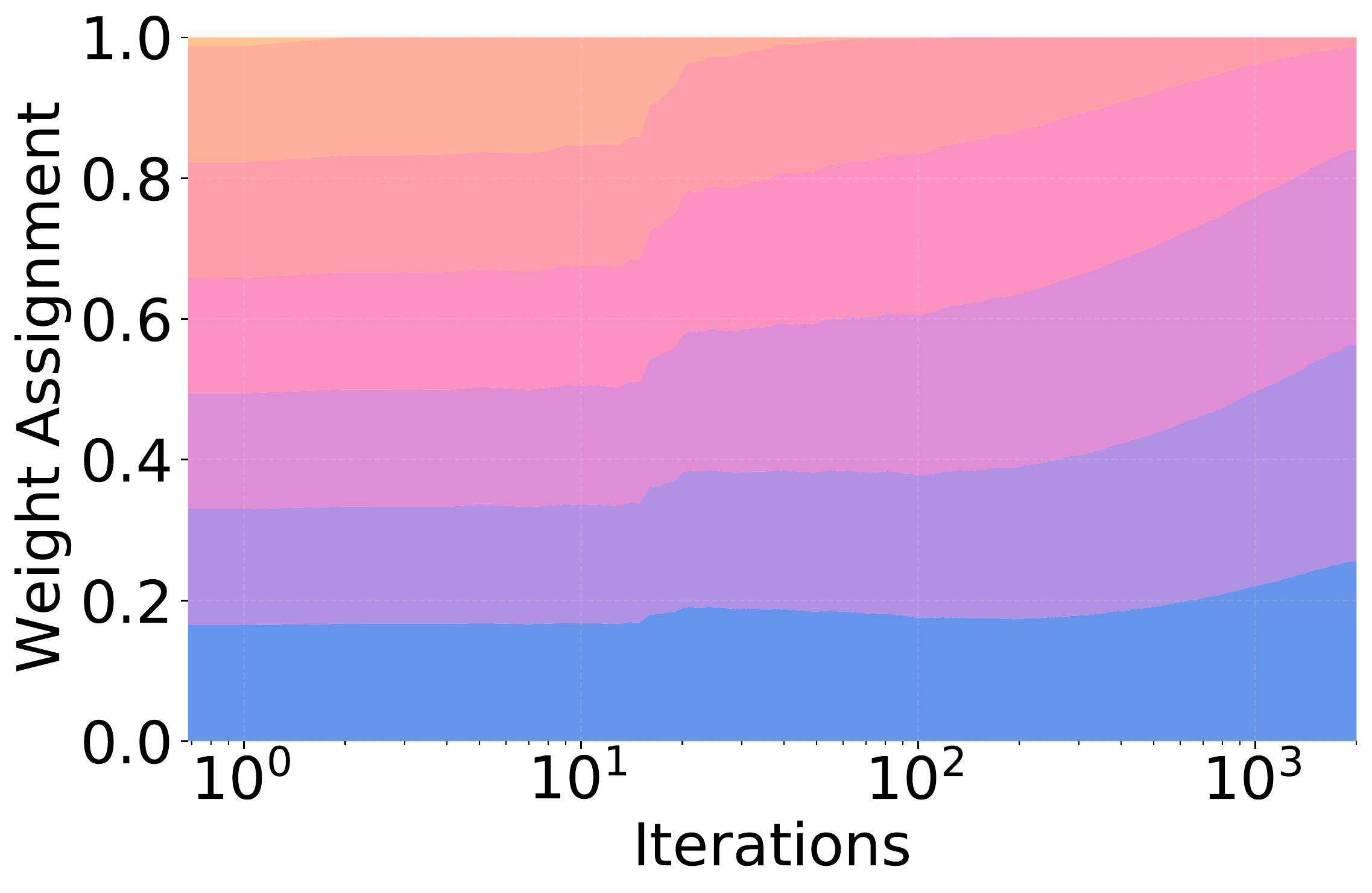} 
            \end{minipage}
            }
            \subfigure[\texttt{Bernoulli shift}]{
            \begin{minipage}[b]{0.46\textwidth}
               \label{tab:main-new-syn-results-b} \includegraphics[width=\textwidth,  height=.575\textwidth]{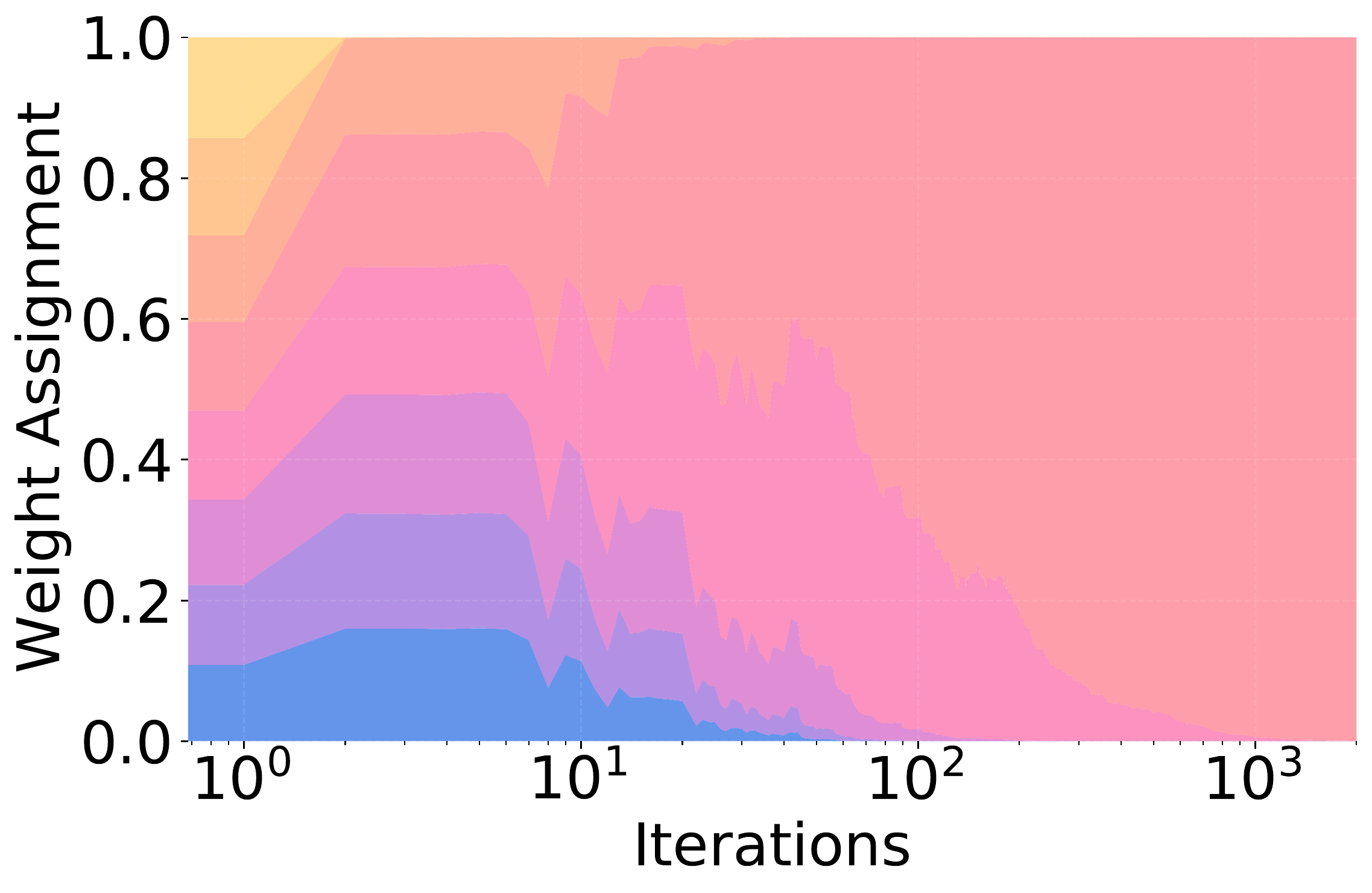}  
            \end{minipage}
            }
            \vspace{-2mm}
            \caption{Weight assigned of the \textsc{Atlas} algorithm for each step size along the learning process. Different colors are used to indicate different step sizes. } \label{fig-meta-weights}
        \end{minipage}
 \end{figure}

%% file: sections/experiments/new-syn-results-optimism.tex
\begin{table}[!t]
    \centering
    \vspace{-1mm}
    \caption{Average error (\%) for \textsc{Atlas-Ada} with  four hint functions under different sample sizes. The best one is emphasized in bold. Besides, $\bullet$ indicates a better result than \textsc{Atlas} without hint (\emph{None}).}
    \vspace{2mm}
    \label{tab:new-syn-results-optimism}
    \resizebox{\textwidth}{!}{%
    \begin{tabular}{crrrrrrrrrrrrrrr}
      \toprule
    \multirow{2}{*}{\begin{tabular}[c]{@{}c@{}}\texttt{Shift} \\ \texttt{Type}\end{tabular}} &
        \multicolumn{5}{c}{\textbf{Sample Size: 1}} &
        \multicolumn{5}{c}{\textbf{Sample Size: 10}} &
        \multicolumn{5}{c}{\textbf{Sample Size: 100}} \\    \cmidrule(lr){2-6} \cmidrule(lr){7-11} \cmidrule(lr){12-16}
        \multicolumn{1}{c}{} &
          \multicolumn{1}{r}{\emph{None}} &
          \multicolumn{1}{r}{\emph{Win}}   &
          \multicolumn{1}{r}{\emph{Peri}}   &
          \multicolumn{1}{r}{\emph{Fwd}} &
          \multicolumn{1}{r}{\emph{OKM}} &
          \multicolumn{1}{r}{\emph{None}} &
          \multicolumn{1}{r}{\emph{Win}}   &
          \multicolumn{1}{r}{\emph{Peri}}   &
          \multicolumn{1}{r}{\emph{Fwd}} &
          \multicolumn{1}{r}{\emph{OKM}} &
          \multicolumn{1}{r}{\emph{None}} &
          \multicolumn{1}{r}{\emph{Win}}   &
          \multicolumn{1}{r}{\emph{Peri}}   &
          \multicolumn{1}{r}{\emph{Fwd}} &
          \multicolumn{1}{r}{\emph{OKM}} \\ \midrule
        \multirow{2}{*}{\texttt{Lin}} &
        6.28 &
        $\bullet$ 5.89 &
        $\bullet$ 5.99 &
        $\bullet$ 6.01 &
        \textbf{$\bullet$ 5.35} &
        5.61 &
        $\bullet$ 5.47 &
        $\bullet$ 5.43 &
        $\bullet$ 5.53 &
        \textbf{$\bullet$ 5.42} &
        5.44 &
        5.44 &
        \textbf{$\bullet$ 5.38} &
        $\bullet$ 5.40 &
        5.45 \\
       &
        $\pm$ 0.21 &
        $\pm$ 0.26 &
        $\pm$ 0.29 &
        $\pm$ 0.31 &
        $\pm$ 0.31 &
        $\pm$ 0.04 &
        $\pm$ 0.04 &
        $\pm$ 0.03 &
        $\pm$ 0.05 &
        $\pm$ 0.05 &
        $\pm$ 0.02 &
        $\pm$ 0.03 &
        $\pm$ 0.02 &
        $\pm$ 0.02 &
        $\pm$ 0.03 \\
        \multirow{2}{*}{\texttt{Squ}} &
        6.03 &
        $\bullet$ 5.83 &
        $\bullet$ 5.27 &
        5.88 &
        \textbf{$\bullet$ 5.07} &
        4.59 &
        4.69 &
        $\bullet$ 3.85 &
        \textbf{$\bullet$ 3.72} &
        $\bullet$ 3.91 &
        4.27 &
        4.68 &
        $\bullet$ 3.39 &
        \textbf{$\bullet$ 3.33} &
        $\bullet$ 3.46 \\
       &
        $\pm$ 0.23 &
        $\pm$ 0.24 &
        $\pm$ 0.20 &
        $\pm$ 0.23 &
        $\pm$ 0.35 &
        $\pm$ 0.02 &
        $\pm$ 0.02 &
        $\pm$ 0.04 &
        $\pm$ 0.02 &
        $\pm$ 0.03 &
        $\pm$ 0.02 &
        $\pm$ 0.02 &
        $\pm$ 0.03 &
        $\pm$ 0.03 &
        $\pm$ 0.04 \\
        \multirow{2}{*}{\texttt{Sin}} &
        6.90 &
        $\bullet$ 6.58 &
        $\bullet$ 6.59 &
        $\bullet$ 6.43 &
        \textbf{$\bullet$ 5.25} &
        6.12 &
        $\bullet$ 5.99 &
        $\bullet$ 5.83 &
        \textbf{$\bullet$ 5.78} &
        $\bullet$ 5.86 &
        5.75 &
        5.78 &
        $\bullet$ 5.53 &
        \textbf{$\bullet$ 5.48} &
        $\bullet$ 5.58 \\
       &
        $\pm$ 0.22 &
        $\pm$ 0.22 &
        $\pm$ 0.25 &
        $\pm$ 0.26 &
        $\pm$ 0.22 &
        $\pm$ 0.07 &
        $\pm$ 0.06 &
        $\pm$ 0.05 &
        $\pm$ 0.05 &
        $\pm$ 0.04 &
        $\pm$ 0.01 &
        $\pm$ 0.01 &
        $\pm$ 0.00 &
        $\pm$ 0.01 &
        $\pm$ 0.00 \\
        \multirow{2}{*}{\texttt{Ber}} &
        5.55 &
        $\bullet$ 5.42 &
        $\bullet$ 5.43 &
        5.63 &
        \textbf{$\bullet$ 4.69} &
        4.39 &
        4.45 &
        4.43 &
        \textbf{$\bullet$ 3.66} &
        $\bullet$ 3.73 &
        4.04 &
        4.29 &
        4.26 &
        \textbf{$\bullet$ 3.19} &
        $\bullet$ 3.45 \\
       &
        $\pm$ 0.09 &
        $\pm$ 0.11 &
        $\pm$ 0.09 &
        $\pm$ 0.16 &
        $\pm$ 0.17 &
        $\pm$ 0.10 &
        $\pm$ 0.08 &
        $\pm$ 0.10 &
        $\pm$ 0.10 &
        $\pm$ 0.06 &
        $\pm$ 0.07 &
        $\pm$ 0.06 &
        $\pm$ 0.06 &
        $\pm$ 0.07 &
        $\pm$ 0.11 \\ \bottomrule
      \end{tabular}
      }\vspace{-6mm}
    \end{table}

%% file: sections/experiments/benchmark_part1.tex
\begin{table*}[!t]
  \centering
  \caption{Average error (\%) of different algorithms on various real-world datasets (\texttt{Lin} and \texttt{Ber}). We report  the mean and standard deviation over five runs. The best algorithms are emphasized in bold. ``$\bullet$'' indicates the algorithms that are significantly inferior to \textsc{Atlas-ada} by the paired $t$-test at a $5\%$ significance level. Here AT-ADA represents \textsc{Atlas-ada} (with \emph{OKM}). The online sample size is set as $N_t = 10$. }
  \vspace{-1mm}
  \resizebox{\textwidth}{!}{%
  \begin{tabular}{crrrrrrrrrrrrrr}
    \toprule
    &
         \multicolumn{7}{c}{\texttt{Lin}} &
         \multicolumn{7}{c}{\texttt{Ber}}
           \\ \cmidrule(lr){2-8} \cmidrule(lr){9-15}
         & 
         \multicolumn{1}{c}{\textbf{FIX}} &
          \multicolumn{1}{c}{\textbf{FTH}} &
          \multicolumn{1}{c}{\textbf{FTFWH}} &
          \multicolumn{1}{c}{\textbf{ROGD}} &
          \multicolumn{1}{c}{\textbf{UOGD}} &
          \multicolumn{1}{c}{\textbf{ATLAS}} &
          \multicolumn{1}{c}{\textbf{AT-ADA}} &
          \multicolumn{1}{c}{\textbf{FIX}} &
          \multicolumn{1}{c}{\textbf{FTH}} &
          \multicolumn{1}{c}{\textbf{FTFWH}} &
          \multicolumn{1}{c}{\textbf{ROGD}} &
          \multicolumn{1}{c}{\textbf{UOGD}} &
          \multicolumn{1}{c}{\textbf{ATLAS}} &
          \multicolumn{1}{c}{\textbf{AT-ADA}}
           \\ \midrule
    \multirow{2}{*}{\textbf{ArXiv}} &
      $\bullet$ 30.28 &
      $\bullet$ 28.18 &
      $\bullet$ 25.74 &
      $\bullet$ 23.09 &
      \textbf{21.04} &
      $\bullet$ 22.10 &
      21.28 &
      $\bullet$ 30.63 &
      $\bullet$ 27.69 &
      $\bullet$ 28.50 &
      $\bullet$ 24.82 &
      $\bullet$ 21.53 &
      $\bullet$ 21.11 &
      \textbf{20.58} \\
     &
      $\pm$0.07 &
      $\pm$0.28 &
      $\pm$0.21 &
      $\pm$0.20 &
      $\pm$0.11 &
      $\pm$0.09 &
      $\pm$0.09 &
      $\pm$0.20 &
      $\pm$0.13 &
      $\pm$0.19 &
      $\pm$0.11 &
      $\pm$0.68 &
      $\pm$0.70 &
      $\pm$0.69 \\
    \multirow{2}{*}{\textbf{EuroSAT}} &
      $\bullet$ 14.06 &
      $\bullet$ 11.16 &
      $\bullet$ {\color{white} 0}9.78 &
      $\bullet$ 12.56 &
      \textbf{7.04} &
      $\bullet$ {\color{white} 0}7.19 &
      7.13 &
      $\bullet$ 14.12 &
      $\bullet$ 10.48 &
      $\bullet$ 10.50 &
      $\bullet$ {\color{white} 0}9.06 &
      $\bullet$ {\color{white} 0}7.28 &
      $\bullet$ {\color{white} 0}6.99 &
      \textbf{6.91} \\
     &
      $\pm$0.09 &
      $\pm$0.11 &
      $\pm$0.12 &
      $\pm$3.16 &
      $\pm$0.11 &
      $\pm$0.10 &
      $\pm$0.11 &
      $\pm$0.13 &
      $\pm$0.09 &
      $\pm$0.08 &
      $\pm$0.05 &
      $\pm$0.04 &
      $\pm$0.03 &
      $\pm$0.05 \\
    \multirow{2}{*}{\textbf{MNIST}} &
      $\bullet$ {\color{white} 0}1.79 &
      $\bullet$ {\color{white} 0}1.38 &
      $\bullet$ {\color{white} 0}1.20 &
      $\bullet$ {\color{white} 0}1.25 &
      \textbf{1.06} &
      \textbf{1.06} &
      \textbf{1.06} &
      $\bullet$ {\color{white} 0}1.81 &
      $\bullet$ {\color{white} 0}1.29 &
      $\bullet$ {\color{white} 0}1.34 &
      $\bullet$ {\color{white} 0}1.33 &
      $\bullet$ {\color{white} 0}1.12 &
      \textbf{1.03} &
      \textbf{1.03} \\
     &
      $\pm$0.02 &
      $\pm$0.03 &
      $\pm$0.02 &
      $\pm$0.02 &
      $\pm$0.02 &
      $\pm$0.02 &
      $\pm$0.02 &
      $\pm$0.05 &
      $\pm$0.03 &
      $\pm$0.03 &
      $\pm$0.03 &
      $\pm$0.02 &
      $\pm$0.02 &
      $\pm$0.02 \\
    \multirow{2}{*}{\textbf{Fashion}} &
    $\bullet$ 11.86 &
    $\bullet$ {\color{white} 0}8.47 &
      \textbf{7.84} &
      8.18 &
      7.95 &
      $\bullet$ {\color{white} 0}8.36 &
      8.04 &
      $\bullet$ 11.85 &
      $\bullet$ {\color{white} 0}8.48 &
      $\bullet$ {\color{white} 0}8.69 &
      $\bullet$ {\color{white} 0}8.72 &
      $\bullet$ {\color{white} 0}8.23 &
      $\bullet$ {\color{white} 0}7.91 &
      \textbf{7.69} \\
     &
      $\pm$0.04 &
      $\pm$0.07 &
      $\pm$0.06 &
      $\pm$0.07 &
      $\pm$0.08 &
      $\pm$0.07 &
      $\pm$0.08 &
      $\pm$0.09 &
      $\pm$0.11 &
      $\pm$0.10 &
      $\pm$0.08 &
      $\pm$0.12 &
      $\pm$0.12 &
      $\pm$0.12 \\
    \multirow{2}{*}{\textbf{CIFAR10}} &
    $\bullet$ 20.77 &
    $\bullet$ 17.36 &
      15.77 &
      $\bullet$ 18.45 &
      \textbf{15.54} &
      $\bullet$ 15.77 &
      15.62 &
      $\bullet$ 20.82 &
      $\bullet$ 17.06 &
      $\bullet$ 16.96 &
      $\bullet$ 17.66 &
      $\bullet$ 15.93 &
      $\bullet$ 14.98 &
      \textbf{14.80} \\
     &
      $\pm$0.12 &
      $\pm$0.14 &
      $\pm$0.12 &
      $\pm$0.47 &
      $\pm$0.15 &
      $\pm$0.11 &
      $\pm$0.14 &
      $\pm$0.12 &
      $\pm$0.14 &
      $\pm$0.15 &
      $\pm$0.13 &
      $\pm$0.29 &
      $\pm$0.30 &
      $\pm$0.29 \\
    \multirow{2}{*}{\textbf{CINIC10}} &
    $\bullet$ 33.98 &
      $\bullet$ 28.85 &
      $\bullet$ 26.87 &
      $\bullet$ 32.54 &
      \textbf{26.21} &
      $\bullet$ 26.66 &
      26.38 &
      $\bullet$ 34.11 &
      $\bullet$ 28.48 &
      $\bullet$ 28.44 &
      $\bullet$ 28.90 &
      $\bullet$ 26.63 &
      $\bullet$ 25.85 &
      \textbf{25.63} \\
     &
      $\pm$0.22 &
      $\pm$0.10 &
      $\pm$0.13 &
      $\pm$2.59 &
      $\pm$0.15 &
      $\pm$0.19 &
      $\pm$0.16 &
      $\pm$0.35 &
      $\pm$0.17 &
      $\pm$0.19 &
      $\pm$0.19 &
      $\pm$0.55 &
      $\pm$0.58 &
      $\pm$0.60 \\ \bottomrule
    \end{tabular}
  }
  \label{tab:benchmark-part1}
  \end{table*}

%% file: sections/experiments/locomotion.tex
\begin{figure}[!t] 
    \begin{minipage}[t]{0.34\columnwidth}
    \centering
    \includegraphics[width=.88\textwidth]{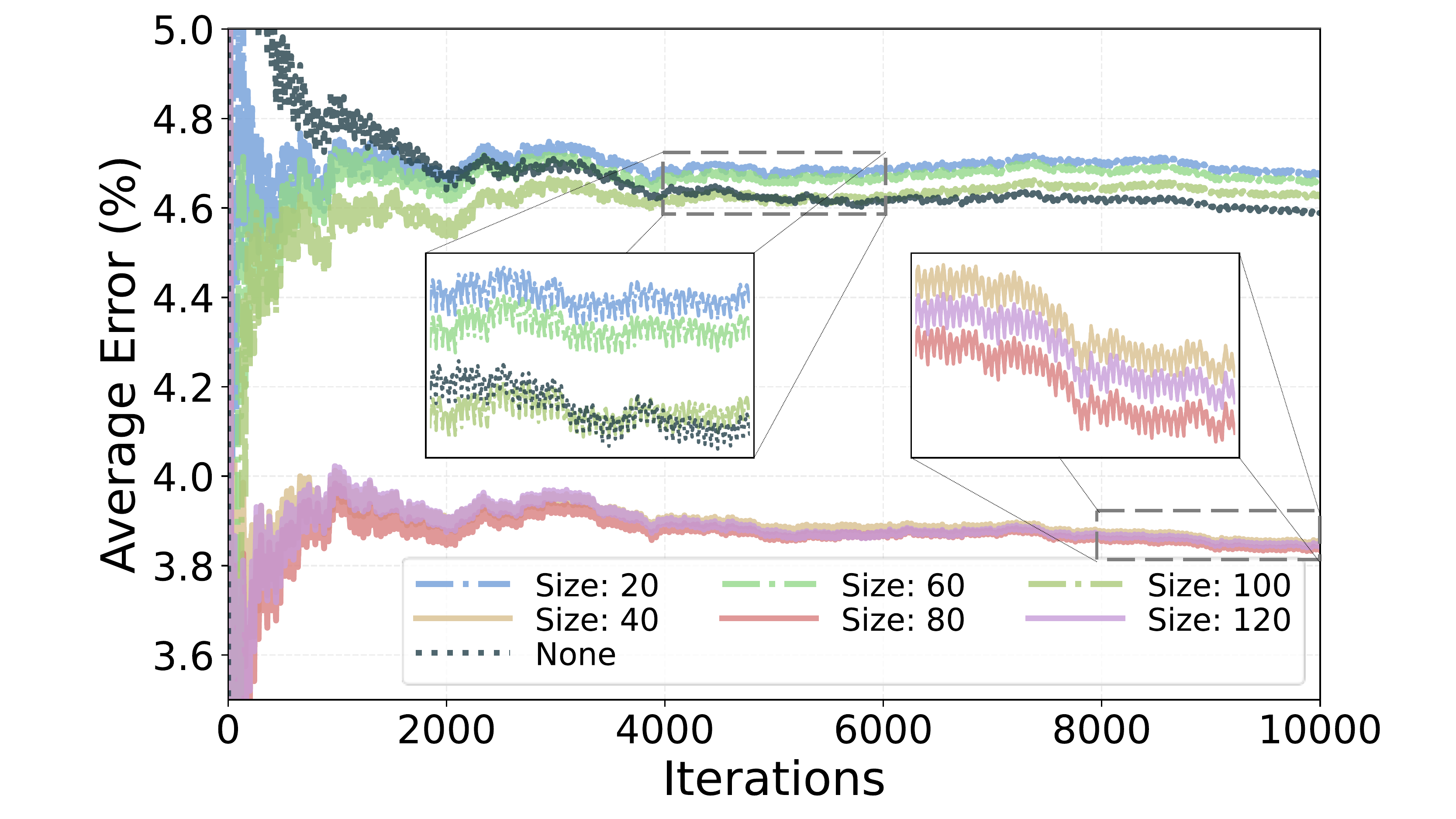} 
    \vspace{-1mm}
    \caption{Average error of \emph{Peri} hint with different buffer sizes. \textsc{Atlas-Ada} enjoys a significant improvement with proper buffer sizes while still safeguarding similar performance as \textsc{Atlas} even the size is misspecified.}
    \label{fig:optimism-po}
    \end{minipage}
    \hfill
	\begin{minipage}[t]{0.62\columnwidth}
        \vspace{-29mm}
        \subfigure[overall performance]{
            \includegraphics[height=0.3\textwidth]{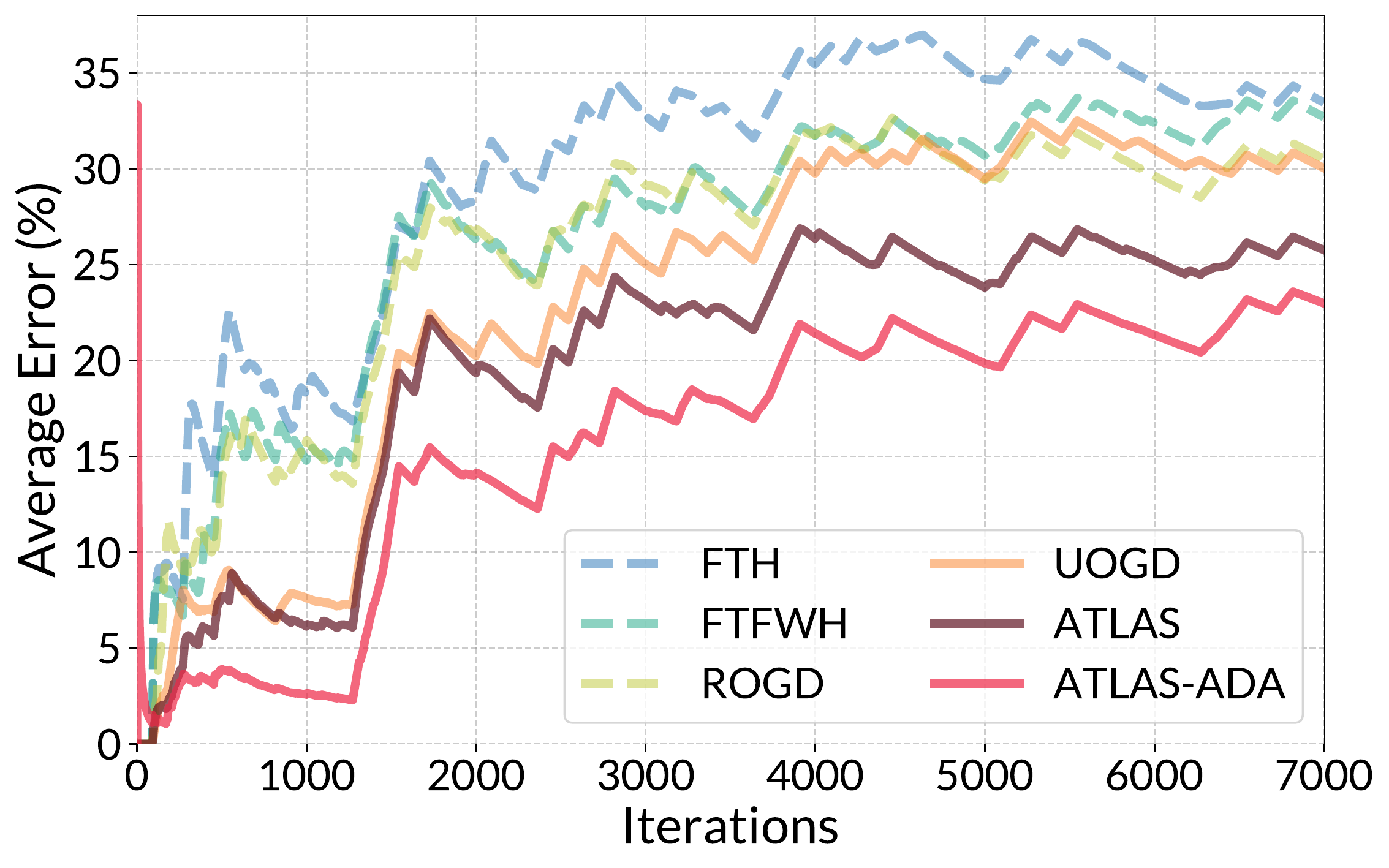}
            \label{fig:result-shl-performance}
        }
        \subfigure[weight heatmap]{
            \includegraphics[width=0.45\textwidth]{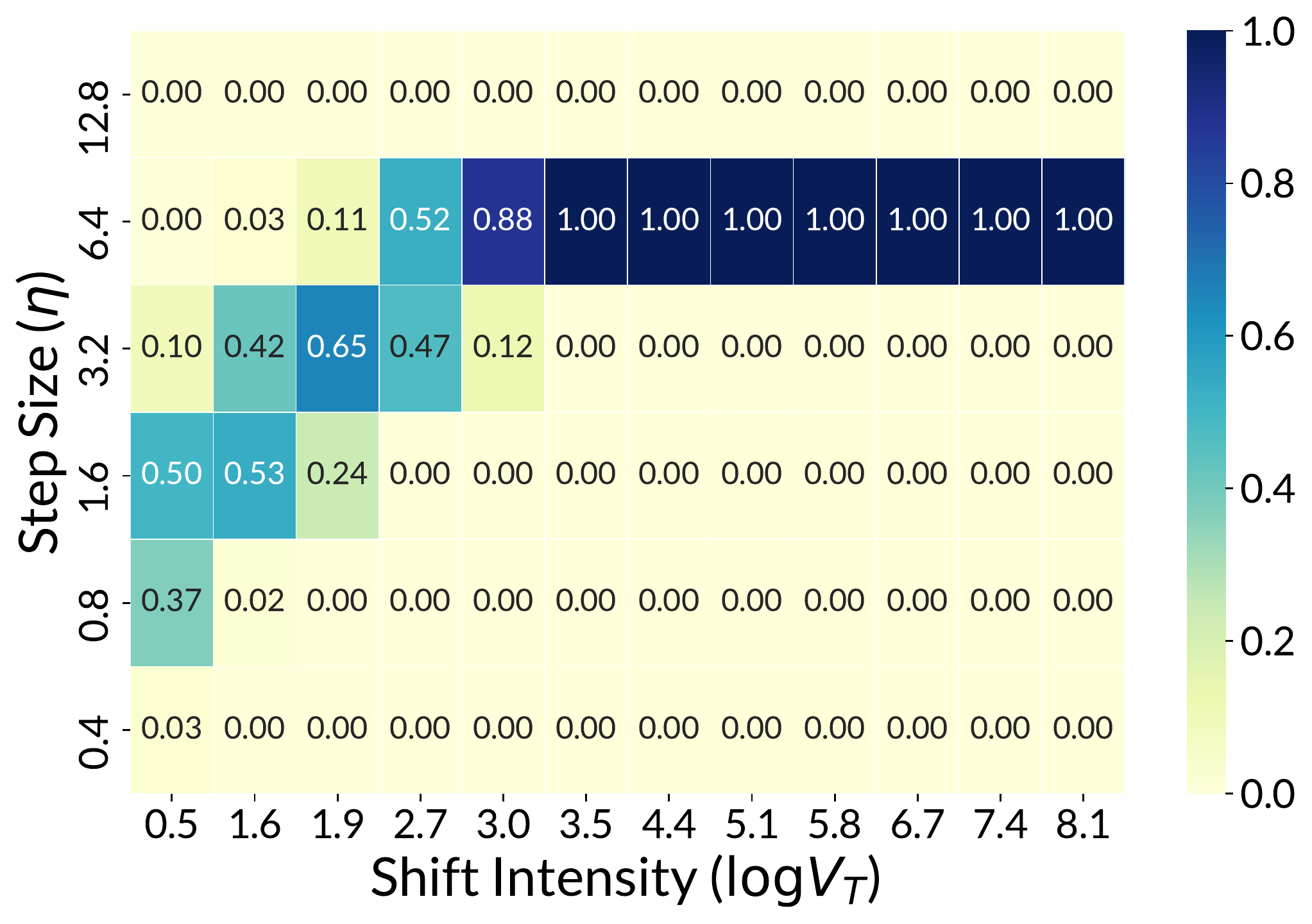}
            \label{fig:result-shl-heatmap}
        }
    \vspace{-1mm}
    \caption{(a) Overall performance comparison on the locomotion detection task.	(b) The heatmap of the final round weights for each step size. A larger $V_T$ implies a larger intensity of distribution shift. The darker color in the heatmap indicates the larger weight.
    }

	\label{fig:result-shl}
    \end{minipage}
    \vspace{-6mm}
\end{figure}

%% file: sections/conclusion.tex

\section{Conclusion}
\label{sec:conslusion}

This paper proposed algorithms for online label shift with provable guarantees. We constructed an unbiased risk estimator without using any supervision at test time. Then, we designed novel online ensemble algorithms that automatically adapt to the non-stationary online label shift and enjoy problem-dependent dynamic regret. Our proposed \textsc{Atlas} algorithm employed a meta-base structure to handle the non-stationarity and obtained an $\O(V_T^{1/3}T^{2/3})$ guarantee, and \textsc{Atlas-ada} further introduced hint functions to exploit the shift structure and obtained an improved $\O(V_T^{1/3}G_T^{1/3}T^{1/3})$ guarantee. Extensive experiments validated the effectiveness of the proposed algorithms. 

Our study serves as a preliminary attempt to bridge the distribution change problem and online learning techniques by focusing on the label shift case. Considering a more general distribution change setting is an important future direction. Besides, our current algorithm is designed for the most challenging unlabeled scenario, and it is interesting to consider relaxed real-world demand where a few labels could be available in the learning process. Moreover, our obtained regret guarantees hold in expectation, and we will take high-probability bounds as the future work.

%% file: sections/appendices/catlog.tex
\clearpage
\onecolumn
\appendix
\begin{center}
  {\bf \LARGE Supplementary Material for ``Adapting to \\ Online Label Shift with Provable Guarantees''}
\end{center}

This is the supplemental material for the paper ``Adapting to Online Label Shift with Provable Guarantees''. The appendix is organized as follows.
\begin{itemize}
    \item Appendix~\ref{sec-appendix:experiments}: details of the experiments.
    \begin{itemize}
        \item Appendix~\ref{appendix-expr-benchmark}: more numerical results on real-world datasets.
        \item Appendix~\ref{appendix-expr-setup}: details of experimental setup.
        \item Appendix~\ref{appendix-expr-results}: additional experimental results.
    \end{itemize}
    \item Appendix~\ref{sec-appendix:preliminary}: a brief introduction to OCO and BBSE for label shift.
    \begin{itemize}
        \item Appendix~\ref{sec-appendix:preliminary-oco}: review of the OCO framework.
        \item Appendix~\ref{sec-appendix:preliminary-bbse}: details of the BBSE method.
    \end{itemize}
    \item Appendix~\ref{sec-appendix:related-work}: related work and discussion.
    \begin{itemize}
        \item Appendix~\ref{sec-appendix:related-work-label-shift}: related work of the offline label shift problem.
        \item Appendix~\ref{sec-appendix:related-work-oco}: related work of non-stationary online convex optimization.
        \item Appendix~\ref{sec-appendix:difference}: comparison on the risk estimator of our method and the work~\citep{NIPS'21:Online-LS}.
    \end{itemize}
    \item Appendix~\ref{sec-appendix:unbiased-estimator}: omitted proofs for Section~\ref{sec:unbiased-risk-estimator}.
    \begin{itemize}
        \item Appendix~\ref{sec-appendix:proof-lemma1}: proof of Lemma~\ref{lemma:unbiased-estimator}.
        \item Appendix~\ref{sec-appendix:risk-estimator-hp}: high-probability bound of the risk estimator.
        \item Appendix~\ref{sec-appendix:static-regret}: static regret for UOGD algorithm.
    \end{itemize}
    
    \item Appendix~\ref{sec-appendix:proof-sec3.2}: omitted proofs for Section~\ref{sec:online-algorithm}.
    \begin{itemize}
        \item Appendix~\ref{appendix:sec-proof-UOGD}: proof of Theorem~\ref{thm:UOGD}.
        \item Appendix~\ref{appendix:sec-proof-overall-UOGD}: proof of Theorem~\ref{thm:overall-UOGD}.
        \item Appendix~\ref{appendix:discuss-optimal}: discussion on the minimax optimality.
        \item Appendix~\ref{appendix:sec-proof-overall-UOGD-lemmas}: useful lemmas in the analysis of Theorem~\ref{thm:UOGD} and Theorem~\ref{thm:overall-UOGD}.
    \end{itemize}
    \item Appendix~\ref{sec-appendix:proof-sec3.3}: algorithm details for \textsc{Atlas-ada} and omitted proofs for Section~\ref{sec:more-adaptive}.
    \begin{itemize}
        \item Appendix~\ref{sec-appendix:atala-ada-algorithm}: algorithm details for \textsc{Atlas-ada} and the design of the hint function.
        \item Appendix~\ref{appendix:sec-proof-Atlas-ada}: proof of Theorem~\ref{thm:overall-guarantees}.
        \item Appendix~\ref{appendix:sec-proof-Atlas-ada-lemmas}: useful lemmas in the analysis of Theorem~\ref{thm:overall-guarantees}.
    \end{itemize}
    \item Appendix~\ref{sec:appendix-lemma}: technical lemmas.
    
\end{itemize}

%% file: sections/appendices/experiments.tex
\section{Omitted Details for Experiments}
\label{sec-appendix:experiments}
In this section, we mainly supplement the omitted details in Section~\ref{sec:experiments}. We first present the omitted numerical results on benchmark datasets in Appendix~\ref{appendix-expr-benchmark}. Then we list the experiment setups in Appendix~\ref{appendix-expr-setup}, followed by additional experimental results in Appendix~\ref{appendix-expr-results}.

 \subsection{More Numerical Results for Section~\ref{sec:expr-benchmark}}
 \label{appendix-expr-benchmark}

\input{sections/experiments/benchmark_part2.tex}

 Table~\ref{tab:benchmark-part2} presents the numerical results omitted in Section~\ref{sec:expr-benchmark}, which reports the average error of different algorithms on various real-world datasets. The results show that the \textsc{Atlas} and \textsc{Atlas-ada} methods outperform other contenders on all tasks. The advantage is particularly significant when the underlying class prior changes fast. The empirical results validate that our method can effectively adapt to the non-stationary label shift.

 \subsection{Experiment Setup}
 \label{appendix-expr-setup}
 This subsection describes details of experimental setups, including contenders, simulated shifts, and datasets. All our experiments are run on a machine with 2 CPUs (24 cores for each). 

\paragraph{Contenders.}
In the experiments, we mainly compare seven online label shift algorithms, including:
\begin{compactitem}
	\item \emph{FIX} predicts with the fixed initialized classifier without any online updates.
    \item \emph{ROGD} is a variant of OGD algorithm proposed by~\citet{NIPS'21:Online-LS}. The algorithm constructs the risk estimator with the $0/1$ loss and uses a re-weighting classifier for the model update. (Detailed description and the comparison with our risk estimator can be found in Appendix~\ref{sec-appendix:difference}).
    \item \emph{FTH} is short for Following The History algorithm proposed by~\citet{NIPS'21:Online-LS}. The method takes the class prior for every iteration $t$ as the average of all previously estimated priors.
    \item \emph{FTFWH} is a variant of FTH proposed as a compared method in experiments of~\citet{NIPS'21:Online-LS}. The method takes an average across previously estimated priors within a sliding window. In all experiments, the length of the sliding window is set as $100$.
	\item \emph{UOGD}/\textsc{Atlas}/\textsc{Atlas-ada}: The \ols algorithms proposed in Section~\ref{sec:approach}.
\end{compactitem}

Following our theoretical results, the step size of \emph{ROGD} and \emph{UOGD} is set to be $\nicefrac{\Gamma}{\left(G\sqrt{T}\right)}$, where $G$ can be estimated during training the initial model and $\Gamma$ is given by the decision set domain. In the experiments, we choose the decision domain as a ball with a fixed diameter for each dataset. For a fair comparison, all contenders use the same decision domain $\mathcal{W}$ with the diameter $\Gamma$ set according to the parameter norm of the initial offline model $f_0$. The settings of step size pool of \textsc{Atlas} and \textsc{Atlas-ada} are guided by Theorem~\ref{thm:overall-UOGD} and Theorem~\ref{thm:overall-guarantees}, respectively. We employ a multinomial logistic regression classifier for \emph{UOGD}, \textsc{Atlas}, and \textsc{Atlas-ada}.

\begin{table}[ht!]
    \centering
    \begin{tabular}{ll}
     \texttt{Lin} & $\raisebox{-.5\height}{\includegraphics[width=0.9\textwidth]{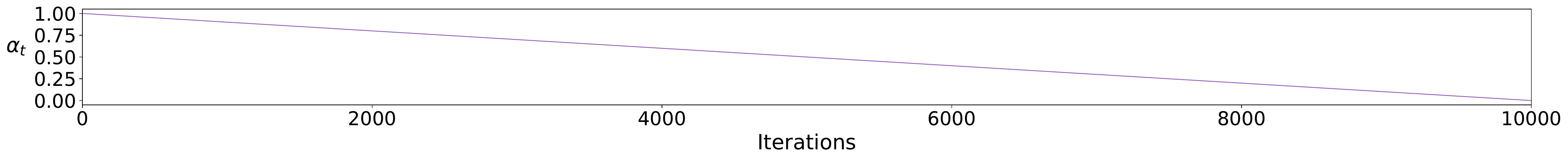}}$ \\
     \texttt{Squ} & $\raisebox{-.5\height}{\includegraphics[width=0.9\textwidth]{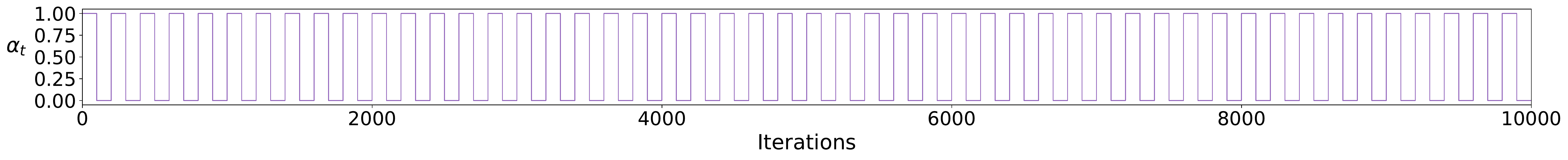}}$ \\
     \texttt{Sin} & $\raisebox{-.5\height}{\includegraphics[width=0.9\textwidth]{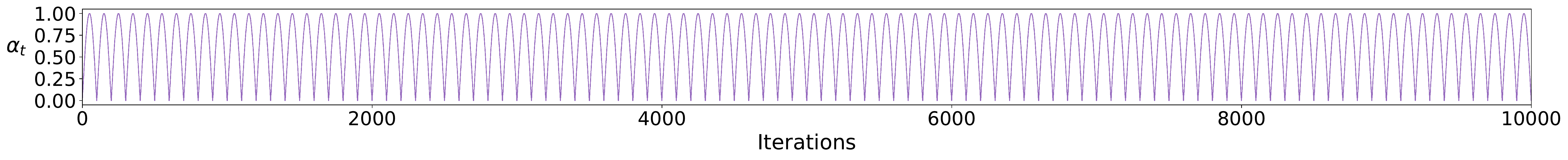}}$ \\
     \texttt{Ber} & $\raisebox{-.5\height}{\includegraphics[width=0.9\textwidth]{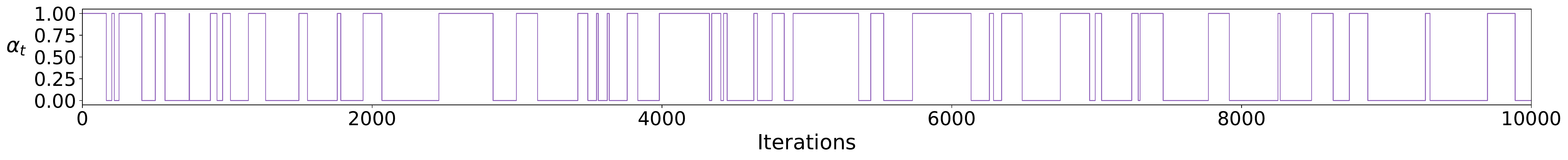}}$
    \end{tabular}
    \vspace{3mm}
    \captionof{figure}{From top to bottom: \texttt{Linear Shift}, \texttt{Square Shift}, \texttt{Sine Shift}, \texttt{Bernoulli Shift}.}
	\label{fig:shift-sketch}
	\vspace{-4mm}
\end{table}

\paragraph{Simulated Shifts.}
We simulate four kinds of label shift patterns to capture different kinds of non-stationary environments. For each shift, the priors are a mixture of two different constant priors $\bm{\mu}_{1}, \bm{\mu}_{2} \in \Delta_K$ with a time-varying coefficient $\alpha_t$: $\bm{\mu}_{y_t} = (1 - \alpha_t) \bm{\mu}_{1} + \alpha_t  \bm{\mu}_{2}$, where $\bm{\mu}_{y_t}$ denotes the label distribution at round $t$ and $\alpha_t$ controls the shift non-stationarity and patterns. We list the details in the following.

\begin{compactitem}
    \item \textbf{\texttt{Linear Shift} (\texttt{Lin})}: the parameter $\alpha_t = \frac{t}{T}$, which represents the gradually changed environments. 
    \item \textbf{\texttt{Square Shift} (\texttt{Squ})}: the parameter $\alpha_t$ switches between $1$ and $0$ every $L/2$ rounds, where $L$ is the periodic length. In the experiments, we set $L = \Theta(\sqrt{T})$ by default, which implies that the fluctuation of the class prior $V_T = \sum_{t=2}^T\norm{\bm{\mu}_{y_t}-\bm{\mu}_{{y}_{t-1}}}_1$ is $\Theta(\sqrt{T})$. The \texttt{Square Shift} simulates a fast-changing environment with periodic patterns.
    
    \item \textbf{\texttt{Sine Shift} (\texttt{Sin})}: $\alpha_t = \sin \frac{i\pi}{L}$, where $i=t\bmod L$ and $L$ is a given periodic length. In the experiments, we set $L = \Theta(\sqrt{T})$ by default. The \texttt{Sine Shift} also simulates a fast-changing environment with periodic patterns.

    \item \textbf{\texttt{Bernoulli Shift} (\texttt{Ber})}: At every iteration, we keep the $\alpha_t = \alpha_{t-1}$ with probability $p\in[0,1]$ and otherwise set $\alpha_t = 1 - \alpha_{t-1}$. In the experiments, the parameter is set as $p = 1/\sqrt{T}$ by default, which implies $V_t = \Theta(\sqrt{T})$. The \texttt{Bernoulli Shift}  simulates a fast-changing environment without periodic patterns.
\end{compactitem}

Figure~\ref{fig:shift-sketch} demonstrates how $\alpha_t$ changes over time. We can observe that $\texttt{Square Shift}$ and $\texttt{Sine Shift}$ change in a periodic pattern while the others do not. Moreover, it can be seen that $\texttt{Linear Shift}$ has simulated a much more tender class prior change than the other three shifts.

\paragraph{Datasets.} 
We conduct experiments on synthetic data, six real-world datasets, and a real-life application of locomotion detection.

\emph{{Synthetic data}.} There are three classes in the synthetic data, where the feature distribution of each class follows a Gaussian distribution. More specifically, for each instance $(\x_i, y_i)$ in the dataset $S_t$ at round $t$, the label is generated from the discrete distribution defined by the given priors $\D_t(y)$, and the feature $\x \in \R^{12}$ is generated from the corresponding multivariate normal distributions $\mathcal{N}(\mu_{y}, \Sigma)$. 
We set $\bm{\mu}_1 = [1/3,1/3,1/3]$ and $\bm{\mu}_2 = [1,0,0]$ for simulated shifts. 

{\emph{Real-world datasets}.} We conduct experiments on six real-world datasets. For each dataset, we set $\bm{\mu}_1 = [1/K, 1/K, ..., 1/K]$ and $\bm{\mu}_2 = [1, 0, ..., 0]$ to generate the simulated shifts. Specifically, we include the following datasets.
\begin{compactitem}
    \item \textbf{ArXiv~\footnote{\url{www.kaggle.com/datasets/Cornell-University/arxiv}}:} A paper classification dataset which contains metadata of the scholarly articles. We select 296,\ 708 papers from the computer science domain, which cover 23 classes, including cs.AI, cs.CC, cs.CL, cs.CR, cs.CV, cs.CY, cs.DB, cs.DC, cs.DM, cs.DS, cs.GT, cs.HC, cs.IR, cs.IT, cs.LG, cs.LO, cs.NE, cs.NI, cs.PL, cs.RO, cs.SE, cs.SI, and cs.SY.~\footnote{See \url{www.arxiv.org/archive/cs} for the full name.} We use a finetuned DistilBERT~\citep{journals/corr/abs-1910-01108} to extract features from authors, titles and abstracts. The papers selected additionally to finetune the DistilBERT do not overlap with the offline and online datasets.
    \item \textbf{EuroSAT}~\citep{DBLP:journals/staeors/HelberBDB19}: A land cover classification dataset, which includes satellite images with the purpose of identifying the visible land use or land cover class. The dataset consists of 27, 000 satellite images from over 30 different European countries. It contains ten different classes, including industrial, residential, annual crop, permanent crop, river, sea and lake, herbaceous vegetation, highway, pasture, and forest. We use a finetuned ResNet~\citep{DBLP:conf/cvpr/HeZRS16} to extract features from the images of EuroSAT and the following four datasets. The images selected to train the ResNet also do not overlap with both the offline or online datasets.
    \item \textbf{MNIST}~\citep{DBLP:journals/pieee/LeCunBBH98}: A widely-used image dataset of handwritten digits, which consists of 70,\ 000 grayscale images with 10 different classes.
    \item \textbf{Fashion}~\citep{DBLP:journals/corr/abs-1708-07747}: A dataset of 70,\ 000 grayscale fashion images, consisting of 10 different classes: T-shirt, trouser, shirt and sneaker, pullover, dress, coat, sandal, bag, and ankle boot.
    \item \textbf{CIFAR10}~\citep{Krizhevsky09learningmultiple}: A dataset consists of 60,\ 000 color images in 10 classes, including airplane, automobile, ship, truck, bird, cat, deer, dog, frog, and horse.
    \item \textbf{CINIC10}~\citep{DBLP:journals/corr/abs-1810-03505}: A tiny ImageNet~\citep{deng2009imagenet} dataset, which consists of images from CIFAR10 and ImageNet, and has the same ten classes as CIFAR10.
\end{compactitem}

\emph{Real-life application.} The real-life application is to distinguish human locomotion through the sensor data collected by the carry-on mobile phones\footnote{\url{www.shl-dataset.org}}. The tabular data covers the sensor data (e.g., acceleration, gyroscope, magnetometer, orientation, gravity, pressure, altitude, and temperature) and the corresponding human motion and timestamp. We sample 30,\ 000 offline data and 77,\ 000 online data from 11 days, covering six classes, including still, walking, run, bike, car, and bus. During the online update, the online samples arrive in real chronological order based on the timestamp.

\subsection{Additional Experimental Results}
\label{appendix-expr-results}
This part further reports the additional experimental results omitted in the main paper. Specifically, we show a supplement for the assigned weight visualization of \textsc{Atlas} and a skyline to illustrate how the meta-algorithm tracks the proper base-algorithms.

\textbf{Weight Assignment of \textsc{Atlas}.} In Figure~\ref{fig-meta-weights}, we show that slower \texttt{Linear Shift} is assigned relatively small step sizes, while faster \texttt{Bernoulli Shift} is assigned relatively large step sizes. Here, we replenish the experimental results for \texttt{Square Shift} and \texttt{Sine Shift}. As these two simulated shifts also change relatively fast like \texttt{Bernoulli Shift}, the assigned weights are also expected to be relatively large, which is in line with the experimental results in Figure~\ref{tab:new-syn-results-c} and~\ref{tab:new-syn-results-d}.

\input{sections/experiments/meta-weights_part2.tex}

\begin{wrapfigure}{r}{0.45\columnwidth}
    \vspace{-4mm}
    \centering
    \begin{minipage}[b]{0.45\columnwidth}
        \includegraphics[width=\textwidth]{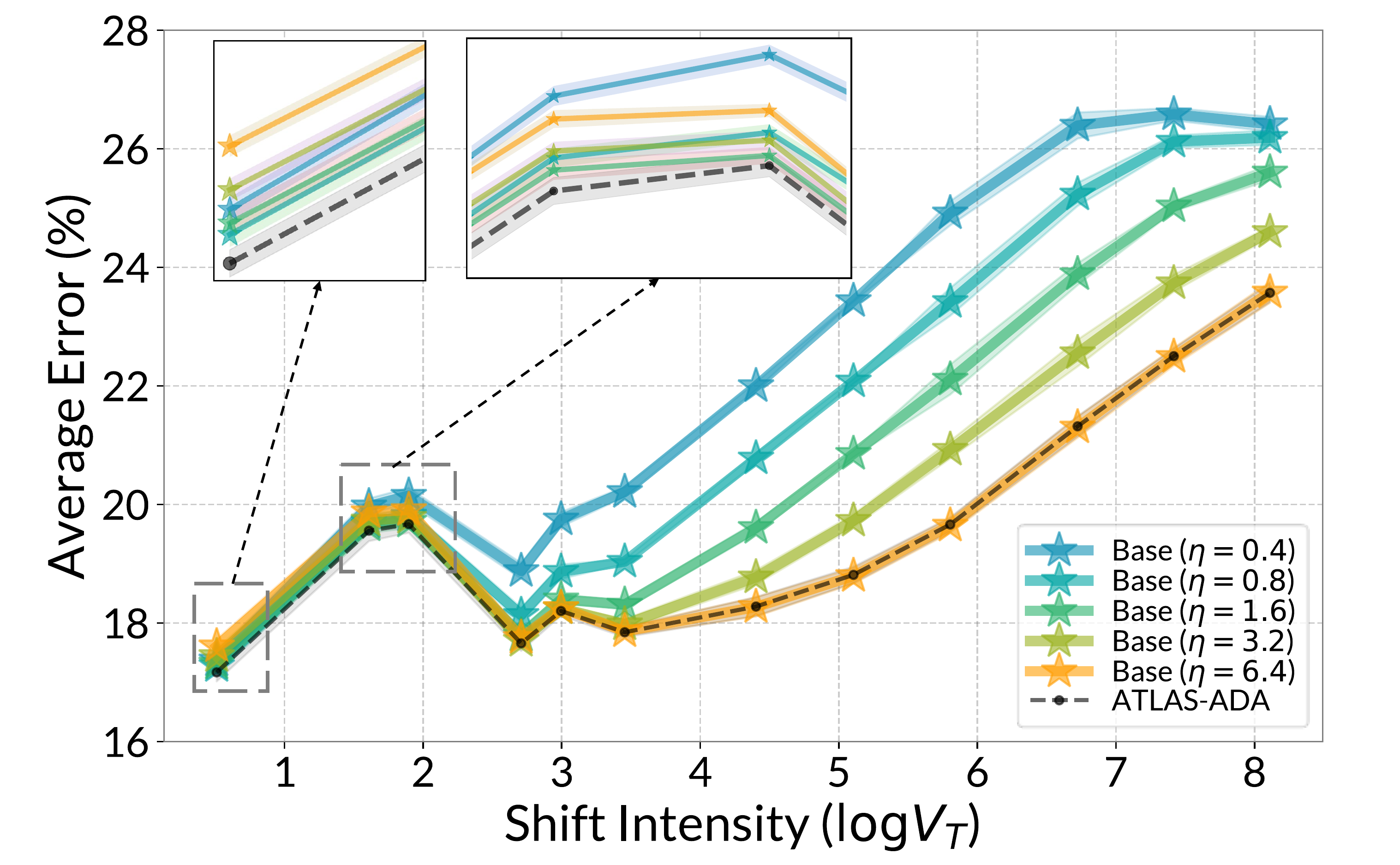}
    \end{minipage}
    \vspace{-3mm}
    \caption{Skyline on the SHL dataset. The intensity of the label distribution shift increases from left to right. The black dashed line represents the average error of \textsc{Atlas-ada}, while the other lines represent that of different base-learners.}
    \vspace{-3mm}
	\label{fig:result-shl-skyline}
\end{wrapfigure}

\textbf{Skyline of \textsc{Atlas-ada}.} The purpose of this part is to answer the question of whether our proposed \textsc{Atlas-ada} can cope with online label shifts, regardless of how fast or slow the change is. To see this, we use the SHL dataset to simulate shifts with different non-stationarities, increasing from left to right side. We plot the final average errors of the major base-learners and \textsc{Atlas-ada} in Figure~\ref{fig:result-shl-skyline}.  Note that base-learners with much smaller or larger step sizes lead to worse results for all situations and are omitted for clarity. As shown in Figure~\ref{fig:result-shl-skyline}, small step sizes achieve the best average errors for tender non-stationarity cases (left side) but fail for dramatic cases (right side), which is consistent with the theoretical results that the optimal step size scales with the non-stationarity measure $V_T$. Moreover, we find that the proposed \textsc{Atlas-ada} can always track the best base-learners, no matter tender cases or dramatic cases, which implies that we can safely employ it without knowing the change of the unknown environment in advance.

%% file: sections/experiments/benchmark_part2.tex
\begin{table*}[!t]
  \centering
  \caption{Average error (\%) of different algorithms on various real-world datasets (\texttt{Sin} and \texttt{Squ}).  We report the mean and standard deviation over five runs. The best algorithms are emphasized in bold. ``$\bullet$'' indicates the algorithms that are significantly inferior to \textsc{Atlas-ada} by the paired $t$-test at a $5\%$ significance level. Here AT-ADA represents \textsc{Atlas-ada} (with \emph{OKM}). The online sample size is set as $N_t = 10$. }
  \resizebox{\textwidth}{!}{%
  \begin{tabular}{crrrrrrrrrrrrrr}
    \toprule
    &
         \multicolumn{7}{c}{\texttt{Sin}} &
         \multicolumn{7}{c}{\texttt{Squ}}
           \\ \cmidrule(lr){2-8} \cmidrule(lr){9-15}
         & 
         \multicolumn{1}{c}{\textbf{FIX}} &
          \multicolumn{1}{c}{\textbf{FTH}} &
          \multicolumn{1}{c}{\textbf{FTFWH}} &
          \multicolumn{1}{c}{\textbf{ROGD}} &
          \multicolumn{1}{c}{\textbf{UOGD}} &
          \multicolumn{1}{c}{\textbf{ATLAS}} &
          \multicolumn{1}{c}{\textbf{AT-ADA}} &
          \multicolumn{1}{c}{\textbf{FIX}} &
          \multicolumn{1}{c}{\textbf{FTH}} &
          \multicolumn{1}{c}{\textbf{FTFWH}} &
          \multicolumn{1}{c}{\textbf{ROGD}} &
          \multicolumn{1}{c}{\textbf{UOGD}} &
          \multicolumn{1}{c}{\textbf{ATLAS}} &
          \multicolumn{1}{c}{\textbf{AT-ADA}}
           \\ \midrule
    \multirow{2}{*}{\textbf{ArXiv}} &
    $\bullet$ 31.58 &
    $\bullet$ 30.63 &
    $\bullet$ 31.90 &
    $\bullet$ 28.35 &
    $\bullet$ 25.64 &
    $\bullet$ 26.03 &
    \textbf{25.08} &
      $\bullet$ 30.35 &
      $\bullet$ 26.72 &
      $\bullet$ 28.05 &
      $\bullet$ 24.44 &
      $\bullet$ 21.96 &
      $\bullet$ 21.36 &
      \textbf{20.80} \\
     &
     $\pm$0.10 &
     $\pm$0.24 &
     $\pm$0.22 &
     $\pm$0.32 &
     $\pm$0.18 &
     $\pm$0.16 &
     $\pm$0.12 &
      $\pm$0.06 &
      $\pm$0.39 &
      $\pm$0.20 &
      $\pm$0.17 &
      $\pm$0.07 &
      $\pm$0.06 &
      $\pm$0.06 \\
    \multirow{2}{*}{\textbf{EuroSAT}} &
    $\bullet$ 13.62 &
    $\bullet$ 10.90 &
    $\bullet$ 10.96 &
    $\bullet$ {\color{white} 0}9.68 &
    $\bullet$ {\color{white} 0}8.03 &
    $\bullet$ {\color{white} 0}8.03 &
    \textbf{7.97} &
      $\bullet$ 14.15 &
      $\bullet$ 10.22 &
      $\bullet$ 10.26 &
      $\bullet$ {\color{white} 0}8.91 &
     $\bullet$ {\color{white} 0}7.30 &
      $\bullet$ {\color{white} 0}6.97 &
      \textbf{6.81} \\
     &
     $\pm$0.13 &
      $\pm$0.03 &
      $\pm$0.02 &
      $\pm$0.08 &
      $\pm$0.06 &
      $\pm$0.06 &
      $\pm$0.08 &
      $\pm$0.11 &
      $\pm$0.08 &
      $\pm$0.06 &
      $\pm$0.05 &
      $\pm$0.07 &
      $\pm$0.08 &
      $\pm$0.06 \\
    \multirow{2}{*}{\textbf{MNIST}} &
    $\bullet$ {\color{white} 0}1.81 &
    $\bullet$ {\color{white} 0}1.46 &
    $\bullet$ {\color{white} 0}1.47 &
    $\bullet$ {\color{white} 0}1.46 &
    $\bullet$ {\color{white} 0}1.30 &
    1.28 &
    \textbf{1.27} &
      $\bullet$ {\color{white} 0}1.79 &
      $\bullet$ {\color{white} 0}1.26 &
      $\bullet$ {\color{white} 0}1.28 &
      $\bullet$ {\color{white} 0}1.32 &
      $\bullet$ {\color{white} 0}1.13 &
      $\bullet$ {\color{white} 0}1.04 &
      \textbf{1.01} \\
     &
     $\pm$0.02 &
     $\pm$0.03 &
     $\pm$0.03 &
     $\pm$0.03 &
     $\pm$0.04 &
     $\pm$0.03 &
     $\pm$0.03 &
      $\pm$0.04 &
      $\pm$0.03 &
      $\pm$0.04 &
      $\pm$0.04 &
      $\pm$0.03 &
      $\pm$0.02 &
      $\pm$0.04 \\
    \multirow{2}{*}{\textbf{Fashion}} &
    $\bullet$ 11.77 &
      9.37 &
      9.39 &
      $\bullet$ {\color{white} 0}9.75 &
      9.36 &
      $\bullet$ {\color{white} 0}9.44 &
      \textbf{9.32} &
      $\bullet$ 11.92 &
      $\bullet$ {\color{white} 0}8.24 &
      $\bullet$ {\color{white} 0}8.35 &
      $\bullet$ {\color{white} 0}8.63 &
      $\bullet$ {\color{white} 0}8.42 &
      $\bullet$ {\color{white} 0}8.05 &
      \textbf{7.73} \\
     &
     $\pm$0.11 &
     $\pm$0.15 &
     $\pm$0.14 &
     $\pm$0.12 &
     $\pm$0.07 &
     $\pm$0.04 &
     $\pm$0.04 &
      $\pm$0.09 &
      $\pm$0.09 &
      $\pm$0.07 &
      $\pm$0.07 &
      $\pm$0.04 &
      $\pm$0.07 &
      $\pm$0.05 \\
    \multirow{2}{*}{\textbf{CIFAR10}} &
    $\bullet$ 21.40 &
    $\bullet$ 18.57 &
    $\bullet$ 18.62 &
    $\bullet$ 19.16 &
    $\bullet$ 18.17 &
    $\bullet$ 18.01 &
    \textbf{17.89} &
      $\bullet$ 20.77 &
      $\bullet$ 16.67 &
      $\bullet$ 16.72 &
      $\bullet$ 17.40 &
      $\bullet$ 16.29 &
      $\bullet$ 15.18 &
      \textbf{14.84} \\
     &
     $\pm$0.09 &
     $\pm$0.07 &
     $\pm$0.08 &
     $\pm$0.12 &
     $\pm$0.07 &
     $\pm$0.07 &
     $\pm$0.05 &
      $\pm$0.08 &
      $\pm$0.12 &
      $\pm$0.12 &
      $\pm$0.11 &
      $\pm$0.09 &
      $\pm$0.07 &
      $\pm$0.05 \\
    \multirow{2}{*}{\textbf{CINIC10}} &
    $\bullet$ 35.29 &
      $\bullet$ 31.17 &
      $\bullet$ 31.20 &
      $\bullet$ 31.46 &
      $\bullet$ 30.22 &
      $\bullet$ 30.15 &
      \textbf{30.06} &
      $\bullet$ 33.99 &
      $\bullet$ 27.99 &
      $\bullet$ 28.08 &
      $\bullet$ 28.58 &
      $\bullet$ 27.00 &
      $\bullet$ 25.94 &
      \textbf{25.56} \\
     &
     $\pm$0.19 &
     $\pm$0.12 &
     $\pm$0.12 &
     $\pm$0.14 &
     $\pm$0.10 &
     $\pm$0.11 &
     $\pm$0.15 &
      $\pm$0.16 &
      $\pm$0.09 &
      $\pm$0.08 &
      $\pm$0.09 &
      $\pm$0.14 &
      $\pm$0.13 &
      $\pm$0.12 \\ \bottomrule
    \end{tabular}
  }
  
  \label{tab:benchmark-part2}
  \end{table*}

%% file: sections/experiments/meta-weights_part2.tex
\begin{figure}[!t]  
    \centering 
        \begin{minipage}[b]{.6\textwidth}
            \centering
            \includegraphics[width=\textwidth]{figures/meta_weights/legend.pdf} 
       \end{minipage}
       \vfill
       \vspace{-2mm}
       \subfigure[\texttt{Linear shift}]{
            \begin{minipage}[b]{0.22\textwidth}
               \label{tab:new-syn-results-a}
               \includegraphics[width=\textwidth, height=.575\textwidth]{figures/meta_weights/lin_shift.pdf} 
            \end{minipage}
            }
       \subfigure[\texttt{Square shift}]{
       \begin{minipage}[b]{0.22\textwidth}
        \label{tab:new-syn-results-c} 
          \includegraphics[width=\textwidth, height=.575\textwidth]{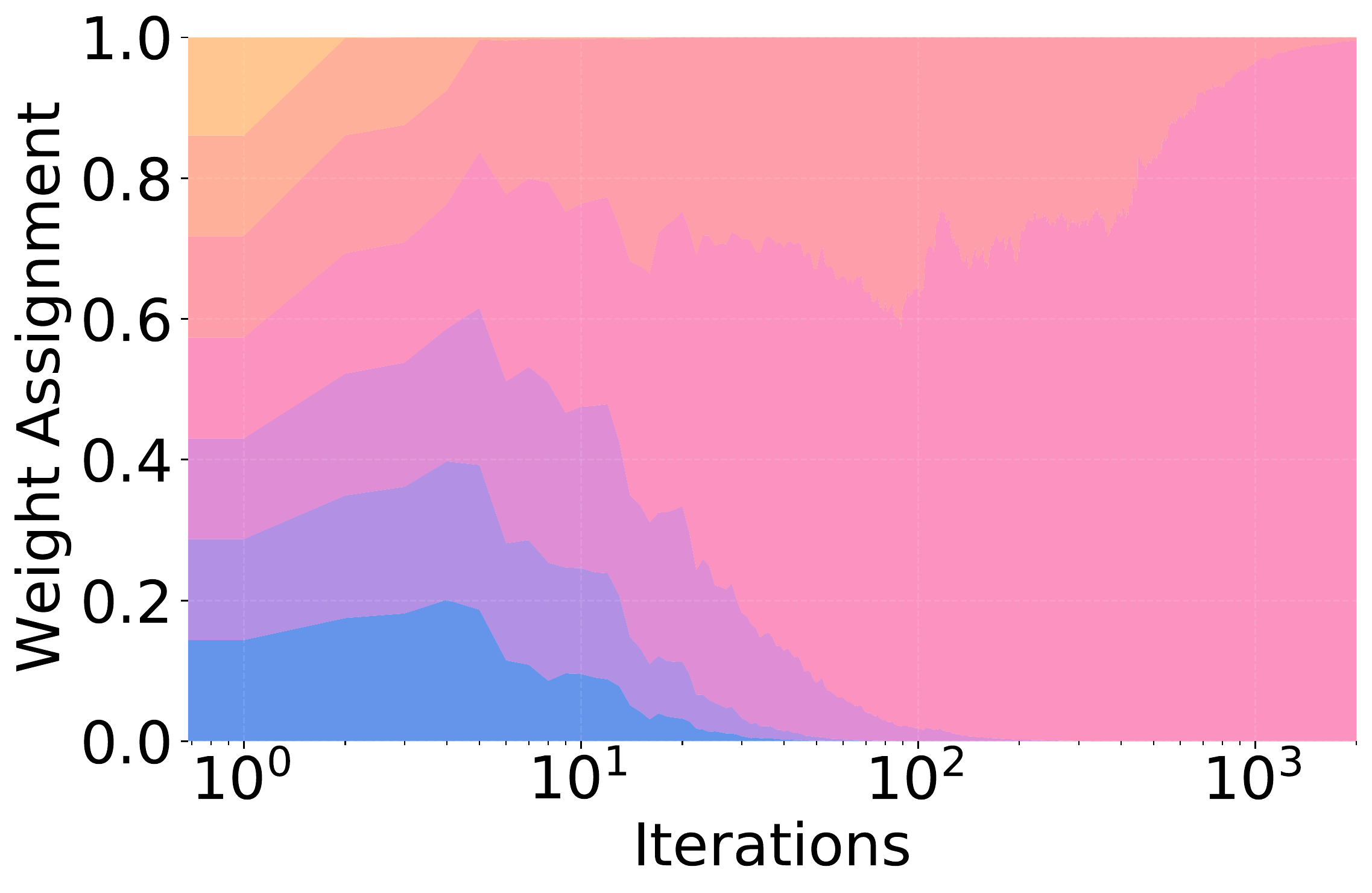} 
       \end{minipage}
       }
       \subfigure[\texttt{Sine shift}]{
       \begin{minipage}[b]{0.22\textwidth}
        \label{tab:new-syn-results-d} 
          \includegraphics[width=\textwidth,  height=.575\textwidth]{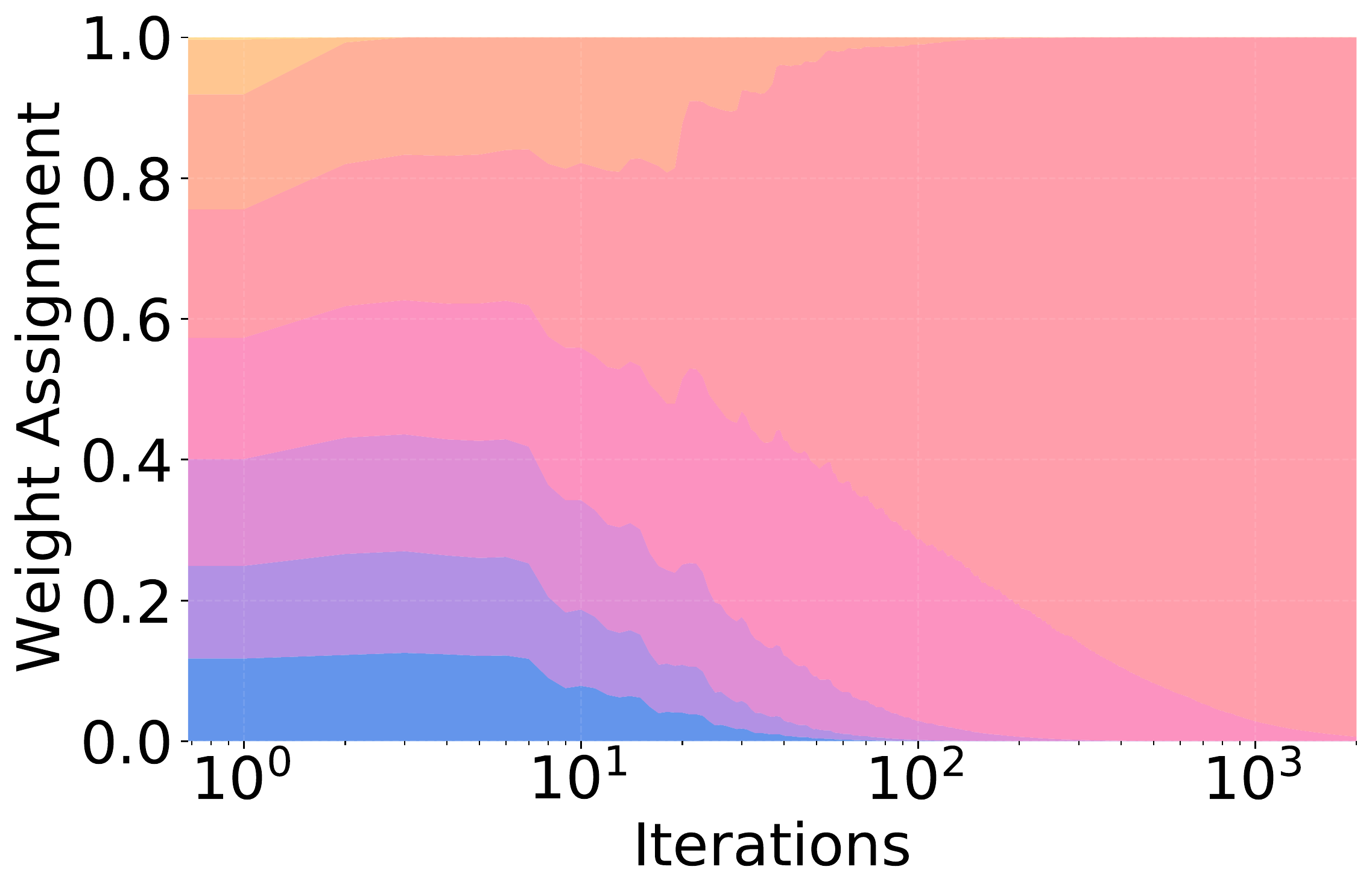}  
       \end{minipage}
       }
       \subfigure[\texttt{Bernoulli shift}]{
            \begin{minipage}[b]{0.22\textwidth}
               \label{tab:new-syn-results-b}
               \includegraphics[width=\textwidth,  height=.575\textwidth]{figures/meta_weights/ber_shift.pdf}  
            \end{minipage}
            }
        \caption{Weight assigned of the \textsc{Atlas} algorithm for each step size along the learning process. Different colors are used to indicate different step sizes.}
        \vspace{-3mm}
        \label{fig-meta-weights-part2}
 \end{figure}

%% file: sections/appendices/preliminary.tex

\section{Preliminaries}
\label{sec-appendix:preliminary}
In this section, we first provide a brief introduction to the online convex optimization framework in Appendix~\ref{sec-appendix:preliminary-oco}, and then in Appendix~\ref{sec-appendix:preliminary-bbse} we present a detailed description of the BBSE method that is designed for the offline label shift problem, as well as its usage in constructing our risk estimator for the online label shift problem.

\subsection{Brief Introduction to Online Convex Optimization}
\label{sec-appendix:preliminary-oco}
This section briefly introduces the online convex optimization framework. In many real-world tasks, data are often accumulated in a sequential way, which requires a learning paradigm to update the model in an online fashion. Taking the locomotion detection task as an example, each time, we receive a few data collected by smartphone sensors and need to predict the motion type immediately. The sequential nature of the data makes it challenging to train a model offline using the standard machine learning methods. These circumstances can be modeled and handled with the Online Convex Optimization (OCO) framework~\citep{ICML'03:zinkvich,book'16:Hazan-OCO, book'12:Shai-OCO}.

The OCO framework can be viewed as a structured repeated game between a learner and the environment. During the learning process, the learner iteratively chooses decisions from a fixed convex decision set $\W$ based on the feedback from the environments. Let $T$ denote the total number of game iterations, and the protocol of OCO is given as follows.

\begin{tcolorbox}
\centerline{\textsc{Online Convex Optimization Framework}}
 \begin{algorithmic}[1]
     \FOR{$t=1$ {\bfseries to} $T$}
       \STATE the learner chooses a decision $\w_t \in \W$, and the environments simultaneously choose a convex loss function $R_t:\W\mapsto\R$;
       \STATE the learner suffers cost $R_t(\w_t)$ and observes certain information about $R_t$.
     \ENDFOR
 \end{algorithmic}
 \end{tcolorbox}
 
  The classic performance measure for the OCO algorithms is the (static) regret defined as $$\SReg = \sum_{t=1}^T R_t(\w_t)-\sum_{t=1}^TR_t(\w_*),$$ which compares the learner's prediction with the best single decision in hindsight $\w_* \in \argmin_{\w\in\W}\sum_{t=1}^TR_t(\w)$. The static regret is a reasonable measure for stationary environments. But the performance measure could be too optimistic in non-stationary environments, where the underlying distribution changes over time, and the single best model could perform badly. Under such a case, a more suitable performance is the \emph{dynamic regret} that competes with the online learner's performance with the best decisions for each round. A more detailed discussion and literature review for the dynamic regret can be found in Appendix~\ref{sec-appendix:related-work-oco}.

  The OCO framework has received extensive studies over the decades. Among them, one of the most prominent algorithms is the Online Gradient Descent (OGD) algorithm~\citep{ICML'03:zinkvich}, which takes a step in the opposite direction to the gradient of the previous risk at each iteration (See Algorithm~\ref{alg:ogd} for the procedures). Despite its simplicity, OGD algorithm is powerful enough to handle a large family of online problems. Specifically, it enjoys a static regret of $\O(\sqrt{T})$ by setting the step size to be $\Theta(1/\sqrt{T})$, which was proven to be minimax optimal~\citep{DBLP:conf/colt/AbernethyBRT08} for convex functions. Latter, tighter static bounds for loss functions with stronger curvature~\citep{journals/ml/HazanAK07} and that can adapt to benign environments~\citep{COLT'08:Hazan-variation,COLT'12:variation-Yang,conf/colt/RakhlinS13} were proposed. For more detailed reviews of the OCO framework, we refer the readers to the seminal books~\citep{book'12:Shai-OCO, book'16:Hazan-OCO}.
  
  \begin{algorithm}[!t]
    \caption{Online Gradient Descent (OGD)}
    \label{alg:ogd}
 \begin{algorithmic}[1]
   \REQUIRE{step size $\eta > 0$}
   \STATE{Initialization: let $\w_{1,i}$ be any point in $\W$.}
     \FOR{$t=1$ {\bfseries to} $T$}
       \STATE the learner plays $\w_t$ and observes loss $R_t(\w_t)$ and gradient $\nabla R_t(\w_t)$;
       \STATE gradient-descent update: $$\w_{t+1} = \Pi_{\W}\left[\w_t - \eta \nabla R_t(\w_t)\right].$$
     \ENDFOR
 \end{algorithmic}
 \end{algorithm}

\subsection{BBSE Method for Online Label Shift}
\label{sec-appendix:preliminary-bbse}

The Black Box Shift Estimation (BBSE) method~\citep{conf/icml/LiptonWS18} is a family of label shift algorithms that uses the confusion matrix of a given black-box classifier to estimate the label distributions from unlabeled samples. This part illustrates how to employ BBSE to 
the online label distribution estimation.

Recall that we have the labeled offline data $S_0$ and the unlabeled data $S_t$, and we need to estimate the label distribution $\bm{\mu}_{y_t}$. Note that $\bm{\mu}_{y_t}$ cannot be estimated directly due to labels of $S_t$ being unavailable. Since $S_0$ is labeled, a possible way is to employ the labels of $S_0$. To this end, we introduce an initial black-box model $f_0$ and have the following equation:
\begin{align}
    \E_{\x \sim \D_t(\x)} \left[\indicator\{f_0(\x)=i\}\right] & 
    = \sum_{j=1}^{K} [\bm{\mu}_{y_t}]_j \cdot \E_{\x \sim \D_t\left(\x \givenn y=j \right)} \left[\indicator\{f_0(\x)=i\}\right]\notag\\
    & = \sum_{j=1}^{K} [\bm{\mu}_{y_t}]_j \cdot \E_{\x \sim \D_0\left(\x \givenn y=j \right)} \left[\indicator\{f_0(\x)=i\}\right] \label{eq:bbse},
\end{align}
which holds for all $i\in [K]$. The first equality holds due to the law of total probability and the second equality is based on Assumption~\ref{assum:label-shift} that $\D_t\left(\x \givenn y \right) = \D_0\left(\x \givenn y \right)$ for any $t\in[T]$.~\eqref{eq:bbse} can be equivalently expressed in matrix notation, which is given by: 
\begin{equation}
    \bm{\mu}_{\hat{y}_t} = C_{f_0} \bm{\mu}_{y_t},\label{eq:bbse-matrix}
\end{equation}
where $\bm{\mu}_{\hat{y}_t}\in\Delta_K$ is the distribution vector of the black-box model's prediction under the distribution $\D_t(\x)$, with the $i$-th entry $[\bm{\mu}_{\hat{y}_t}]_i = \E_{\x \sim \D_t(\x)} \left[\indicator\{f_0(\x)=i\}\right]$.  $C_{f_0}\in \R^{K\times K}$ is the confusion matrix, whose $(i,j)$-th entry $[C_{f_0}]_{ij} = \E_{\x\sim\D_0 \left(\x \givenn y=j \right)}[\indicator\{f_0(\x) = i\}]$ is the classification rate that the initial model $f_0$ predicts samples from class $i$ as class $j$. Here we assume the confusion matrix $C_{f_0}$ is invertible.
By solving~\eqref{eq:bbse-matrix}, we have
\begin{equation}
    \bm{\mu}_{y_t} = C_{f_0}^{-1} \bm{\mu}_{\hat{y}_t}.\label{eq:bbse-matrix-solve}
\end{equation}
The LHS of~\eqref{eq:bbse-matrix-solve} $\bm{\mu}_{y_t}$ is the label distribution we want. In the RHS, we note that the distribution $\D_t(\x)$ and $\D_0 \left(\x \givenn y \right)$ is empirically accessible via the unlabeled data $S_t$ and the labeled data $S_0$, so both $C_{f_0}$ and $\bm{\mu}_{\hat{y}_t}$ can be unbiasedly estimated with its empirical version.
Specifically, $C_{f_0}$ can be estimated with the offline labeled data $S_0$:  
\begin{equation}
\label{eq:definition-C}
    [\hat{C}_{f_0}]_{ij} = \sum_{(\x, y)\in S_0} \frac{  \indicator\{f_0\left(\x\right)=i \text { and } y=j\}}{\indicator\{y=j\}}.
\end{equation}
And $\bm{\mu}_{\hat{y}_t}$ in the RHS can be estimated with the online data $S_t$, which is given by 
\begin{equation}
    \label{eq:definition-hat-u-hat-y}
[\hat{\bm{\mu}}_{\hat{y}_t}]_j = \frac{1}{\vert S_t\vert}\sum_{\x\in S_t}{\indicator\{f_0(\x) = j\}}.
\end{equation}
With the above estimation, we construct the final estimator for the label distribution vector:
$$
\hat{\bm{\mu}}_{y_t} =\hat{C}_{f_0}^{-1} \hat{\bm{\mu}}_{\hat{y}_t}.
$$

Note that although BBSE has benign unbiasedness, it has limitations arising from the inverse operation, which might amplify slight errors in the confusion matrix.  Other powerful unbiased estimators can also be applied to our \ols method directly.

%% file: sections/appendices/related_work.tex

\section{Related Work and Discussion}
\label{sec-appendix:related-work}

This section introduces related works to our paper, including supervised learning with label shift in Appendix~\ref{sec-appendix:related-work-label-shift}, and online learning measures for non-stationary environments in Appendix~\ref{sec-appendix:related-work-oco}. Furthermore, in Appendix~\ref{sec-appendix:difference}, we clarify the main difference between~\citet{NIPS'21:Online-LS} and our work.

\subsection{Supervised Learning with Label Shift}
\label{sec-appendix:related-work-label-shift}
Label shift is a typical scenario of learning with distribution change problem~\citep{edit:Quinonero-Candela+etal:2009}. Most existing works have focused on the offline setting, where the label distribution varies from the source to target stages, but the label-conditional density remains the same. The main challenge of offline label shift problem lies in the estimation of the target label distribution. With such knowledge, the learner can train classifiers guaranteed to perform well over the target distribution.~\citet{DBLP:journals/neco/SaerensLD02} first introduced two kinds of solutions for label distribution estimation, including the confusion matrix-based method and the maximum likelihood estimation.~\citet{conf/icml/LiptonWS18} provided a convergence analysis of the matrix-based method with black box models and~\citet{conf/iclr/Azizzadenesheli19} enhanced the matrix-based method by regularization. Another line of studies investigates the label distribution estimation problem via distribution matching~\citep{conf/icml/ZhangSMW13, DBLP:journals/nn/PlessisS14, DBLP:conf/acml/NguyenPS15,DBLP:conf/nips/GargWBL20}.~\citet{DBLP:conf/nips/GargWBL20} further showed that the matrix-based method can be interpreted from the distribution matching view. Beyond the label shift, the label distribution estimation with labeled data from the source domain and unlabeled data from the target domain is also known as the mixture proportion estimation problem~\citep{journals/jmlr/BlanchardLS10,MLJ'17:class-prior,conf/icml/RamaswamyST16}, which has been wildly applied in learning with noisy label~\citep{conf/aistats/Scott15}, positive and unlabeled learning~\citep{conf/nips/PlessisNS14,conf/icml/PlessisNS15}, and learning with unknown classes~\citep{NeurIPS'20:EULAC}, etc.

The above methods only considered the scenario where the label distribution changes once, and sufficient unlabeled samples from the target distribution can be collected in advance. In many real-world applications, data are accumulated with time, and thus it is important to consider the online version of the label shift problem.~\citet{NIPS'21:Online-LS} first introduced online label shift and considered the scenario where test samples come in sequence and label distribution changes with time. To address the \ols problem,~\citet{NIPS'21:Online-LS} proposed a risk estimator and used the vanilla OGD algorithm to learn over the online streams with label shift. They derived a static regret of order $\O(\sqrt{T})$, but the analysis crucially relied on the assumption of the convexity of the (expected) risk function, which is hard to be theoretically verified due to the use of 0/1-loss and the argmax operation. In contrast, by constructing an unbiased risk estimator with nice properties, our methods can adapt to dynamic environments with provable grantees. We provide more detailed discussions on the differences between the risk estimators in Appendix~\ref{sec-appendix:difference}.

\subsection{Online Learning in Non-stationary Environments}
\label{sec-appendix:related-work-oco}
This section first reviews existing results for non-stationary online convex optimization, including the worst-case dynamic regret and the universal dynamic regret. Then, we discuss the difference between our dynamic regret bounds and those in the previous works. We also present a brief introduction to readers unfamiliar with the OCO paradigm in Appendix~\ref{sec-appendix:preliminary-oco}.

\paragraph{Worst-case Dynamic Regret.} The rationale behind the static regret is that the best single decision could perform well over the time horizon. However, the underlying environments could change over the online learning process, making the static regret unsuitable. A better measure for the non-stationary environments is the worst-case dynamic regret $\DReg = \sum_{t=1}R_t(\w_t) - \sum_{t=1}^T R(\w_t^*)$, which compares the learner's decision with the best decision at each iteration $\w_t^* \in \argmin_{\w\in\W}R_t(\w)$ and has draw growing attention recently~\citep{ICML'03:zinkvich,OR'15:dynamic-function-VT,AISTATS'15:dynamic-optimistic,ICML'16:Yang-smooth,NIPS:2017:Zhang,ICML'18:zhang-dynamic-adaptive,conf/nips/BabyW19,OR'19:V_T-pq,UAI'20:Simple,L4DC'21:SC-Smooth}.

It is well-known that a sublinear regret bound is impossible for the dynamic regret in the worst case unless imposing certain regularities on the non-stationarity of the environments. There are two commonly studied non-stationarity measures. The first one is the \emph{path length} $P_T^* = \sum_{t=2}^T\norm{\w^*_t-\w^*_{t-1}}_2$, which reflects the fluctuation of the comparator sequence. When the loss functions are convex,~\citet{ICML'16:Yang-smooth} showed that OGD algorithm attains an $\O(\sqrt{T(1+P_T^*)})$ dynamic regret. The bound can be further improved to $\O(P_T^*)$ when the functions are convex and smooth, and the minimizer $\w_t^*$ lies in the interior of the feasible domain $\W$~\citep{ICML'16:Yang-smooth}, or the loss functions are strongly convex and smooth~\citep{CDC'16:dynamic-sc}. Another commonly studied non-stationarity measure is the \emph{temporal variability} $V_T^f = \sum_{t=2}^T \sup_{\w\in\W} \vert f_{t}(\w) - f_{t-1} (\w)\vert$, which quantifies the variation of the online functions. When the online functions are convex,~\citet{OR'15:dynamic-function-VT} showed an $\O(T^{2/3}{V_T^f}^{1/3})$ dynamic regret by the restarted OGD algorithm, and the bound can be further improved to $\O(T^{1/2}{V_T^f}^{1/2})$ for strongly convex functions. Then, the temporal variability bound was generalized to high-order deviations~\citep{conf/nips/BabyW19} and the $p,q$-norm~\citep{OR'19:V_T-pq}. There are also studies on the best-of-both-worlds bounds~\citep{AISTATS'15:dynamic-optimistic,UAI'20:Simple,L4DC'21:SC-Smooth}, where the path-length bound and temporal variability bound are achieved simultaneously by a single algorithm. For strongly convex and smooth functions, the best known result is $\O(\min\{P^*_T, S^*_T,V_T^f\})$ due to~\citet{L4DC'21:SC-Smooth}, where $S^*_T = \sum_{t=2}^T\norm{\w_t^* - \w_{t-1}^*}_2^2$ is the squared path length.

\paragraph{Universal Dynamic Regret.} The worst-case dynamic regret will sometimes be too pessimistic to guide the algorithm design as the function minimizer $\w_t^*$ can easily overfit to the noise~\citep{NIPS'18:Zhang-Ader}. To this end, the universal dynamic regret is proposed and defined as $\DReg(\u_1,\dots,\u_T) = \sum_{t=1}^T R_t(\w_t)-\sum_{t=1}^T R_t(\u_t)$, which supports the comparison to any feasible comparator sequence $\{\u_t\}_{t=1}^T$ with $\u_t\in\W$, and the measure has gained more and more attention~\citep{ICML'03:zinkvich,NIPS'18:Zhang-Ader,NIPS'20:sword,AISTATS'20:Zhang,ICML'20:Ashok,arXiv'21:Sword++,JMLR'21:BCO,conf/colt/BabyW21,AISTATS'22:memory,NeurIPS:2021:Zhang:A,ICML'22:mdp,ICML'22:TVgame}.~\citet{ICML'03:zinkvich} showed that OGD can achieve an $\O((1+P_T)\sqrt{T})$ universal dynamic regret, where $P_T = \sum_{t=2}^T\norm{\u_t-\u_{t-1}}_2$ is the path length of the comparator sequence $\{\u_t\}_{t=1}^T$. Nevertheless, the guarantee is far from the $\Omega(\sqrt{T(1+P_T)})$ minimax lower bound for convex functions as shown by~\citet{NIPS'18:Zhang-Ader}, who further developed an algorithm with an $\O(\sqrt{T(1+P_T)})$ universal dynamic regret and hence closed the gap~\citep{NIPS'18:Zhang-Ader}. Later, many studies are devoted to developing algorithms with tighter guarantees that can benefit from stronger curvature of loss functions~\citep{conf/colt/BabyW21}, or can adapt to the benign environments~\citep{NIPS'20:sword,arXiv'21:Sword++}, or to the more challenging setting with bandit feedback~\citep{JMLR'21:BCO}.

\paragraph{Discussion on Our Dynamic Regret Bounds. } Although there are many studies on non-stationary online learning, it is hard to apply existing methods to \ols, and our regret bound is novel. First of all, although our performance measure shares seemingly similarity with the worst-case dynamic regret, the algorithm with such bounds is not suitable for the \ols problem. The main reason is that existing methods require access to the loss function $R_t$ for the model update. However, in the \ols problem, the true loss function $R_t$ is established on the inaccessible data distribution $\D_t$ and the learner can only approximate it with $\hat{R}_t$ based on the empirical data. A direct application of algorithms with a worst-case dynamic regret can only lead to regret bounds defined over $\hat{R}_t$ such that the learned model would suffer from severe overfitting on the sample noise. A few previous works~\citep{OR'15:dynamic-function-VT,ICML'16:Yang-smooth,conf/nips/BabyW19} have attempted to develop methods to learn with the noisy loss function, but nevertheless the methods in~\citep{OR'15:dynamic-function-VT,ICML'16:Yang-smooth} require knowing the non-stationarity of the underlying distribution $\D_t$ ahead of time,  which is generally unavailable.

Given that the comparators in our regret measure are \emph{not} the optimizers of the loss functions, our dynamic regret measure is essentially a kind of \emph{universal} dynamic regret rather than the worst-case dynamic regret. Our results differ from the existing universal dynamic regret bounds appearing in the literature by carefully exploiting the structure of online label shift. In particular, existing bounds always scale with the path length of  compactors $P_T =\sum_{t=2}^T \norm{\u_t-\u_{t-1}}_2$, whereas our dynamic regret can adapt to the variation of label distribution $V_T = \sum_{t=2}^T \norm{\bm{\mu}_{y_t} - \bm{\mu}_{y_{t-1}}}_1$ directly, which is a more interpretable and suitable measure for \ols.

From the technical side, the most related works to us are~\citep{NIPS'18:Zhang-Ader,NIPS'20:sword} as all algorithms use the meta-base structure to achieve the dynamic regret bound. Besides the difference in the non-stationarity measure, our method improves the regret bound of~\citet{NIPS'18:Zhang-Ader} for convex functions by replacing the dependence on $T$ with the problem-dependent quantity $\G_T$. The $\G_T$ quantity can be much smaller than $T$ in benign environments, which implies a better adaptivity of our algorithm to underlying environments. The first dynamic regret bound with such adaptivity was achieved by~\citet{NIPS'20:sword}, but their work requires an additional smoothness assumption on the loss functions. We remove such a constraint via refined analyses for the meta-algorithm, which updates the weights with function values and an improved algorithm design for base-algorithm, which trains the model following an implicit update rule. The new finding might be of independent interest to the general online convex optimization. We note that a similar implicit update rule  also appeared in a concurrent work~\citep[Section 5]{colt'22:parameter-free-mirror-descent} for achieving problem-dependent \emph{static} regret. Our dynamic regret is much more challenging, as we not only need an implicit update in base-algorithms but also need a refined analysis of the meta-algorithm, which is unique in online ensemble analysis of dynamic regret minimization.

%% file: sections/appendices/comparison.tex
\subsection{Detailed Comparison with~\citet{NIPS'21:Online-LS} }
\label{sec-appendix:difference}
As mentioned in the main paper the work of~\citet{NIPS'21:Online-LS} is the closest one to our proposal, who also proposed a risk estimator to cope with \ols. In this part, we first review their result by restating their proposed risk estimator and then elaborate on the salient differences between their method and ours. The comparison demonstrates both theoretical and practical appeals of our proposed risk estimator in that it makes the OCO framework still applicable to \ols.

\paragraph {Estimation of $0/1$ risk.} The performance measure of~\citet{NIPS'21:Online-LS} is slightly different from ours, where the authors aim to minimize the regret defined over the $0/1$ risk
\begin{align*}
\mathop{\bm{\mathrm{Reg}}}\nolimits_{T}^{\mathtt{01}}=\sum_{t=1}^T R_{t}^{\mathtt{01}}(g_t) -\min_{g\in\mathcal{G}} \sum_{t=1}^T R^{\mathtt{01}}_t(g).
\end{align*}
In above, the $0/1$ risk is defined as $R_t^{\mathtt{01}}(g) = \E_{(\x,y)\sim\D_t}[\indicator\{g(\x)\neq y\}]$ and $g:\X\mapsto\Y$ is the classifier whose definition will be clear latter.
  Denoting by $\bm{\mu}_{y_t}$ the class prior vector with $[\bm{\mu}_{y_t}]_k = \D_t(y=k)$, the expected $0/1$ risk can be rewritten as
\begin{align}
    R_t^{\mathtt{01}}(g) ={}& \sum_{k=1}^K \E_{\x\sim\D_t\left(\x \givenn y = k \right)}\left[\indicator\{g(\x)\neq k\} \right]\cdot \left[\bm{\mu}_{y_t}\right]_{k}\notag\\
    ={}& \sum_{k=1}^K \E_{\x\sim\D_0\left(\x \givenn y = k\right)}\left[\indicator\{g(\x)\neq y\} \right]\cdot\left[\bm{\mu}_{y_t}\right]_{k}.\label{eq:0-1loss-rewriting}
\end{align}
In decomposition~\eqref{eq:0-1loss-rewriting}, the first term $\E_{\x\sim\D_0\left(\x \givenn y=k\right)}[\indicator\{g(\x)\neq y\}]$ is established on the class-conditional distribution $\D_0\left(\x \givenn y\right)$, which can be empirically approximated with the offline data $S_0$. The second term is the class prior at each iteration. One can estimate it with the BBSE estimator introduced in Section~\ref{sec:approach}. As a consequence, ~\citet{NIPS'21:Online-LS} empirically approximated $R_t^{\mathtt{01}}(g)$ by
\begin{align*}
    \hat{R}_t^{\mathtt{01}}(g) = \sum_{k=1}^K [\hat{\bm{\mu}}_{y_t}]_k\cdot \hat{R}^{\mathtt{01}}_{t,k}(g),
\end{align*}
where $\hat{\bm{\mu}}_{y_t} = C_{f_0}^{-1}\cdot \hat{\bm{\mu}}_{\hat{y}_t}$ is the estimated class prior returned by the BBSE model and $\hat{R}^{\mathtt{01}}_{t,k} = \frac{1}{\vert S_{0}^k\vert}\sum_{\x_n\in S_0^k}\indicator\{g(\x_n)\neq k\}$ is the model's empirical risk over $S_0^k$. 

\paragraph{Reweighting classifier.} To directly optimize the $0/1$ loss,~\citet{NIPS'21:Online-LS} further restricted the model's structure by focusing on the reweighting algorithm. The basic idea is, the optimal classifier that minimizes $R_t^{\mathtt{01}}(g)$ is $g_*(\x) = \argmax_{k\in[K]} \D_t\left(y = k \givenn \x\right)$. Under the class shift condition, they rewrote $\D_t\left(y=k \givenn \x\right)$ as
\begin{align}
    \D_t\left(y=k \givenn \x \right) ={}& \frac{\D_t(y = k)\D_t\left(\x \givenn y = k\right)}{\D_t(\x)} = \frac{\D_t(y = k)\D_0\left(\x \givenn y = k\right)}{\D_t(\x)}\notag\\
    ={}&\frac{\D_t(y=k)}{\D_0(y=k)}\frac{\D_0(\x)}{\D_t(\x)}\D_0\left(y=k \givenn \x \right)\propto \frac{\D_t(y=k)}{\D_0(y=k)}\D_0\left(y=k \givenn \x \right).\label{eq:0-1 classifier}
\end{align}

Since the $\D_0$ can be approximate by the offline data $S_0$, the leaner can train a probability model $g_0:\X\mapsto\Delta_K$ on $S_0$ to approximate $\D_0\left(y\givenn \x\right)$. Then, by the relationship~\eqref{eq:0-1 classifier}, one can construct a classifier $g_{\p}:\X\mapsto[K]$ parameterized by the reweighting parameter $\p\in\Delta_K$ to approximate the optimal classifier $g_*$ by
\begin{align}
    g_{\p}(\x) = \argmax_{y\in[K]}  \frac{1}{Z(\x)} \frac{[\p]_k}{\D_0(y=k)}\cdot [g_0(\x)]_k, \label{eq:0-1-estimate}
\end{align}
where $Z(\x) = \sum_{k=1}^K\frac{[\p]_k}{\D_0(y=k)}\cdot [g_0(\x)]_k $ is the normalization factor and the initial label distribution $\D_0(y=k)$ can be estimated with the offline data $S_0$. When $g_0$ provides a sufficiently accurate approximation of $\D_t\left(y \givenn \x \right)$, one can show that $g_* = \argmin_{\p\in\Delta_{K}} R_t^{\mathtt{01}}(g_{\p})$. 

\paragraph{Comparisons.}
In summary, the differences between the risk estimator of~\citet{NIPS'21:Online-LS} and ours are mainly in the following two aspects:
\begin{enumerate}[(i)]
    \item \textbf{Loss function.} As shown in \eqref{eq:0-1loss-rewriting},~\citet{NIPS'21:Online-LS} used the non-convex $0/1$ loss $\indicator\{g(\x)\neq y\}$ for evaluation. Theoretically, the non-convexity of $0/1$ loss makes it hard to confirm the convexity of the $0/1$ risk, which is the critical condition for the regret bound. Practically, the gradient of $0/1$ loss is always zero, which brings troubles to the optimization. By contrast, we use a convex surrogate loss in our risk estimator to address the above problems. 
    \item \textbf{Classifier.} To match the 0/1 loss function,~\citet{NIPS'21:Online-LS} designed a classifier based on the reweighting algorithm, which directly updates the reweighting parameter $\mathbf{p}$ in a simplex. Theoretically, as shown in~\eqref{eq:0-1-estimate}, there is a troublesome argmax operation in the formulation of the classifier, which makes it non-convex and then hurts the convexity of the entire risk. Practically, only adjusting the outputs of the initial classifier $f_0^\mathtt{p}$ by reweighting might limit the learning capability of the model, especially when the initial classifier does not perform well enough. By comparison, we employ a strictly convex classifier and update the whole parameters in a convex parameter space, which confirms the convexity of the risk and the flexibility to better adapt to the environmental changes. The classifier in~\eqref{eq:0-1-estimate} is hard to be convex even if we replace the 0/1 loss with convex surrogate loss (details elaborated below).
\end{enumerate}

We finally remark that even if we modify the risk estimator of~\citet{NIPS'21:Online-LS} by optimizing the convex surrogate loss rather than the $0/1$ loss, it remains unclear how to apply the OCO framework to this modified estimator. The critical challenge is that even if the argmax operation can be avoided, another renormalization process is required due to the reweighting mechanism, making it hard to confirm the convexity of online functions. By contrast, our risk estimator well fits the OCO framework and thus can achieve favorable dynamic regret guarantees with suitable online update rules. 

%% file: sections/appendices/proofs.tex

\section{Omitted Details for Section~\ref{sec:unbiased-risk-estimator}}
\label{sec-appendix:unbiased-estimator}
This section presents the omitted details for Section~\ref{sec:unbiased-risk-estimator}.
\subsection{Proof of Lemma~\ref{lemma:unbiased-estimator}}
\label{sec-appendix:proof-lemma1}
\begin{proof}[Proof of Lemma~\ref{lemma:unbiased-estimator}]
Recall the definition of the expected  risk~\eqref{eq:risk-decomposition} as
$$R_t(\w) =\sum_{k=1}^K [\bm{\mu}_{y_t}]_k\cdot R_0^k(\w),$$
where $\bm{\mu}_{y_t}$ is the true class prior at iteration $t$ with $[\bm{\mu}_{y_t}]_k = \D(y = k)$. Besides, the risk estimator~\eqref{eq:unbiased-risk-estimator} is defined by
$$
\hat{R}_t(\w) 
= \sum_{k=1}^K [\hat{C}_{f_0}^{-1}\hat{\bm{\mu}}_{\hat{y}_t}]_k\cdot \hat{R}_0^k(\w), 
$$
where $\hat{\bm{\mu}}_{\hat{y}_t}\in\Delta_K $ with $[\hat{\bm{\mu}}_{\hat{y}_t}]_k = ({1}/{\vert S_t\vert})\sum_{\x\in S_t}{\indicator\{f_0(\x) = k\}}$ is the estimated class prior of the prediction $f_0(\x)$.

Given that the initial offline data has sufficient samples such that $\hat{C}_{f_0} = C_{f_0}$ and $\hat{R}_0^k(\w) = R_0^k(\w)$ and the model $\w\in\W$ is independent of $S_t$, we have
\begin{align*}
\E_{S_t\sim\D_t}[\Rh_t(\w)] ={}& \E_{S_t\sim\D_t}\left[\sum_{k=1}^K [{\hat{C}}_{f_0}^{-1}\hat{\bm{\mu}}_{\hat{y}_t}]_k\cdot {\hat{R}}_0^k(\w)\right]\\ 
={}&\sum_{k=1}^K \left[{\hat{C}}_{f_0}^{-1}\E_{S_t\sim\D_t}\big[\hat{\bm{\mu}}_{\hat{y}_t}\big]\right]_k\cdot {\hat{R}}_0^k(\w)\\
={}&\sum_{k=1}^K \left[{\hat{C}}_{f_0}^{-1}{\bm{\mu}}_{\hat{y}_t}\right]_k\cdot {\hat{R}}_0^k(\w),
\end{align*}
where the first equality is due to the definition of $\hat{R}_t(\w)$. The second inequality comes from the fact that $\Chf$ and $\Rh_0^k$ are independent of $S_t$. The last equality is a consequence of the unbiasedness of $\hat{\bm{\mu}}_{\hat{y}_t}$, i.e., $\E_{S_t\sim\D_t}[\hat{\bm{\mu}}_{\hat{y}_t}] = {\bm{\mu}}_{\hat{y}_t}$, where $[{\bm{\mu}}_{\hat{y}_t}]_k = \E_{\x\sim\D_t(\x)}[\indicator\{f_0(\x)=k\}]$ is the expected class prior of the prediction $f_0(\x)$. 

Then, under the condition that $\hat{C}_{f_0} = C_{f_0}$ and $\hat{R}_0^k(\w) = R_0^k(\w)$, we can further obtain 
\begin{align*}
    \E_{S_t\sim\D_t}[\Rh_t(\w)] =\sum_{k=1}^K \left[{{C}}_{f_0}^{-1}{\bm{\mu}}_{\hat{y}_t}\right]_k\cdot {{R}}_0^k(\w) = \sum_{k=1}^K \left[\bm{\mu}_{y_t}\right]_k\cdot {{R}}_0^k(\w) = R_t(\w),
\end{align*}
where the second equality is due to the relationship $\bm{\mu}_{y_t} = C_{f_0}^{-1}\bm{\mu}_{\hat{y}_t}$ as introduced by~\eqref{eq:bbse-matrix}. The last inequality comes from the definition of $R_t(\w)$, which completes the proof.
\end{proof}

\subsection{Concentration of Risk Estimator}

\label{sec-appendix:risk-estimator-hp}
In this section, we show that the risk estimator converges to the expected risk at the rate of $\mathcal{O}(\sqrt{1/\vert S_0\vert})$ with high probability.

\begin{myLemma}
\label{lemma:high-prob}
  Let $\delta\in(0,1/4]$. For any $\w\in\W$ independent of the dataset $S_t$, with probability at least $1-(K+3)\delta$, the risk estimator $\Rh_t(\w)$ in \pref{eq:unbiased-risk-estimator} satisfies
  \begin{align*}
    &\vert \E_{S_t\sim \D_t}[\Rh_t(\w)] - R_t(\w)\vert   \leq\frac{4\sqrt{K}B}{\kappa^2\sigma^2}\left( \frac{\log(2K/\delta)}{\vert S_0 \vert} +\sqrt{\frac{2\log (2K/\delta)}{\vert S_0 \vert}}\right),
  \end{align*}
  given that $\vert S_0\vert\geq \left(25\log(2K/\delta)\right)/(\kappa\sigma)^2$, where $\kappa = \min_{k\in[K]} \D_0(y = k)$ and $\sigma$ is the minimum singular value of $\Cf$.
\end{myLemma}

\begin{proof}[Proof of Lemma~\ref{lemma:high-prob}] For notation convenience, we denote by $\bm{R}_0:\W\mapsto\R^{K}$ the vector risk function with its $k$-th entry $[\bm{R}_0(\w)]_k =  R_0^k(\w)$ and its empirical version $\hat{\bm{R}}_0:\W\mapsto\R^K$ with $[\hat{\bm{R}}_0(\w)]_k = \hat{R}_0^k(\w)$. Under such a case, we can rewrite the risk estimator as $\Rh_t(\w) = \hat{\bm{\mu}}_{\hat{y}_t}^\top \Chf^{-1}\hat{\bm{R}}_0(\w)$ and the expected risk as $R_t(w) = {\bm{\mu}}_{\hat{y}_t}^\top \Cf^{-1}{\bm{R}}_0(\w)$. Then, since $S_t$ is independent of $S_0$, we have
\begin{align*}
\E_{S_t\sim\D_t}[\hat{R}_t(\w)] 
= \E_{S_t\sim\D_t}[\hat{\bm{\mu}}_{\hat{y}_t}^\top \Chf^{-1}\hat{\bm{R}}_0(\w)] = \E_{S_t\sim\D_t}[\hat{\bm{\mu}}_{\hat{y}_t}^\top] \Chf^{-1}\hat{\bm{R}}_0(\w)=
{\bm{\mu}}_{\hat{y}_t}^\top \Chf^{-1}\hat{\bm{R}}_0(\w).
\end{align*}

Let $W_{f_0}\in\R^{K}$ be the diagonal matrix with $[W_{f_0}]_{ii} = \D_0(y = i)$ and $[J_{f_0}]_{ij} =  \E_{(\x,y)\sim \D_0}[\indicator\{f_0(\x) = i \mbox{ and } y=j\}]$ be the confusion matrix defined over the joint distribution of $f_0(\x)$ and $y$. Besides, denote by $[\hat{W}_{f_0}]_{ii} = ({1}/{\vert S_0\vert})\sum_{(\x,y)\in S_0}\indicator\{y = i\}$ and $\hat{J}_{f_0} = (1/\vert S_0\vert)\sum_{(\x,y)\in S_0} \indicator\{f_0(\x) = i \mbox{ and } y = j\}$ the empirical versions. Definition~\eqref{eq:definition-C} indicates that 
\begin{equation}
\label{eq:defnition-J}
    J_{f_0} = W_{f_0}\cdot \Cf \quad \mbox{and} \quad \hat{J}_{f_0} = \hat{W}_{f_0}\cdot \Chf.
\end{equation}

Under such a case, the empirical and expected risk can be further rewritten as 
\[
\E_{S_t\sim\D_t}[\Rh_t(\w)] = \bm{\mu}_{\hat{y}_t}^\top \hat{J}_{f_0}^{-1} \hat{W}_{f_0} \bm{\hat{\bm{R}}_0(\w)}\quad\mbox{and}\quad R_t(\w) = \bm{\mu}_{\hat{y}_t}^\top J_{f_0}^{-1} W_{f_0} \bm{\bm{R}_0(\w)}.
\]

So, we have 
\begin{align*}
    {}&\E_{S_t\sim\D_t}\left[{\Rh_t(\w)}\right] - R_t(\w)\\
    \leq{}& \bm{\mu}_{\hat{y}_t}^\top \hat{J}_{f_0}^{-1} \hat{W}_{f_0} \bm{\hat{\bm{R}}_0(\w)} -\bm{\mu}_{\hat{y}_t}^\top J_{f_0}^{-1} W_{f_0} \bm{\bm{R}_0(\w)} \\
    \leq{}&\underbrace{\bm{\mu}_{\hat{y}_t}^\top \hat{J}_{f_0}^{-1} \hat{W}_{f_0} \bm{\hat{\bm{R}}_0(\w)} -\bm{\mu}_{\hat{y}_t}^\top \Jf^{-1} \hat{W}_{f_0} {\hat{\bm{R}}_0(\w)}}_{\mathtt{term~(a)}} + \underbrace{\bm{\mu}_{\hat{y}_t}^\top \Jf^{-1} \hat{W}_{f_0} {\hat{\bm{R}}_0(\w)}-\bm{\mu}_{\hat{y}_t}^\top \Jf^{-1} W_{f_0} {\bm{R}_0(\w)}}_{\mathtt{term~(b)}}.
\end{align*}

For term~(a), with probability at least $1-\delta$, we have 
\begin{align*}
    \texttt{term~(a)} ={}& \bm{\mu}_{\hat{y}_t}^\top (\Jf^{-1}-\Jhf^{-1}) \hat{W}_{f_0} {\hat{\bm{R}}_0(\w)}\\
    \leq{}&\norm{\bm{\mu}_{\hat{y}_t}}_2\cdot\norm{\Whf\Rhb_0(\w)}_2\cdot\norm{\Jhf^{-1}-\Jf^{-1}}_2\\
    \leq{}&\sqrt{K}B\norm{\Jhf^{-1}-\Jf^{-1}}_2\\
    \leq {}&\frac{2\sqrt{K}B}{\kappa^2\sigma^2}\left(\frac{\log(2K/\delta)}{\vert S_0 \vert} + \sqrt{\frac{2\log (2K/\delta)}{\vert S_0 \vert}}\right),
\end{align*}
where the second inequality is due to the Cauchy-Schwarz inequality and the property of the operator norm. The third inequality holds since $\norm{\bm{\mu}_{\hat{y}_t}}_2\leq 1$ and $\norm{\Whf\Rhb_0(\w)}_2\leq \norm{\Whf}_2\cdot\norm{\Rhb_0(\w)}_2\leq \sqrt{K}B$. The last inequality is due to Lemma~\ref{lemma:concentration-J}.

Next, we analyze the term~(b).
\begin{align*}
    \texttt{term~(b)} ={}& \bm{\mu}_{\hat{y}_t}^\top \Jf^{-1} (\hat{W}_{f_0} {\hat{\bm{R}}_0(\w)}-{W}_{f_0} {{\bm{R}}_0(\w)})\\
    \leq{}& \norm{\bm{\mu}_{\hat{y}_t}}_2\cdot\norm{\Jf^{-1}}_2\cdot \norm{\hat{W}_{f_0} {\hat{\bm{R}}_0(\w)}-{W}_{f_0} {{\bm{R}}_0(\w)}}_2\\
    \leq{}&\frac{1}{\kappa\sigma} \norm{\hat{W}_{f_0} {\hat{\bm{R}}_0(\w)}-{W}_{f_0} {{\bm{R}}_0(\w)}}_2,
\end{align*}
where the third inequality is due to the fact that $\norm{\bm{\mu}_{\hat{y}_t}}_2\leq 1$ and $\norm{J_{f_0}^{-1}}_2\leq \norm{\Wf^{-1}\Cf^{-1}}_2\leq 1/(\kappa\sigma)$. The term $\norm{\hat{W}_{f_0} {\hat{\bm{R}}_0(\w)}-{W}_{f_0} {{\bm{R}}_0(\w)}}_2$ can be further decomposed as 
\begin{align*}
    \norm{\hat{W}_{f_0} {\hat{\bm{R}}_0(\w)}-{W}_{f_0} {{\bm{R}}_0(\w)}}_2 ={}&\norm{\hat{W}_{f_0} {\hat{\bm{R}}_0(\w)}-{W}_{f_0} {{\hat{\bm{R}}}_0(\w)}}_2 + \norm{{W}_{f_0} {\hat{\bm{R}}_0(\w)}-{W}_{f_0} {{\bm{R}}_0(\w)}}_2 \\
    \leq {}& \norm{\Rhb_0(\w)}_2\cdot\norm{\Whf-\Wf}_2 + \norm{\Wf}_2\cdot\norm{\Rhb_0(\w) - \bm{R}_0(\w)}_2\\
    \leq {}& B \norm{\Whf-\Wf}_2 + \norm{\Rhb_0(\w) - \bm{R}_0(\w)}_2,
\end{align*}
where the first inequality is due to the property of the operator norm. The second inequality is a consequence of the fact that $\norm{\Rhb(\w)}_2\leq B$ for any $\w\in\W$ and $\norm{\Wf}_2\leq 1$. 

Then, by the matrix Bernstein inequality in~\citet[Theorem 1.4]{FCM'12:Matrix-Concentration}, we have
\begin{align}
\label{eq:proof-of-termb-1}
\norm{W_{f_0} -\hat{W}_{f_0}}_2 \leq \frac{2\log(2K/\delta)}{3\vert S_0 \vert} + \sqrt{\frac{2\log (2K/\delta)}{\vert S_0 \vert}}
\end{align}
with probability at least $1-\delta$.

Now, we turn to bound the term $\norm{\Rhb_0(\w) - \bm{R}_0(\w)}_2$. Let $N_{\min} = \min\{\vert S_0^1\vert,\dots,\vert S_0^{K}\vert\}$. According to Lemma~\ref{lemma:number-of-sample}, we can first show that $N_{\min}\geq (\kappa \vert S_0 \vert)/2$ with probability at least $1-\delta$ for any $\delta\in(0,1/4]$. Then, we can bound the last term by 
\begin{align*}
    \norm{\Rhb_0(\w) - \bm{R}_0(\w)}_2 \leq \sqrt{K}\max_{k\in[K]}\{\vert
R_0^{k}(\w)-\Rh^{k}_0(\w)\vert\}.
\end{align*}
According to the Hoeffding's lemma, for each $k\in[K]$, we can bound 
\begin{align*}
    \vert
R_0^{k}(\w)-\Rh^{k}_0(\w)\vert\leq \sqrt{\frac{B^2\log (1/\delta)}{2 \vert S_0^k\vert}} \leq \sqrt{\frac{B^2\log (1/\delta)}{2 N_{\min}}}
\end{align*}
with probability at least $1-\delta$. Combining the $K$ events, which holds with probability  $1-\delta$, and the event that $N_{\min}\geq (\kappa \vert S_0 \vert)/2$, we can obtain that  
\begin{align}
\label{eq:proof-of-termb-2}
    \norm{\Rhb_0(\w) - \bm{R}_0(\w)}_2 \leq \sqrt{\frac{KB^2\log({1}/{\delta})}{\kappa \vert S_0 \vert}}
\end{align}
with probability at least $1-(K+1)\delta$. 

Putting~\eqref{eq:proof-of-termb-1} and~\eqref{eq:proof-of-termb-2} together, we have 
\begin{align*}
    \texttt{term~(b)} \leq \frac{B}{\kappa\sigma}\left(\frac{2\log(2K/\delta)}{3\vert S_0 \vert} + \sqrt{\frac{2\log (2K/\delta)}{\vert S_0 \vert}} +\sqrt{\frac{K\log({1}/{\delta})}{\kappa \vert S_0 \vert}}\right)
\end{align*}
with probability at least $1-(K+2)\delta$. Combining the upper bounds for term (a) and term (b), we can complete the proof by
\begin{align*}
    &\vert \E_{S_t\sim \D_t}[\Rh_t(\w)] - R_t(\w)\vert   \leq\frac{4\sqrt{K}B}{\kappa^2\sigma^2}\left( \frac{\log(2K/\delta)}{\vert S_0 \vert} +\sqrt{\frac{2\log (2K/\delta)}{\vert S_0 \vert}}\right)
\end{align*}
with probability at least $1-(K+3)\delta$.

\end{proof}

\subsection{Static Regret of UOGD Algorithm}
\label{sec-appendix:static-regret}

The following lemma presents the static regret guarantees of the UOGD algorithm when the step size is set as $\eta = \Theta(T^{-1/2})$.

\begin{myLemma}
\label{lemma:OGD-2}
{}Under the same assumptions of Lemma~\ref{lemma:unbiased-estimator}, UOGD in~\pref{eq:UOGD} with a step size $\eta = \Theta (T^{-1/2})$ satisfies
\begin{align*}
\E_{1:T}\left[\SReg\right] \leq \frac{2G\Gamma}{\sigma}\sqrt{KT} = \O(\sqrt{T}),
\end{align*} 
where $\E_{1:T}[\cdot]$ denotes the expectation taken over the random draw of dataset $\{S_t\}_{t=1}^T$. The constant $\sigma > 0 $ denotes the minimum singular value of confusion matrix $C_{f_0}$. 
\end{myLemma}

\subsection{Useful Lemmas}
This section presents several useful lemmas for the proofs in Appendix~\ref{sec-appendix:unbiased-estimator}. 

\begin{myLemma}
\label{lemma:concentration-J}
Let $\Jf$ and $\Jhf$ be the matrix defined as~\eqref{eq:defnition-J}. Then, for any $\delta\in[0.1)$, with probability at least $1-\delta$  we have 
  \begin{align*}
    \norm{\Jhf^{-1}-\Jf^{-1}}_2 \leq     \frac{2\log(2K/\delta)}{\kappa^2\sigma^2\vert S_0 \vert} + \frac{2}{\kappa^2\sigma^2}\sqrt{\frac{2\log (2K/\delta)}{\vert S_0 \vert}}, 
  \end{align*}
  given that $\vert S_0\vert\geq \left(25\log(2K/\delta)\right)/(\kappa\sigma)^2$, where $\kappa = \min_{k\in[K]} \D_0(y = k)$ and $\sigma$ is the minimum singular value of $\Cf$.
\end{myLemma}
\begin{proof}[Proof of Lemma~\ref{lemma:concentration-J}]

Denoting by $\Delta_J = \Jhf-\Jf$, according to~\citet[Lemma~2]{conf/iclr/Azizzadenesheli19}, we have
\begin{align*}
    \norm{\Delta_J}_2\leq     \frac{2\log(2K/\delta)}{3\vert S_0 \vert} + \sqrt{\frac{2\log (2K/\delta)}{\vert S_0 \vert}}\leq \frac{\kappa\sigma}{2}
\end{align*}
with probability at least $1-\delta$, where the last inequality is due to the assumption that $\vert S_0\vert\geq (\log (2K/\delta))/(\sigma\kappa)^2$. We can check the condition for Lemma~\ref{lemma:matrix-gap}
\begin{equation*}
    \norm{\Jf^{-1}\Delta_J}_2 \leq \norm{\Cf^{-1}}_2\cdot\norm{\Wf^{-1}}_2\cdot\norm{\Delta_J}_2\leq \frac{1}{\kappa \sigma}\cdot \frac{\kappa \sigma}{2},
\end{equation*}
where the last inequality holds since $\sigma$ is the minimum singular value of $\Cf$ and $\kappa$ is that of $\Wf$.

Then, according to Lemma~\ref{lemma:matrix-gap}, we can complete the proof by
\begin{align*}
    \norm{\Jhf^{-1}-\Jf^{-1}}_2 \leq 2 \norm{\Jf^{-1}}_2^2 \norm{\Delta}_2\leq     \frac{2\log(2K/\delta)}{\kappa^2\sigma^2\vert S_0 \vert} + \frac{2}{\kappa^2\sigma^2}\sqrt{\frac{2\log (2K/\delta)}{\vert S_0 \vert}}. 
\end{align*}
\end{proof}

\begin{myLemma}
\label{lemma:number-of-sample}
Let $\vert S_0 \vert\geq \left(16\log (1/\delta)\right)/\kappa^2$. Then, for any $\delta\in(0,1/4]$, with probability at least $1-\delta$ the following holds: 
$$
N_{\min} \geq \frac{\kappa \vert S_0 \vert}{2},
$$
where $N_{\min} = \min\{\vert S_0^1\vert,\dots,\vert S_0^K\vert\}$ and $\kappa = \min_{k\in[K]}\D_0(y = k)$.
\end{myLemma}

\begin{proof}[Proof of Lemma~\ref{lemma:number-of-sample}]
For notation simplicity, we denote by $\tilde{\bm{\mu}}_0$ the empirical class prior at $t=0$, i.e. $[\tilde{\bm{\mu}}_0]_k = \vert S_0^k\vert /  \vert S_0\vert$ and $\bm{\mu}_0$ the expected class prior with $[\bm{\mu}_0]_k = \D_0(y=k)$. Then, according to~\citet[Proposition 19]{JCSS'12:Hsu}, we have 
\begin{align*}
    \norm{\tilde{\bm{\mu}}_0 - \bm{\mu}_0}_2\leq \frac{1}{\sqrt{\vert S_0 \vert}}+\sqrt{\frac{\log (1/\delta)}{\vert S_0 \vert}}
\end{align*}
with probability at least $1-\delta$, which implies that 
\begin{align*}
    [\tilde{\bm{\mu}}_0]_k\geq [{\bm{\mu}}_0]_k - \frac{1}{\sqrt{\vert S_0 \vert}} - \sqrt{\frac{\log (1/\delta)}{\vert S_0 \vert}}, \quad \forall k\in[K]
\end{align*}
holds with probability at least $1-\delta$. We complete the proof by noticing that $[\bm{\mu}_0]_k\geq \kappa$ holds for any $k\in[K]$ and  further noting the assumption that $\vert S_0 \vert\geq \frac{16\log(1/\delta)}{\kappa^2}$.
\end{proof}

\section{Omitted Proofs for Section~\ref{sec:online-algorithm}}
\label{sec-appendix:proof-sec3.2}

This section provides the proof omitted in section~\ref{sec:online-algorithm}. We will first present the regret bound for the UOGD algorithm without step size tuning (Theorem~\ref{thm:UOGD}), and then show the overall dynamic regret bound of the \textsc{Atlas} algorithm (Theorem~\ref{thm:overall-UOGD}). At the end of this section, we list several useful lemmas.

Before presenting the proofs, we highlight the main challenge of the regret analysis. As noted by Section~\ref{sec:online-algorithm} and Appendix~\ref{sec-appendix:related-work-oco}, our performance measure~\eqref{eq:dynamic-regret-def} is different from the conventional notion of (worst-case) dynamic regret~\citep{AISTATS'15:dynamic-optimistic}, where the performance of the learned model is measured by the observed loss function. However, in the \ols problem, the model is desired to perform well over the underlying distribution, and its quality is evaluated over the expected risk function $R_t(\cdot)$. The main challenge here is that the expected risk function $R_t(\cdot)$ is unavailable in the learning process, which makes it hard to apply the analysis of (worst-case) dynamic regret for our case. 

\subsection{Proof of Theorem~\ref{thm:UOGD}}
\label{appendix:sec-proof-UOGD}
This part presents the proof of Theorem~\ref{thm:UOGD}. The key idea is to decompose the overall dynamic regret into two parts by introducing a reference sequence, which changes in a piecewise-stationary manner. In our analysis, the first part measures the regret of our algorithm when compared with the reference sequence, and the second part reflects the quality of the reference sequence. Since the reference sequence appears only in the analysis, we can make a tight regret bound by choosing a proper one to balance the first and second parts.

\begin{proof}[Proof of Theorem~\ref{thm:UOGD}]
We can decompose the regret bound into two parts by introducing a reference sequence $\{\u_t\}_{t=1}^T$ that only changes every $\Delta$ iteration. More specifically, denoting by $\mathcal{I}_m = [(m-1)\Delta+1,m\Delta]$ the $m$-th time interval, any comparator $\u_t$ that falls into $\mathcal{I}_m$ is taken as the single best decision over the interval, i.e. $\u_t = \w_{\mathcal{I}_m}^* \in \argmin_{\w\in\W} \sum_{t\in\mathcal{I}_m} R_t(\w)$ for any $t\in\mathcal{I}_m$. We have
\begin{align*}
&\E_{1:T}\left[\sum_{t=1}^T R_t(\w_{t})\right] - \sum_{t=1}^{T} R_t(\w^*_t)\\
={}&\E_{1:T}\left[\sum_{t=1}^T R_t(\w_{t})\right] -\sum_{t=1}^{T} R_t(\u_t) +\sum_{t=1}^{T} R_t(\u_t) - \sum_{t=1}^{T} R_t(\w^*_t)\\
={}&\underbrace{\E_{1:T}\left[\sum_{t=1}^T R_t(\w_{t})\right] - \sum_{m=1}^{M}\sum_{t\in\mathcal{I}_m} R_t(\w_{\mathcal{I}_m}^*)}_{\texttt{term~(a)}} + \underbrace{\sum_{m=1}^M\sum_{t\in\mathcal{I}_m} R_t(\w_{\mathcal{I}_m}^*) - \sum_{t=1}^{T} R_t(\w^*_t)}_{\texttt{term~(b)}},
\end{align*}
where $M =\left\lceil\frac{T}{\Delta}\right\rceil \leq T/\Delta + 1$ is the number of the intervals. Then, we turn to analyze term (a) and term (b), respectively.

\paragraph{Analysis for term (a).} Term (a) measures the regret of our algorithm when compared with a piecewise-stationary sequence. We first show that the regret defined over the expected risk $R_t(\cdot)$ can be related to the empirical risk estimator $\hat{R}_t(\cdot)$ due to its unbiased property.
\begin{align}
&\texttt{term~(a)}\notag\\
={}&\E_{1:T}\left[\sum_{t=1}^TR_t(\w_{t})\right] - \sum_{t=1}^TR_t(\u_t)\notag \notag\\
\leq{}& \E_{1:T}\left[\sum_{t=1}^T\inner{\nabla R_t(\w_{t})}{\w_{t}-\u_t}\right]\notag\\
={}&\underbrace{\E_{1:T}\left[{\sum_{t=1}^T\inner{\nabla R_t(\w_{t})-\nabla \Rh_t(\w_{t})}{\w_{t}-\u_t}}\right]}_{\texttt{term~(a-1)}}+\underbrace{\E_{1:T}\left[\sum_{t=1}^T[\inner{\nabla \Rh_t(\w_{t})}{\w_{t}-\u_t}\right]}_{\texttt{term~(a-2)}},\label{eq:proof-lemma-OGD-A1}
\end{align}
where the first inequality is due to the convexity of the risk function $R_t(\cdot)$. For term (a-1), we have 
\begin{align*}
\mathtt{term~(a}\texttt{-}\mathtt{1)} ={}& \E_{1:T}\left[\inner{\nabla R_t(\w_{t})-\nabla \Rh_t(\w_{t})}{\w_{t}-\u_t}\right]\notag\\
={}&\E_{1:t-1}\left[\inner{\nabla R_t(\w_{t})-\E_{t}\left[\nabla \Rh_t(\w_{t}) \givenn 1:t-1\right]}{\w_{t}-\u_t}\right]=0,
\end{align*}
where the last equality is due to the unbiasedness of the risk estimator $\hat{R}_t$ such that $\nabla R_t(\w_{t})=\E_{t}\left[\nabla \Rh_t(\w_{t}) \givenn 1:t-1\right]$. Thus, it is sufficient to analyze term (a-2) to provide an upper bound for term (a). For the model sequence $\{\w_t\}_{t=1}^T$ generated by the UOGD algorithm in~\pref{eq:UOGD}, we have the following lemma, whose proof is deferred to Appendix~\ref{appendix:sec-proof-overall-UOGD-lemmas}. 

\begin{myLemma}
\label{lemma:OGD-P_T}
Under same assumptions of Theorem~\ref{thm:UOGD}, UOGD in~\pref{eq:UOGD} with a step size $\eta > 0$ satisfies
\begin{align*}
\sum_{t=1}^T\inner{\nabla \hat{R}_t(\w_t)}{\w_t-\u_t} \leq \frac{2\eta K G^2T}{\sigma^2} + \frac{2\Gamma P_T + \Gamma^2}{2\eta}
\end{align*}
for any comparator sequence $\{\u_t\}_{t=1}^T$ with $\u_{t}\in\W$. Moreover, $P_T = \sum_{t=2}^T \norm{\u_t-\u_{t-1}}_2$ measures the variation of the comparator sequence.
\end{myLemma}

Since the comparator sequence in term (a) only changes $M-1$ times, its variation is bounded by $P_T\leq \Gamma (M-1)\leq (\Gamma T)/\Delta$. By Lemma~\ref{lemma:OGD-P_T} and taking an expectation over both sides, we have
\begin{align*}
\texttt{term~(a-2)}\leq \frac{2\eta K G^2T}{\sigma^2} + \frac{2\Gamma^2 T/\Delta + \Gamma^2}{2\eta}.
\end{align*}
Combining the upper bounds of term(a-1) and term (a-2) yields
\begin{align}
\label{eq:proof-thm1-terma}
\texttt{term~(a)}\leq \texttt{term~(a-1)} + \texttt{term~(a-2)}\leq \frac{2\eta K G^2T}{\sigma^2} + \frac{2\Gamma^2 T/\Delta + \Gamma^2}{2\eta}.
\end{align}

\paragraph{Analysis for term (b).} For term (b), we can follow the reasoning in~\citet{OR'15:dynamic-function-VT} and subsequently simplified analysis in~\citet[Lemma 1]{AAAI'20:DynamicOMD} to show that
\begin{align}
    \label{eq:proof-theorem1-V_T}
\texttt{term~(b)}={}& \sum_{m=1}^M\sum_{t\in\mathcal{I}_m}\left( R_t(\w_{\mathcal{I}_m}^*) - R_t(\w^*_t)\right)\notag\\
\leq{}&\sum_{m=1}^M\sum_{t\in\mathcal{I}_m}\left( R_t(\w_{s_m}^*) - R_t(\w^*_t)\right)\notag\\
={}&\sum_{m=1}^M\sum_{t\in\mathcal{I}_m}\left( R_t(\w_{s_m}^*) - R_{s_m}(\w_{s_m}^*) + R_{s_m}(\w_{s_m}^*) - R_t(\w^*_t)\right)\notag\\
\leq{}&\sum_{m=1}^M\sum_{t\in\mathcal{I}_m}\left( R_t(\w_{s_m}^*) - R_{s_m}(\w_{s_m}^*) + R_{s_m}(\w_{t}^*) - R_t(\w^*_t)\right)\notag\\
\leq {}& 2\Delta \sum_{m=1}^M\sum_{t\in\mathcal{I}_m} \sup_{\w\in\W }\left\vert R_t(\w) - R_{t-1}(\w)\right\vert \notag\\
= {}& 2\Delta \sum_{t=2}^T \sup_{\w\in\W }\left\vert R_t(\w) - R_{t-1}(\w)\right\vert\notag\\
\triangleq {}& 2B\Delta V_T^{R},
\end{align}
where $s_m = (m-1)\Delta+1$ is the first time step at interval $\mathcal{I}_m$. In the above, the first inequality is due to the optimality of $\w_{\mathcal{I}_m}^*$ over the interval $\mathcal{I}_m$. The second inequality holds since $\w_{s_m}^* \in \argmin_{\w\in\W}R_{s_m}(\w)$. 

Combining term (a) and term (b), we have 
\begin{align}
    \label{uogd-vt-bound}
\E_{1:T}\left[\sum_{t=1}^T R_t(\w_{t})\right] - \sum_{t=1}^{T} R_t(\w^*_t)\leq{}& \frac{2\eta K G^2T}{\sigma^2} + \frac{2\Gamma^2 T/\Delta + \Gamma^2}{2\eta} + 2B\Delta V_T^{R}\notag\\
\leq{}& \frac{2\eta K G^2T}{\sigma^2} + \frac{\Gamma^2}{2\eta} + 4\Gamma\sqrt{\frac{BTV_T^{R}}{\eta}} + 2BV_T^{R}\notag\\
\leq {}&\frac{2\eta K G^2T}{\sigma^2} + \frac{\Gamma^2}{2\eta} + 4\Gamma\sqrt{\frac{BTV_T^{R}}{\eta}} + 4B\sqrt[3]{T^2V_T^{R}}\notag\\
\leq {}& \big(\frac{2 K G^2}{\sigma^2}+2B^2\big)\eta T + \frac{\Gamma^2}{\eta} + 4(\Gamma+1)\sqrt{\frac{BTV_T^{R}}{\eta}},
\end{align}
where the second inequality is due to the parameter setting $\Delta = \lceil\sqrt{{\Gamma^2 T}/({\eta B V_T^R})}\rceil$. The third inequality is due to the fact that $V_T^R\leq 2T$. The last inequality holds by the AM-GM inequality $3B\sqrt[3]{T^2V_T^R}\leq 2\sqrt{\frac{BTV_T^R}{\eta}} +B^2\eta T $. 

Inspired by~\citet[Proposition 1]{AISTATS'22:OnlineContinualAdaptation}, we know that the bounded distance between probability distributions can lead to the temporal variability condition.
In the OLaS problem, with the label shift condition, we can further bound the variation of the loss function by the variation of the class prior $V_T = \sum_{t=2}^T \norm{\bm{\mu}_{y_t} - \bm{\mu}_{y_{t-1}}}_1$:
\begin{align}
    \label{eq:variation-bound}
V_T^R \triangleq{}& \sum_{t=2}^T \sup_{\w\in\W}\vert R_t(\w)-R_{t-1}(\w)\vert\notag\\
 ={}& \sum_{t=2}^T \sup_{\w\in\W} \left\vert\sum_{k=1}^K ([\bm{\mu}_{y_t}]_k-[\bm{\mu}_{y_{t-1}}]_k)R_0^k(\w)\right\vert\notag\\
\leq{}& \sum_{t=2}^T B\sum_{k=1}^K \left\vert[\bm{\mu}_{y_t}]_k-[\bm{\mu}_{y_{t-1}}]_k\right\vert\notag\\
={}& B \sum_{t=2}^T \norm{\bm{\mu}_{y_t}-\bm{\mu}_{y_{t-1}}}_1 \\
\triangleq{}& B V_T.
\end{align}

Plug~\eqref{eq:variation-bound} into~\eqref{eq:dynamic-UOGD-vtf} of Lemma~\ref{lemma:UOGD}, we have 
$$
\E\left[\DReg\right] \leq \big(\frac{2 K G^2}{\sigma^2}+2B^2\big)\eta T + \frac{\Gamma^2}{\eta} + 4(\Gamma+1)\sqrt{\frac{BTV_T}{\eta}},
$$
which completes the proof.

\end{proof}

\subsection{Proof of Theorem~\ref{thm:overall-UOGD}}
\label{appendix:sec-proof-overall-UOGD}

\begin{proof}[Proof of Theorem~\ref{thm:overall-UOGD}]
We can decompose the dynamic regret into two parts with the piecewise-stationary sequence $\{\u_t\}_{t=1}^T$ introduced in the proof of Theorem~\ref{thm:UOGD}.
\begin{align*}
&\E_{1:T}\left[\sum_{t=1}^T R_t(\w_{t})\right] - \sum_{t=1}^{T} R_t(\w^*_t)\\
={}&\E_{1:T}\left[\sum_{t=1}^T R_t(\w_{t})\right] -\sum_{t=1}^{T} R_t(\u_t) +\sum_{t=1}^{T} R_t(\u_t) - \sum_{t=1}^{T} R_t(\w^*_t)\\
={}&\underbrace{\E_{1:T}\left[\sum_{t=1}^T R_t(\w_{t})\right] - \sum_{m=1}^{M}\sum_{t\in\mathcal{I}_m} R_t(\w_{\mathcal{I}_m}^*)}_{\texttt{term~(a)}} + \underbrace{\sum_{m=1}^M\sum_{t\in\mathcal{I}_m} R_t(\w_{\mathcal{I}_m}^*) - \sum_{t=1}^{T} R_t(\w^*_t)}_{\texttt{term~(b)}},
\end{align*}
where term (b) can be bounded by 
\begin{align*}
\mathtt{term~(b)}\leq 2B\Delta V_T
\end{align*}
following the same arguments in deriving~\eqref{eq:proof-theorem1-V_T} and~\eqref{eq:variation-bound}. So, we only need to focus on the analysis of term (a), the regret of the model sequence returned by \textsc{Atlas} algorithm to the piecewise-stationary compactors. 

For any $i$-th base UOGD algorithm, we can further decompose term (a) into the meta-regret and the base-regret
\begin{align}
&\E_{1:T}\left[\sum_{t=1}^T R_t(\w_{t})\right] - \sum_{t=1}^T R_t(\u_t)\notag\\
={}& \underbrace{\E_{1:T}\left[\sum_{t=1}^T R_t(\w_{t}) - \sum_{t=1}^T R_t(\w_{t,i})\right]}_{\texttt{meta-regret}} +  \underbrace{\E_{1:T}\left[\sum_{t=1}^T R_t(\w_{t,i})\right] - \sum_{t=1}^T R_t(\u_t)}_{\texttt{base-regret}},
\label{eq:altas-decompose-meta-base}
\end{align}
where the meta-regret measures the performance gap between the model returned by \textsc{Atlas} algorithm and that by the $i$-th base UOGD algorithm, and the base-regret is the regret of the $i$-th base UOGD algorithm running with step size $\eta_i>0$. 

\paragraph{Analysis for meta-regret.} For the meta-regret, we have
\begin{align}
    &\E_{1:T}\left[\sum_{t=1}^T R_t(\w_t) - \sum_{t=1}^TR_t(\w_{t,i})\right]\notag\\
    \leq{}& \E_{1:T}\left[\sum_{t=1}^T\sum_{j=1}^N p_{t,j} R_t(\w_{t,j}) - \sum_{t=1}^TR_t(\w_{t,i})\right]\notag\\
    ={}&\E_{1:T}\left[\sum_{t=1}^T\sum_{j=1}^N p_{t,j} \Rh_t(\w_{t,j}) - \sum_{t=1}^T\Rh_t(\w_{t,i})\right] \notag\\
    & + \E_{1:T}\left[\sum_{t=1}^T\sum_{j=1}^N p_{t,j} (R_t(\w_{t,j})-\Rh_t(\w_{t,j})) + \sum_{t=1}^T(R_t(\w_{t,i})-\Rh_t(\w_{t,i}))\right]\notag\\
    = {}&\E_{1:T}\left[\sum_{t=1}^T\sum_{j=1}^N p_{t,j} \Rh_t(\w_{t,j}) - \sum_{t=1}^T\Rh_t(\w_{t,i})\right]\label{eq:proof-meta-regret-1},
\end{align}
where the first inequality is due to the Jensen's inequality and the last equality is due to the unbiasedness of the risk estimator such that $\E_t[\hat{R}_t(\w_{t,i})] = R_t(\w_{t,i})$ for any $i\in[N]$. Then, we can upper bound the meta-regret by the following lemma, whose proof is deferred to Appendix~\ref{appendix:sec-proof-overall-UOGD-lemmas}. 

\begin{myLemma}
\label{lemma:Hedge}
By setting the learning rate $\varepsilon = \frac{\sigma}{B}\sqrt{\frac{\ln N +2}{ KT}}$, the meta-algorithm of \textsc{Atlas} (Algorithm~\ref{alg:atlas-meta-hedge}) satisfies
\begin{align*}
\sum_{t=1}^T\sum_{j=1}^N p_{t,j} \Rh_t(\w_{t,j}) - \sum_{t=1}^T\Rh_t(\w_{t,i})\leq \frac{2B}{\sigma}\sqrt{(\ln N+2)KT}
\end{align*}
for any $i\in[N]$, where $B \triangleq \sup_{(\x,y)\in\X\times\Y,\w\in\W} \vert \ell(f(\w,\x),y)\vert$ is defined as the upper bound of the loss function.
\end{myLemma}

As a consequence of Lemma~\ref{lemma:Hedge}, we can upper bound the meta-regret as
\begin{align*}
\texttt{meta-regret} \leq \frac{2B}{\sigma} \sqrt{(\ln N +2) KT}.
\end{align*}

\paragraph{Analysis for base-regret.} Since the base-algorithm of \textsc{Atlas} algorithm (Algorithm~\ref{alg:atlas-base-hedge}) is taken as the UOGD algorithm, following a similar argument for obtaining Lemma~\ref{lemma:OGD-P_T}, we have
\begin{align*}
\E_{1:T}\left[\sum_{t=1}^T R_t(\w_{t,i})\right] - \sum_{t=1}^T R_t(\u_t)\leq \frac{2\eta_i K G^2T}{\sigma^2} + \frac{2\Gamma P_T + \Gamma^2}{2\eta_i}
\end{align*}
for any base-algorithm with the index $i\in[N]$, where $P_T$ is the variation of the comparator sequence $P_T = \sum_{t=2}^T \norm{\u_t-\u_{t-1}}_2$. Then, by combining the meta-regret and the base-regret, we can obtain that 
\begin{align*}
\texttt{term~(a)} ={}& \E_{1:T}\left[\sum_{t=1}^T (\w_t)\right] - \sum_{t=1}^T R_t(\u_t)\\
\leq{}& \frac{2B}{\sigma} \sqrt{(\ln N +2) KT} + \frac{2\eta_i K G^2T}{\sigma^2} + \frac{2\Gamma P_T + \Gamma^2}{2\eta_i}
\end{align*}
for any base-algorithm $i\in[N]$. 

It remains to identify the optimal step size $\eta_{i_*}$ to make the bound tight. Due to the construction of the step size pool\
\begin{align*}
\H =\left\{ \frac{\Gamma\sigma}{2G\sqrt{KT}}\cdot 2^{i-1} \mid i\in[N]\right\}
\end{align*}
with $N = 1+ \lceil \frac{1}{2}\log_2 (1 +2T) \rceil$, we can make sure that the optimal step size $\eta^* =\frac{\sigma}{2G}\sqrt{\frac{\Gamma^2+2\Gamma P_T}{KT}}$ is covered by the step size pool. Even better, due to the logarithmic construction of the pool, we claim that there must exist an $i_*\in[N]$ satisfies that $\eta_{i_*}/2\leq\eta^*\leq \eta_{i_*}$. So, we further bound term (a) as 
\begin{align*}
\texttt{term~(a)} \leq \frac{2B}{\sigma} \sqrt{(\ln N +2) KT} + \frac{3G}{\sigma}\sqrt{KT(2\Gamma P_T+\Gamma^2)}
\end{align*}
by comparing with the $i_*$-th base-algorithm in the decomposition~\eqref{eq:altas-decompose-meta-base}.

\paragraph{Overall dynamic regret bound.}
Combining term~(a) and term~(b) and noticing that $\{\u_t\}_{t=1}^T$ is the piecewise stationary comparator sequences with $P_T = \sum_{t=2}^T\norm{\u_t-\u_{t-1}}_{t=1}^T\leq \Gamma T/\Delta$, we can bound the overall dynamic regret as
\begin{align*}
\DReg  = {}&\E_{1:T}\left[\sum_{t=1}^T R_t(\w_{t})\right] - \sum_{t=1}^{T} R_t(\w^*_t)\\
\leq{}& 2B\Delta V_T + \frac{3G\Gamma}{\sigma}\sqrt{K T\left(1+\frac{2 T}{\Delta}\right)} + \frac{2B}{\sigma} \sqrt{(\ln N +2) KT} \\
\leq{}& 2B\Delta V_T + \frac{3G\Gamma T}{\sigma}\sqrt{\frac{2K}{\Delta}}+ \frac{3G\Gamma}{\sigma}\sqrt{K T} + \frac{2B}{\sigma} \sqrt{(\ln N +2) KT},
\end{align*}
where the last inequality is due to the fact that $\sqrt{a+b}\leq \sqrt{a} + \sqrt{b}$ for any $a,b\geq0$. We can complete the proof by choosing $\Delta = \lceil(\frac{3GT\Gamma \sqrt{2K}}{4\sigma BV_T})^{\frac{2}{3}}\rceil$ and obtain the bound as 
\begin{align*}
\DReg\leq{}&3 \left(\frac{9KBG^2\Gamma^2}{\sigma^2}\right)^{\frac{1}{3}}\cdot{V_T^{1/3}T^{2/3}} + \frac{3\sqrt{K}(G\sqrt{\Gamma}+B\sqrt{\ln N+2})}{\sigma}\sqrt{T} + 2BV_T \\
\leq{}& 3\left(\frac{9KBG^2\Gamma^2}{\sigma^2}\right)^{\frac{1}{3}}\cdot{V_T^{1/3}T^{2/3}} + \frac{3\sqrt{K}(G\sqrt{\Gamma}+B\sqrt{\ln N+2})}{\sigma}\sqrt{T} + 4BV_T^{1/3}T^{2/3},
\end{align*}
where the last inequality is due to the fact that $V_T\leq 2T$.
\end{proof}

\subsection{Discussion on Minimax Optimality}
\label{appendix:discuss-optimal}
For online convex optimization with a general convex function $R_t:\W\mapsto \R$,~\citet[Theorem 2]{OR'15:dynamic-function-VT} established an $\Omega\left((V_T^R)^{1/3} T^{2/3}\right)$ lower bound for the dynamic regret $\sum_{t=1}^T\E[R_t(\w_t)]-\sum_{t=1}^T \inf_{\w\in\W}R_t(\w)$ when the learner can only observe a noisy estimate of the full information function $R_t$. In the above, the quantity $V_T^R = \sum_{t=2}^T\sup_{\w\in\W} \vert R_t(\w)-R_{t-1}(\w)\vert$ measures the non-stationarity of the environments by the variation of the loss functions. In addition to the $\O(V_T^{1/3}T^{2/3})$ upper bound stated in Theorem~\ref{thm:UOGD}, we can show that the UOGD algorithm can actually achieve an $\O\left((V_T^{R})^{1/3}T^{2/3}\right)$ dynamic regret bound based on the following lemmas.

\begin{myLemma}
    \label{lemma:UOGD}
    Under the same assumptions as Lemma~\ref{lemma:unbiased-estimator}, UOGD in~\pref{eq:UOGD} with step size $\eta$ satisfies
    \begin{align}
    \label{eq:dynamic-UOGD-vtf}
    \E\big[\DReg\big] {}&\leq 2(\frac{KG^2}{\sigma^2}+B^2)\eta T + \frac{\Gamma^2}{\eta} + 4(\Gamma+1)\sqrt{\frac{B V_T^R T}{\eta}} \notag\\
    {}&= \O\Big( \eta T + 1/\eta + \sqrt{(V_T^R T)/\eta} \Big),
    {}\end{align}
    where the constant $\sigma > 0 $ denotes the minimum singular value of the invertible confusion matrix $C_{f_0}$. Moreover, $V_T^R \triangleq \sum_{t=2}^T \sup_{\w\in\W }\left\vert R_t(\w) - R_{t-1}(\w)\right\vert$ measures the variation of the loss function, usually known as the temporal variability or function variation in the online learning literature. 
\end{myLemma}
\begin{proof}[Proof of Lemma \ref{lemma:UOGD}]
This lemma is actually an intermediate result~\eqref{uogd-vt-bound} in the proof of Theorem~\ref{thm:UOGD}.
\end{proof}
By setting the step size in~\eqref{eq:dynamic-UOGD-vtf} optimally as $\eta=\Theta\big(T^{-\frac{1}{3}} (V_T^R)^{\frac{1}{3}}\big)$, UOGD enjoys an $\O\big((V_T^R)^{\frac{1}{3}} T^{\frac{2}{3}}\big)$ dynamic regret. Such a rate matches the $\Omega\big((V_T^R)^{1/3} T^{2/3}\big)$ lower bound~\citep{OR'15:dynamic-function-VT} and thus demonstrates the minimax optimality of the UOGD algorithm in the non-degenerated cases with the noisy feedback function. Moreover, we can use a similar argument to show that our \textsc{Atlas} also enjoys an $\O(\max\{(V_T^R)^{\frac{1}{3}}T^{\frac{2}{3}},\sqrt{T}\})$ dynamic regret guarantee in addition to the $\O(\max\{\V_T^{\frac{1}{3}}T^{\frac{2}{3}},\sqrt{T}\})$ bound exhibited in Theorem~\ref{thm:overall-UOGD}. Finally, we mention that although our algorithms are minimax optimal in terms of $T$ and $V_T^R$, while it remains unclear about the tightness of the dynamic regret bounds scaling with the variation of class priors $V_T$, which is left as future work to investigate.

We also make another remark on the situation when the online function observed by the learner is noiseless, an even better dynamic regret upper bound of order $\O(V_T)$ is achievable in the OCO setting~\citep{L4DC'21:sc_smooth}, see more discussions on the worst-case dynamic regret in Appendix~\ref{sec-appendix:related-work-oco}. Note that such a rate is also attainable for our problem under the assumption that the distribution $\D_t(\x)$ is \emph{available} at every iteration, however, this assumption can hardly be true in practice since the data samples at the online adaptation stage are usually very few. Thus, we focus on the non-degenerated case where the noise exists due to the sampling from the unknown data distribution.

\setcounter{myLemma}{5}
\subsection{Useful Lemmas in the Analysis of Theorem~\ref{thm:UOGD} and Theorem~\ref{thm:overall-UOGD}}
\label{appendix:sec-proof-overall-UOGD-lemmas}
This section presents several useful lemmas that are omitted in the proof of Theorem~\ref{thm:UOGD} and Theorem~\ref{thm:overall-UOGD}.

\begin{myLemma}
Under same assumptions of Theorem~\ref{thm:UOGD}, UOGD in~\pref{eq:UOGD} with a step size $\eta > 0$ satisfies
\begin{align*}
\sum_{t=1}^T\inner{\nabla \hat{R}_t(\w_t)}{\w_t-\u_t} \leq \frac{2\eta K G^2T}{\sigma^2} + \frac{2\Gamma P_T + \Gamma^2}{2\eta} 
\end{align*}
for any comparator sequence $\{\u_t\}_{t=1}^T$ with $\u_{t}\in\W$. Moreover, $P_T = \sum_{t=2}^T \norm{\u_t-\u_{t-1}}_2$ measures the variation of the comparator sequence.
\end{myLemma}

\begin{proof}[Proof of Lemma~\ref{lemma:OGD-P_T}] 
We can decompose the L.H.S. of the inequality in Lemma~\ref{lemma:OGD-P_T} as
\begin{align*}
\inner{\nabla \Rh_t(\w_{t})}{\w_{t}-\u_t} =\underbrace{\inner{\nabla \Rh_t(\w_t)}{\w_{t}-\w_{t+1}} }_{\texttt{term~(a)}}  + \underbrace{\inner{\nabla \Rh_t(\w_{t})}{\w_{t+1}-\u_t}}_{\texttt{term~(b)}} .
\end{align*}

For term (a), we have
\begin{align*}
    \texttt{term~(a)} ={}& \inner{\nabla \Rh_t(\w_t)}{\w_{t}-\w_{t+1}} \\ 
    \leq{}&\norm{\nabla \Rh_t(\w_t)}_2\norm{\w_{t}-\w_{t+1}}_2\\
    \leq{}&2\eta\norm{\nabla \Rh_t(\w_t)}_2^2 + \frac{1}{2\eta}\norm{\w_{t}-\w_{t+1}}_2^2,
\end{align*}
where the first inequality is due to the Cauchy-Schwarz inequality and the second inequality comes from the AM-GM inequality.
Then, we bound term (b) by~\citet[Lemma 7]{NIPS'20:sword},
\begin{align*}
    \texttt{term~(b)}\leq\frac{1}{2\eta}\left(\norm{\w_{t} - \u_t}_2^2 - \norm{\w_{t+1} - \u_t}_2^2 - \norm{\w_{t} - \w_{t+1}}_2^2 \right).
\end{align*}
Combining the two terms and taking the summation over $T$ iterations, we have 
\begin{align}
    &\sum_{t=1}^T\inner{\nabla \Rh_t(\w_{t})}{\w_{t}-\u_t}\notag\\
    \leq{}& 2\eta\sum_{t=1}^T\norm{\nabla \Rh_t(\w_{t})}_2^2 + \frac{1}{2\eta}\sum_{t=1}^T\left(\norm{\w_{t}-\u_t}_2^2 - \norm{\w_{t+1}-\u_t}_2^2\right)\notag\\
    \leq{}& 2\eta\sum_{t=1}^T\norm{\nabla \Rh_t(\w_{t})}_2^2 + \frac{1}{2\eta}\sum_{t=2}^T\left(\norm{\w_{t}-\u_{t}}_2^2 - \norm{\w_{t}-\u_{t-1}}_2^2\right)  + \frac{\Gamma^2}{2\eta}\notag\\    
    \leq {}&2\eta\sum_{t=1}^T\norm{\nabla \Rh_t(\w_{t})}_2^2 + \frac{1}{2\eta}\sum_{t=2}^T(\norm{\w_t-\u_t}_2+\norm{\w_t-\u_{t-1}}_2)\cdot\norm{\u_t-\u_{t-1}}_2+\frac{\Gamma^2}{2\eta} \notag\\
    \leq {}&2\eta\sum_{t=1}^T\norm{\nabla \Rh_t(\w_{t})}_2^2 + \frac{\Gamma}{\eta}\sum_{t=2}^T\norm{\u_t-\u_{t-1}}_2+\frac{\Gamma^2}{2\eta} \notag\\
    ={}& 2\eta\sum_{t=1}^T\norm{\nabla \Rh_t(\w_{t})}_2^2 + \frac{2\Gamma P_T + \Gamma^2}{2\eta} \label{eq:proof-lemma-OGD-A2},
\end{align}
where we denote by $P_t = \sum_{t=2}^T\norm{\u_t-\u_{t-1}}_2$ the variation of the comparator sequence $\{\u_t\}_{t=1}^T$. The third inequality is due to the fact $\norm{a}_2^2 - \norm{b}_2^2\leq (\norm{a}_2+\norm{b}_2)\cdot\norm{a-b}_2$ for any $a,b\in\R^d$. 

We can further bound the first term of~\eqref{eq:proof-lemma-OGD-A2} by
\begin{align}
2\eta \sum_{t=1}^T\norm{\nabla \Rh_t(\w_{t})}_2^2\overset{\eqref{eq:unbiased-risk-estimator}}{=}{}&2\eta\sum_{t=1}^T\left\Vert\sum_{k=1}^K [C_{f_0}^{-1}\hat{\bm{\mu}}_{\hat{y}_t}]_k\cdot {\nabla R}_0^k(\w_t)\right\Vert_2^2\notag\\
\leq{}& 2\eta \sum_{t=1}^T \left(\sum_{k=1}^K \left\vert[ C_{f_0}^{-1}\hat{\bm{\mu}}_{\hat{y}_t}]_k\right\vert \norm{\nabla R_0^k(\w_t)}_2\right)^2\notag\\
\leq{}& 2\eta G^2  \sum_{t=1}^T \left\Vert C_{f_0}^{-1}\hat{\bm{\mu}}_{\hat{y}_t}\right\Vert_1^2\notag\\
\leq{}& 2\eta KG^2  \sum_{t=1}^T \left\Vert C_{f_0}^{-1}\hat{\bm{\mu}}_{\hat{y}_t}\right\Vert_2^2\notag\\
\leq{}& 2\eta KG^2  \sum_{t=1}^T \left\Vert C_{f_0}^{-1}\right\Vert_2^2\cdot\left\Vert\hat{\bm{\mu}}_{\hat{y}_t}\right\Vert_2^2\notag\\
\leq{}& 2\eta KG^2  \sum_{t=1}^T \norm{C_{f_0}^{-1}}_2^2\leq \frac{2\eta KG^2T}{\sigma^2}\label{eq:proof-lemma-OGD-A4},
\end{align}
where $\sigma$ is the minimum singular value of the matrix $C_{f_0}$. In above, the second inequality is due to the definition $R_0^k(\w) = \E_{\x\sim \D_0\left(\x \givenn y=k \right)}[\ell(f(\w,\x),k)]$ and the norm of gradient of the loss function $\norm{\nabla_\w \ell(f(\w,\x),y)}_2$ is bounded by $G$ for any $\x\in\X$ and $y\in[K]$. The third inequality is a consequence of the fact that $\norm{\x}_1\leq \sqrt{\x}_2$ for any $\x\in\R^K$. The last second inequality hold since $\hat{\bm{\mu}}_{\hat{y}_t}\in\Delta_{K-1}$ and then we have $\norm{\hat{\bm{\mu}}_{\hat{y}_t}}_2^2\leq 1$.
We complete the proof by plugging~\eqref{eq:proof-lemma-OGD-A4} into~\eqref{eq:proof-lemma-OGD-A2}.
\end{proof}

Then, we provide the proof of Lemma~\ref{lemma:Hedge}, which provide an upper bound for the meta-algorithm of the \textsc{Atlas} algorithm.

\begin{myLemma}
By setting the learning rate $\varepsilon = \frac{\sigma}{B}\sqrt{\frac{\ln N +2}{ KT}}$, the meta-algorithm of \textsc{Atlas} (Algorithm~\ref{alg:atlas-meta-hedge}) satisfies
\begin{align*}
\sum_{t=1}^T\sum_{j=1}^N p_{t,j} \Rh_t(\w_{t,j}) - \sum_{t=1}^T\Rh_t(\w_{t,i})\leq \frac{2B}{\sigma}\sqrt{(\ln N+2)KT}
\end{align*}
for any $i\in[N]$, where $B \triangleq \sup_{(\x,y)\in\X\times\Y,\w\in\W} \vert \ell(f(\w,\x),y)\vert$ is defined as the upper bound of the loss function.
\end{myLemma}

\begin{proof}
Since \textsc{Atlas} takes the Hedge algorithm as the meta-algorithm, we can exploit the standard analysis to upper bound the meta-regret. According to the regret guarantee of the Hedge algorithm (Lemma~\ref{lemma: Hedge-NIPS-15} in Appendix~\ref{sec:appendix-lemma}) and setting $\mathbf{m}_t = 0$, we can bound the meta-regret bound by
\begin{align}
\label{eq:proof-lemma-hedge-1}
\sum_{t=1}^T\sum_{j=1}^N p_{t,j} \Rh_t(\w_{t,j}) - \sum_{t=1}^T\Rh_t(\w_{t,i})\leq \frac{\ln N+2}{\epsilon} + \epsilon \sum_{t=1}^T \vert\max_{i\in[N]}\{\hat{R}(\w_{t,i})\}\vert^2,
\end{align}
where the last term can be further bounded by 
\begin{align}
    \vert\Rh_t(\w_{t,i})\vert ={}& \left\vert\sum_{k=1}^K [C^{-1}_{f_0}\hat{\bm{\mu}}_{\hat{y}_t}]_k\cdot R_0^k(\w_{t,i})\right\vert\notag\\
    \leq {}& B\norm{C^{-1}_{f_0}\hat{\bm{\mu}}_{\hat{y}_t}}_1 \leq B\sqrt{K}\norm{C^{-1}_{f_0}\hat{\bm{\mu}}_{\hat{y}_t}}_2\notag\\
    \leq {}&B\sqrt{K}\norm{C^{-1}_{f_0}}_2\cdot \norm{\hat{\bm{\mu}}_{\hat{y}_t}}_2\leq \frac{B\sqrt{K}}{\sigma}, \label{eq:proof-lemma-hedge-2}
\end{align}
where the third inequality is due to the Cauchy–Schwarz inequality. The last inequality comes from the fact that $\norm{C^{-1}_{f_0}}_2\leq \sigma^{-1}$ and $\norm{\hat{\bm{\mu}}_{\hat{y}_t}}_2\leq 1$. Plugging~\eqref{eq:proof-lemma-hedge-2} into~\eqref{eq:proof-lemma-hedge-1} yields 
\begin{align*}
\sum_{t=1}^T\sum_{j=1}^N p_{t,j} \Rh_t(\w_{t,j}) - \sum_{t=1}^T\Rh_t(\w_{t,i})\leq \frac{\ln N+2}{\epsilon} + \epsilon \sum_{t=1}^T \frac{B^2KT}{\sigma^2}.
\end{align*}
We complete the proof by setting $\epsilon = \frac{\sigma}{B}\sqrt{\frac{(\ln N +2)}{KT}}$.
\end{proof}

\section{Omitted Details for Section~\ref{sec:more-adaptive}}
\label{sec-appendix:proof-sec3.3}
In this section, we first introduce the omitted algorithm details for the \textsc{Atlas-Ada} algorithm in Appendix~\ref{sec-appendix:atala-ada-algorithm}. Then, we present the proof of the Theorem~\ref{thm:overall-guarantees} (Appendix~\ref{appendix:sec-proof-Atlas-ada}), followed by the useful lemmas in the regret analysis.

\subsection{Algorithm Details for \textsc{Atlas-Ada}}
\label{sec-appendix:atala-ada-algorithm}
This part provides algorithm details of the \textsc{Atlas-Ada}. The main procedures are presented in Algorithm~\ref{alg:atlas-ada-base} (base-algorithm) and Algorithm~\ref{alg:atlas-ada-meta} (meta-algorithm).

\paragraph{Hint Functions.} 
As shown in Section~\ref{eq:upper-bound-optimism}, a guidance for designing the hint function is to approximate the true priors $\bm{\mu}_{y_t}$ with hint priors $\bm{h}_{y_t}$. 
A direct way is to use current data $S_t$ to estimate the hint priors, which we call \emph{Forward Hint}.
However, when the sample size $|S_t|$ is small at each round, the estimated priors can be far from the true priors, which motivates us to reuse the historical information. The intuition is that we can benefit from the historical data if there are some regular shift patterns. To see this, we design three additional hint functions. Details are listed in the following. 

\begin{compactitem}
    \item \textbf{Forward Hint}. We perform transductive learning with the current data $S_t$ by
    $\bm{h}_{y_t} = \bm{\hat{\mu}}_{y_t}$.

    \item \textbf{Window Hint}. With the belief that recent data are useful, we set a sliding window size of $L_{m}$ and use the average of the estimated priors in the window as the hint priors: $$\bm{h}_{y_t} = \frac{1}{L_m} \sum_{s=t-L_m}^{t-1} \bm{\hat{\mu}}_{y_s}.$$
    
    \item \textbf{Periodic Hint}. For a periodic situation where a similar pattern recurs at a specific time interval, we consider reusing the historical data in a periodic manner by maintaining a buffer $\{\bm{\bar{\mu}}_i \mid i\in[L_p]\}$ with the size of $L_p$. At round $t$, we utilize the estimated prior $\bm{\hat{\mu}}_{y_t}$ to update the priors in the buffer: 
    $$
    \bm{\bar{\mu}}_i^{(t)} = \begin{cases}
    \bm{\bar{\mu}}_i^{(t-1)} + \lambda_t \left(\bm{\hat{\mu}}_{y_t} - \bm{\bar{\mu}}_i^{(t-1)}\right) & i = t\bmod L_p \\
    \bm{\bar{\mu}}_i^{(t-1)} & \text{otherwise}
    \end{cases},
    $$
    where $i \in [L_p]$ and $\bm{\bar{\mu}}_i^{(t)}$ denotes the value of $\bm{\bar{\mu}}_i$ at round $t$. $\lambda_t \in [0, 1]$ is a coefficient to balance the historical and present information, which can be set to be $1/t$ or a constant value.
    Then we set $\bm{h}_{y_t} =\bm{\bar{\mu}}_{i_t}^{(t)}$, where $i_{t} = t\bmod L_p$.
    \item \textbf{Online KMeans Hint}. To further use the feature similarity of the historical data, we design a hint function with the online k-means clustering algorithms~\citep{DBLP:journals/ijns/BarbakhF08}. 
    We maintain $L_k$ prototypes $\{(\Bar{\x}_i, \Bar{\bm{\mu}}_i) \mid i\in[L_k]\}$. At round $t$, we find the prototype $i_t$, which is the most similar to the current unlabeled data $S_t$:
    $$
    i_t = \argmin_{i\in[L_k]} \frac{1}{|S_t|} \sum_{\x \in S_t}\|\Bar{\x}_{i} - \x\|^2,
    $$
    and then we update the corresponding prototype by
    $$
    \Bar{\x}_{i_t}^{(t)} = \Bar{\x}_{i_t}^{(t-1)} + \kappa_t\left[\left(\frac{1}{|S_t|} \sum_{\x \in S_t}\x\right) - \Bar{\x}_{i_t}^{(t-1)}\right],
    $$
    $$
    \Bar{\bm{\mu}}_{i_t}^{(t)} = \bm{\bar{\mu}}_{i_t}^{(t-1)} + \kappa_t \left(\bm{\hat{\mu}}_{y_t} - \bm{\bar{\mu}}_{i_t}^{(t-1)}\right),
    $$
    where $\left( \Bar{\x}_{i_t}^{(t)}, \Bar{\bm{\mu}}_{i_t}^{(t)}\right)$ represents the value of $i_t$-th prototype at round $t$. $\kappa_t \in [0, 1]$ is a coefficient to balance the historical and present information, which can be set to be $1/t$ or a constant value.
    Then we set $\bm{h}_{y_t} =\bm{\bar{\mu}}_{i_t}^{(t)}$.
\end{compactitem}

\begin{figure}[!t]
    \begin{minipage}{0.49\textwidth}
    \begin{algorithm}[H]
       \caption{\textsc{Atlas-ada}: base-algorithm}
       \label{alg:atlas-ada-base}
    \begin{algorithmic}[1]
      \REQUIRE{step size $\eta_i \in \H$}
      \STATE{Let $\w_{1,i}$ be any point in $\W$}
        \FOR{$t=2$ {\bfseries to} $T$}
          \STATE construct the risk estimator $\hat{R}_{t-1}$ as~\eqref{eq:unbiased-risk-estimator}
          \STATE update the model of base-learner by \vspace{-1mm}
            \begin{flalign*}
            &\resizebox{0.85\hsize}{!}{$\wh_{t,i} = \mathop{\Pi}\nolimits_{\W}[\wh_{t-1,i} - \eta_i \nabla \Rh_{t-1}(\w_{t-1,i})]$}\\
            &\resizebox{0.85\hsize}{!}{$\w_{t,i} = \mathop{\argmin}\limits_{\w\in\W}  H_t(\w) + \frac{1}{2\eta_i}\norm{\w-\hat{\w}_{t,i}}_2^2.$} &
            \end{flalign*}\vspace{-1mm}
          \STATE send $\w_{t,i}$ to the meta-algorithm
        \ENDFOR
    \end{algorithmic}
    \end{algorithm}
    \end{minipage}
    \hfill
    \begin{minipage}{0.49\textwidth}
    \begin{algorithm}[H]
       \caption{\textsc{Atlas-ada}: meta-algorithm}
       \label{alg:atlas-ada-meta}
       \begin{spacing}{1.11}
        \begin{algorithmic}[1]
          \REQUIRE{step size pool $\H$; learning rate $\varepsilon$}
          \STATE{initialization: $\forall i\in [N], p_{1,i} = 1/N$}
            \FOR{$t=2$ {\bfseries to} $T$}
              \STATE  receive $\{\w_{t,i}\}_{i=1}^N$ from base-learners
              \STATE update weight $\p_t \in \Delta_N$ according to \vspace{-1mm}
              \begin{flalign*}
              \resizebox{0.85\hsize}{!}{$p_{t, i} \propto \exp\Big({ -\varepsilon\big(\sum\limits_{s=1}^{t-1}\Rh_s(\w_{s,i})+H_{t}(\w_{t,i})\big)}\Big)$} &&
              \end{flalign*}\vspace{-1mm}
              \STATE predict final output $\w_{t} = \sum_{i=1}^{N} p_{t,i} \w_{t,i}$
            \ENDFOR
        \end{algorithmic}
       \end{spacing}
    \end{algorithm}
    \end{minipage}
\end{figure}

\subsection{Proof of Theorem~\ref{thm:overall-guarantees}}
\label{appendix:sec-proof-Atlas-ada}
The proof of Theorem~\ref{thm:overall-guarantees} shares a similar spirit with that of Theorem~\ref{thm:overall-UOGD}, where we decompose the overall dynamic regret into meta-regret and base-regret. The new challenge is that the \textsc{Atlas-Ada} algorithm involves the hint function $H_t(\cdot)$ to exploit the structure of distribution change, which requires a more delicate analysis for both the meta- and base-regret.

\begin{proof}[Proof of Theorem~\ref{thm:overall-guarantees}]
Similar to the proof of Theorem~\ref{thm:UOGD} and Theorem~\ref{thm:overall-UOGD}, we can decompose the dynamic regret into two parts with the piecewise-stationary sequence $\{\u_t\}_{t=1}^T$,
\begin{align*}
&\E_{1:T}\left[\sum_{t=1}^T R_t(\w_{t})\right] - \sum_{t=1}^{T} R_t(\w^*_t)\\
={}&\underbrace{\E_{1:T}\left[\sum_{t=1}^T R_t(\w_{t})\right] - \sum_{m=1}^{M}\sum_{t\in\mathcal{I}_m} R_t(\w_{\mathcal{I}_m}^*)}_{\texttt{term~(a)}} + \underbrace{\sum_{m=1}^M\sum_{t\in\mathcal{I}_m} R_t(\w_{\mathcal{I}_m}^*) - \sum_{t=1}^{T} R_t(\w^*_t)}_{\texttt{term~(b)}}.
\end{align*}
According to~\eqref{eq:proof-theorem1-V_T} and~\eqref{eq:variation-bound}, we have $\texttt{term~(b)}\leq 2B\Delta V_T$. Besides, when the parameter is set as $\Delta = 1$, we have $\texttt{term~(b)} = 0$. Thus, the term (b) can be bounded as
\begin{align}
    \label{eq:proof-thm3-termb}
    \texttt{term~(b)}\leq \indicator\{\Delta> 1\}\cdot 2B\Delta V_T.
\end{align}

As for term~(a), it can be further decomposed into the meta-regret and base-regret.
\begin{align*}
\texttt{term~(a)} = {}&\E_{1:T}\left[\sum_{t=1}^T R_t(\w_{t})\right] - \sum_{t=1}^T R_t(\u_t)\\
={}& \underbrace{\E_{1:T}\left[\sum_{t=1}^T R_t(\w_{t}) - \sum_{t=1}^T R_t(\w_{t,i})\right]}_{\texttt{meta-regret}} +  \underbrace{\E_{1:T}\left[\sum_{t=1}^T R_t(\w_{t,i})\right] - \sum_{t=1}^T R_t(\u_t)}_{\texttt{base-regret}},
\end{align*}
where meta-regret measures the performance gap between the model returned by \textsc{Atlas-Ada} and that by the $i$-th base algorithm. Base-regret is regret of the $i$-th base algorithm with step size $\eta_i>0$. 

\textbf{Analysis for the meta-regret.}~~~ 
Since $\w_{t}$ is independent of the sample observed at iteration $t$, following the same arguments in obtaining~\eqref{eq:proof-meta-regret-1}, we can bound the meta-regret as 
\begin{align*}
    &\E_{1:T}\left[\sum_{t=1}^T R_t(\w_t) - \sum_{t=1}^TR_t(\w_{t,i})\right]\leq \E_{1:T}\left[\sum_{t=1}^T\sum_{j=1}^N p_{t,j} \Rh_t(\w_{t,j}) - \sum_{t=1}^T\Rh_t(\w_{t,i})\right].
\end{align*}
\begin{myRemark}
The main difference between the analysis of the meta-algorithm for \textsc{Atlas-Ada} and that of \textsc{Atlas} is that the former method involves an update step with the hint function. To obtain a meta-regret bound that adapts the quality of the hint function, the previous work~\citep{NIPS'20:sword} crucially relies on the \emph{smoothness} of the loss function, which is generally hard to be satisfied in the \ols problem (e.g., the commonly used hinge loss is not smooth). A more delicate analysis shows that the smoothness assumption is no longer necessary to achieve an adaptive meta-regret bound. The result is stated as the following lemma, whose proof is deferred to Appendix~\ref{appendix:sec-proof-Atlas-ada-lemmas}.
\end{myRemark}

\begin{myLemma}
\label{lemma:Hedge-ada}
By setting the learning rate $\epsilon = \sqrt{\frac{\ln N+2}{ 1+ \sum_{t=1}^T(\max_{i\in[N]}\{\vert \tilde{R}_t(\w_{t,i}) - \tilde{H}_t(\w_{t,i})\vert\})^2}}$, the meta-algorithm of \textsc{Atlas-Ada} (Algorithm~\ref{alg:atlas-ada-meta}) satisfies
\begin{align*}
\sum_{t=1}^T\sum_{j=1}^N p_{t,j} \Rh_t(\w_{t,j}) - \sum_{t=1}^T\Rh_t(\w_{t,i})\leq  \Gamma\sqrt{(\ln N+2)\bigg(1+\sum_{t=1}^T \sup_{\w\in\W}\norm{\nabla \Rh_t(\w) - \nabla H_t(\w)}_2^2\bigg)}
\end{align*}
for any $i\in[N]$, where $\tilde{R}_{t}(\w_{t,i}) = \Rh_{t}(\w_{t,i}) - \Rh_t(\w_{\mathtt{ref}})$ and $\tilde{H}_{t}(\w_{t,i}) = H_t(\w_{t,i}) - H_t(\w_{\mathtt{ref}})$ are the reference loss functions built upon an arbitrary reference point $\w_{\mathtt{ref}}\in\W$.
\end{myLemma}
According to Lemma~\ref{lemma:Hedge-ada}, we can bound the meta-regret by 
\begin{align*}
    &\E_{1:T}\left[\sum_{t=1}^T R_t(\w_t) - \sum_{t=1}^TR_t(\w_{t,i})\right]\\
    \leq{}& \Gamma\E_{1:T}\left[ \sqrt{(\ln N+2)\sum_{t=1}^T \sup_{\w\in\W}\norm{\nabla \Rh_t(\w) - \nabla H_t(\w)}_2^2}\right]\\
    \leq{}& \Gamma \sqrt{(\ln N+2)\left(1+\sum_{t=1}^T \E_t\left[\sup_{\w\in\W}\norm{\nabla \Rh_t(\w) - \nabla H_t(\w)}_2^2\right)\right]},
\end{align*}
where the last inequality is due to the Jensen's inequality and the concavity of the square root function.

\textbf{Analysis for the base-regret.}~~~Since the model $\w_t$ is independent of the draw of dataset $S_t$ received at iteration $t$, following a similar argument in obtain~\eqref{eq:proof-thm1-terma}, we can bound the base-regret as 
\begin{align*}
\E_{1:T}\left[\sum_{t=1}^T R_t(\w_{t,i})\right] - \sum_{t=1}^T R_t(\u_t)\leq \E_{1:T}\left[\sum_{t=1}^T[\inner{\nabla \Rh_t(\w_{t,i})}{\w_{t,i}-\u_t}\right].
\end{align*}

Different from the analysis for the base-algorithm of $\textsc{Atlas}$, the base-algorithm of $\textsc{Atlas-ada}$ exploit a hint function $H_t(\cdot)$ to better exploit the pattern of the distribution challenge. We have the following regret guarantees on the output $\w_{t,i}$ of the base-algorithm.
\begin{myLemma}
\label{lemma:altas-base-regret-noexpectation}
Suppose the hint function $H_t: \W \mapsto \R$ is convex. The base-algorithm of \textsc{Atlas-Ada} (Algorithm~\ref{alg:atlas-ada-base}) with a step size $\eta_{t,i} > 0$ satisfies
\begin{align*}
\sum_{t=1}^T\inner{\nabla \Rh_t(\w_{t,i})}{\w_{t,i}-\u_t}\leq 2\eta_i\sum_{t=1}^T\sup_{\w\in\W}\norm{\nabla \Rh_t(\w)-\nabla H_t(\w)}_2^2 + \frac{\Gamma^2 + 2\Gamma P_T}{2\eta_i}
\end{align*}
for any comparator sequence $\{\u_t\}_{t=1}^T$ with $\u_{t}\in\W$. Moreover, $P_T = \sum_{t=2}^T \norm{\u_t-\u_{t-1}}_2$ measures the variation of the comparator sequence.
\end{myLemma}

By Lemma~\ref{lemma:altas-base-regret-noexpectation} and taking the expectation over both sides, we can bound the base-regret as
\begin{align*}
&\E_{1:T}\left[\sum_{t=1}^T R_t(\w_{t,i})\right] - \sum_{t=1}^T R_t(\u_t)\leq 2\eta_i\sum_{t=1}^T\E_t\left[\sup_{\w\in\W}\norm{\nabla \Rh_t(\w)-\nabla H_t(\w)}_2^2\right] + \frac{\Gamma^2 + 2\Gamma P_T}{2\eta_i}.
\end{align*}
Then, combining the meta- and base-regret, we can obtain that 
\begin{align*}
\texttt{term(a)}\leq 2\eta_iG_T + \frac{\Gamma^2 + 2\Gamma P_T}{2\eta_i} + \Gamma \sqrt{(\ln N+2) (1+G_T)}
\end{align*}
for any base-algorithm $i\in[N]$, where $G_T = \sum_{t=1}^T \E_t\left[\sup_{\w\in\W}\norm{\nabla \Rh_t(\w) - \nabla H_t(\w)}_2^2\right]$ measures the reusability of the history information.

It remains to tune the step size to make the bound tight. We can bound the term $G_t$ by
\begin{align*}
G_T = {}&\sum_{t=1}^T \E_t\left[\sup_{\w\in\W}\norm{\nabla \Rh_t(\w) - \nabla H_t(\w)}_2^2\right]\\
\leq{}& \sum_{t=1}^T \E_t\left[2\sup_{\w\in\W}\norm{\nabla \Rh_t(\w)}_2^2 + 2\sup_{\w\in\W}\norm{\nabla H_t(\w)}_2^2\right]\leq \frac{4T G^2K}{\sigma^2},
\end{align*}
where the last inequality is due to the fact that $\sup_{\w\in\W}\norm{\nabla \hat{R}_t(\w_t)}_2 \leq (G\sqrt{K})/\sigma$ (shown as~\eqref{eq:proof-lemma-OGD-A4}) and the assumption that $\sup_{\w\in\W}{\norm{\nabla H_t(\w)}_2}\leq (G\sqrt{K})/\sigma$. In such a case, we can show that the optimal step size $\eta_* = \frac{1}{2}\sqrt{\frac{\Gamma^2 + 2\Gamma P_T}{1+G_T}}$ lies in the range $[\frac{1}{2}\sqrt{\frac{\Gamma^2\sigma^2}{\sigma^2+4TG^2K}}, \frac{1}{2}\sqrt{\Gamma^2 + 2\Gamma^2 T}]$. So, we can construct the following step size pool to cover the optimal step size
\begin{align*}
\H =\left\{ \frac{\Gamma\sigma}{\sqrt{\sigma^2 + 4TG^2K}}\cdot 2^{i-1} \mid i\in[N]\right\}
\end{align*}
with $N = 1+ \left\lceil \frac{1}{2}\log_2 \big((1 +2T)(1+ 4TG^2K/\sigma^2)\big) \right\rceil$. Even better, due to the logarithmic construction of the pool, we claim that there must exist an $i_*\in[N]$ satisfies that $\eta_{i_*}/2\leq\eta_*\leq \eta_{i_*}$. So, we further bound term (a) as 
\begin{align*}
\texttt{term(a)}\leq 3\sqrt{(1+G_T)(\Gamma^2 + 2\Gamma P_T)} + \Gamma \sqrt{(\ln N+2) (1+G_T)}.
\end{align*}

\textbf{Overall dynamic regret bound.}~~~We can then use similar arguments in the proof of Theorem~\ref{thm:overall-UOGD} to obtain the final dynamic regret bound. Combining term~(a) and term~(b) and noticing that $\{\u_t\}_{t=1}^T$ is the piecewise stationary comparator sequences with $P_T = \sum_{t=2}^T\norm{\u_t-\u_{t-1}}_{t=1}^T\leq \Gamma T/\Delta$, we can bound the overall dynamic regret as
\begin{align*}
{}&\E_{1:T}\left[\DReg\right]  = \E_{1:T}\left[\sum_{t=1}^T R_t(\w_{t})\right] - \sum_{t=1}^{T} R_t(\w^*_t)\\
\leq {}& \indicator\{\Delta >1\}\cdot 2B\Delta V_T + 3\sqrt{(1+G_T)(\Gamma^2 + 2\Gamma P_T)} + \Gamma \sqrt{(\ln N+2) (1+G_T)}\\
\leq{}& \indicator\{\Delta >1\}\cdot2B\Delta V_T + 3\Gamma\sqrt{\frac{2(1+G_T)T}{\Delta}}+ 3\Gamma\sqrt{1+G_T} + \Gamma \sqrt{(\ln N+2) (1+G_T)},
\end{align*}
where the last inequality is due to the fact that $\sqrt{a+b}\leq \sqrt{a} + \sqrt{b}$ for any $a,b\geq0$. To obtain an upper bound of the above inequality, we consider the two following cases,

\textbf{Case 1}: ${3\Gamma\sqrt{2(1+G_T)T}}> {4BV_T} $, in such a case, we can set $\Delta = \lceil \frac{3\Gamma\sqrt{2(1+G_T)T}}{4BV_T}\rceil$. Then, we can bound the dynamic regret by
\begin{align*}
\E_{1:T}\left[\DReg\right]\leq{}& 3\left({9B\Gamma^2}\right)^{\frac{1}{3}}\cdot{V_T^{1/3}T^{1/3}(1+G_T)^{1/3}} + (3+ \sqrt{2+\ln N})\Gamma \sqrt{1+G_T}\\
={}& \O(V_T^{1/3}G_T^{1/3}T^{1/3} +\sqrt{1+G_T}).
\end{align*}

\textbf{Case 2:} ${3\Gamma\sqrt{2(1+G_T)T}}\leq {4BV_T} $, in such a case, we can set $\Delta = 1$. Then, we have 
\begin{align*}
\E_{1:T}\left[\DReg\right]\leq{}& 3\Gamma\sqrt{{2(1+G_T)T}}+ 3\Gamma\sqrt{1+G_T} + \Gamma \sqrt{(\ln N+2) (1+G_T)}\\
\leq{}&  \left(6\Gamma(1+G_T)T\right)^{1/3}\left(12\Gamma BV_T\right)^{1/3} + 3\Gamma\sqrt{1+G_T} + \Gamma \sqrt{(\ln N+2) (1+G_T)}\\
={}& \O(V_T^{1/3}G_T^{1/3}T^{1/3} +\sqrt{1+G_T}),
\end{align*}
where the second inequality is due to the inequality that ${3\Gamma\sqrt{2(1+G_T)T}}\leq {4BV_T} $. We complete the proof by combining the two cases.
\end{proof}

\subsection{Useful Lemmas in Analysis of Theorem~\ref{thm:overall-guarantees}}
\label{appendix:sec-proof-Atlas-ada-lemmas}
\setcounter{myLemma}{7}
\begin{myLemma}
By setting the learning rate $\epsilon = \sqrt{\frac{\ln N+2}{ 1+ \sum_{t=1}^T(\max_{i\in[N]}\{\vert \tilde{R}_t(\w_{t,i}) - \tilde{H}_t(\w_{t,i})\vert\})^2}}$, the meta-algorithm of \textsc{Atlas-Ada} (Algorithm~\ref{alg:atlas-ada-meta}) satisfies
\begin{align*}
\sum_{t=1}^T\sum_{j=1}^N p_{t,j} \Rh_t(\w_{t,j}) - \sum_{t=1}^T\Rh_t(\w_{t,i})\leq  2\Gamma\sqrt{(\ln N+2)\bigg(1+\sum_{t=1}^T \sup_{\w\in\W}\norm{\nabla \Rh_t(\w) - \nabla H_t(\w)}_2^2\bigg)}
\end{align*}
for any $i\in[N]$, where $\tilde{R}_{t}(\w_{t,i}) = \Rh_{t}(\w_{t,i}) - \Rh_t(\w_{\mathtt{ref}})$ and $\tilde{H}_{t}(\w_{t,i}) = H_t(\w_{t,i}) - H_t(\w_{\mathtt{ref}})$ are the reference loss functions built upon an arbitrary reference point $\w_{\mathtt{ref}}\in\W$.
\end{myLemma}

\begin{proof}[Proof of Lemma~\ref{lemma:Hedge-ada}]
The key challenge in the proof is that the meta-algorithm is updated with the loss function $\Rh_t(\w_{t,i})$ and the hint function $H_t(\w_{t,i})$, while we desire a regret that can scale with the gradient to make it compatible with the base-regret bound. To this end, we introduce the reference loss $\tilde{R}_{t}(\w_{t,i}) = \Rh_{t}(\w_{t,i}) - \Rh_t(\w_{\mathtt{ref}})$ and $\tilde{H}_{t}(\w_{t,i}) = H_t(\w_{t,i}) - H_t(\w_{\mathtt{ref}})$. It is easy to verify that the update procedure~\eqref{alg:atlas-meta-hedge-update} with the original loss function is identical to that with the reference loss:  
\begin{align*}
\resizebox{\hsize}{!}{$p_{t,i} = \frac{\exp\big({ -\varepsilon\big(\sum_{s=1}^{t-1}\Rh_s(\w_{s,i})+H_{t}(\w_{t,i})\big)}\big)}{\sum_{i=1}^K \exp\big({ -\varepsilon\big(\sum_{s=1}^{t-1}\Rh_s(\w_{s,i})+H_{t}(\w_{t,i})\big)}\big)} = \frac{\exp\big({ -\varepsilon\big(\sum_{s=1}^{t-1}\tilde{R}_s(\w_{s,i})+\tilde{H}_{t}(\w_{t,i})\big)}\big)}{\sum_{i=1}^K \exp\big({ -\varepsilon\big(\sum_{s=1}^{t-1}\tilde{R}_s(\w_{s,i})+\tilde{H}_{t}(\w_{t,i})\big)}\big)}.$}
\end{align*}

As a consequence, we can bound the meta-regret by
\begin{align}
&\sum_{t=1}^T \sum_{j=1}^N p_{t,j} \Rh_t(\w_{t,j}) - \sum_{t=1}^T \Rh_t(\w_{t,i})\notag\\
    ={}&\sum_{t=1}^T \sum_{j=1}^N p_{t,j} \tilde{R}_t(\w_{t,j}) - \sum_{t=1}^T \tilde{R}_t(\w_{t,i})\notag\\    
    \leq{}&\varepsilon\sum_{t=1}^T \big(\max_{i\in[N]}\{\vert\tilde{R}_t(\w_{t,i}) - \tilde{H}_t(\w_{t,i})\vert\}\big)^2 + \frac{\ln N+2}{\varepsilon}\notag\\
    \leq{}& 2\sqrt{\sum_{t=1}^T\big(\max_{i\in[N]}\{\vert \tilde{R}_t(\w_{t,i}) - \tilde{H}_t(\w_{t,i})\vert\}\big)^2(\ln N+2)} \label{eq:proof-meta-term-A},
\end{align}
where the first inequality is due to Lemma~\ref{lemma: Hedge-NIPS-15} and the second inequality is by the step size setting of $\epsilon = \sqrt{{(\ln N+2)}/{ (1 +\sum_{t=1}^T\max_{i\in[N]}\{\vert \tilde{R}_t(\w_{t,i}) - \tilde{H}_t(\w_{t,i})\vert\}})}$. 

Then, we show that the deviation between the loss function $\max_{i\in[N]}\{\vert\tilde{R}_t(\w_{t,i})-\tilde{H}_t(\w_{t,i})\vert\}$ can be converted to the deviation between the gradient. Let $G_t(\w_{t,i}) = \Rh_t(\w_{t,i})-{H}_t(\w_{t,i})$, for any $i\in[N]$,  we have
\begin{align}
    {}&\vert \tilde{R}_t(\w_{t,i}) - \tilde{H}_t(\w_{t,i})\vert =\vert G_t(\w_{t,i}) - G_t(\w_{\mathtt{ref}})\vert = \vert\inner{\nabla G_t(\bm{\xi}_{t,i})}{\w_{t,i}} \vert \notag\\
     \leq {}&\Gamma\norm{\nabla \Rh_t(\bm{\xi}_{t,i}) - \nabla H_t(\bm{\xi}_{t,i})}_2 \leq \Gamma\sup_{\w\in\W}\norm{\nabla \Rh_t(\w) - \nabla H_t(\w)}_2\label{eq:proof-meta-term-B},
\end{align}
where the first equality is due to the mean value theorem and  $\bm{\xi}_{t,i} = c_{t,i} \w_{t,i} + (1-c_{t,i})\w_{\mathtt{ref}}$ with a certain constant $c_{t,i}\in[0,1]$. Plugging~\eqref{eq:proof-meta-term-B} into~\eqref{eq:proof-meta-term-A}, we obtain that 
\begin{align*}
\sum_{t=1}^T \sum_{i=1}^N p_{t,i} \Rh_t(\w_{t-1,i}) - \sum_{t=1}^T \Rh_t(\w_{t-1,i}) \leq 2\Gamma\sqrt{(\ln N+2)\sum_{t=1}^T \sup_{\w\in\W}\norm{\nabla \Rh_t(\w) - \nabla H_t(\w)}_2^2},
\end{align*}
which completes the proof.
\end{proof}

Then, we provide the proof of the regret of the base-algorithm for \textsc{Atlas-Ada}.
\begin{myLemma}
Suppose the hint function $H_t: \W \mapsto \R$ is convex. The base-algorithm of \textsc{Atlas-Ada} (Algorithm~\ref{alg:atlas-ada-base}) with a step size $\eta_{t,i} > 0$ satisfies
\begin{align*}
\sum_{t=1}^T\inner{\nabla \Rh_t(\w_{t,i})}{\w_{t,i}-\u_t}\leq 2\eta_i\sum_{t=1}^T\sup_{\w\in\W}\norm{\nabla \Rh_t(\w)-\nabla H_t(\w)}_2^2 + \frac{\Gamma^2 + 2\Gamma P_T}{2\eta_i},
\end{align*}
for any comparator sequence $\{\u_t\}_{t=1}^T$ with $\u_{t}\in\W$. Moreover, $P_T = \sum_{t=2}^T \norm{\u_t-\u_{t-1}}_2$ measures the variation of the comparator sequence.
\end{myLemma}

\begin{proof}[Proof of Lemma~\ref{lemma:altas-base-regret-noexpectation}] We decompose the instantaneous loss of the base algorithm as,
\begin{align*}
    &\inner{\nabla \Rh_t(\w_{t,i})}{\w_{t,i}-\u_t} \\
    \leq{}&  \underbrace{\inner{\nabla \Rh_t(\w_{t,i})-\nabla H_t(\w_{t,i})}{\w_{t,i}-\wh_{t+1,i}} }_{\texttt{term~(a)}}  + \underbrace{\inner{\nabla H_t(\w_{t,i})}{\w_{t,i}-\wh_{t+1,i}}}_{\texttt{term~(b)}}  + \underbrace{\inner{\nabla \Rh_t(\w_{t,i})}{\wh_{t+1,i}-\u_t}}_{\texttt{term~(c)}}. 
\end{align*}
For term (a), by Cauchy-Schwarz inequality and AM-GM inequality, we have
\begin{align*}
    \texttt{term~(a)} \leq{}&\norm{\nabla \Rh_t(\w_{t,i})-\nabla H_t(\w_{t,i})}_2\norm{\w_{t,i}-\wh_{t+1,i}}_2\\
    \leq{}&2\eta_i\norm{\nabla \Rh_t(\w_{t,i})-\nabla H_t(\w_{t,i})}_2^2 + \frac{1}{2\eta_i}\norm{\w_{t,i}-\wh_{t+1,i}}_2^2.
\end{align*}
Then, we bound term (b) by Proposition 4.1 of~\citet{NIPS'20:Implict-V_T},
\begin{align*}
    \texttt{term~(b)}\leq\frac{1}{2\eta_i}\left(\norm{\wh_{t+1,i} - \wh_{t,i}}_2^2 - \norm{\wh_{t+1,i} - \w_{t,i}}_2^2 - \norm{\wh_{t,i} - \w_{t,i}}_2^2 \right).
\end{align*}
The term (c) is bounded by Lemma 7 of~\citet{NIPS'20:sword},
\begin{align*}
    \texttt{term~(c)}\leq\frac{1}{2\eta_i}\left(\norm{\u_t - \wh_{t,i}}_2^2 - \norm{\u_t - \wh_{t+1,i}}_2^2 - \norm{\wh_{t,i} - \wh_{t+1,i}}_2^2 \right).
\end{align*}
Combining the three upper bounds yields 
\begin{align*}
    &\sum_{t=1}^T\inner{\Rh_t(\w_{t,i})}{\w_{t,i}-\u_t}\\
    \leq{}& 2\eta_i\sum_{t=1}^T\norm{\nabla \Rh_t(\w_{t,i})-\nabla H_t(\w_{t,i})}_2^2 + \frac{1}{2\eta_i}\sum_{t=1}^T\left(\norm{\u_t - \wh_{t,i}}_2^2 - \norm{\u_t - \wh_{t+1,i}}_2^2\right)  + \frac{\Gamma^2}{2\eta_i}\\
    \leq {}&2\eta_i\sum_{t=1}^T\sup_{\w\in\W}\norm{\nabla \Rh_t(\w)-\nabla H_t(\w)}_2^2 + \frac{\Gamma^2 + 2\Gamma P_T}{2\eta_i},
\end{align*}
which completes the proof.
\end{proof}

%% file: sections/appendices/technique_lemma.tex
\section{Technical Lemmas}
\label{sec:appendix-lemma}

\begin{myLemma}
  \label{lemma:matrix-gap}
  Let $A\in \R^{K\times K}$ be an invertible matrix and $\Delta\in\R^{K\times K}$ be a squared matrix such that $\norm{A^{-1}\Delta}_2\leq 1/2$. For $B = A + \Delta$, we have the following two claims:
  \begin{compactitem}
  \item $B$ is an invertible matrix. 
  \item  $\norm{B^{-1}-A^{-1}}_2  \leq 2{\norm{A^{-1}}_2^2\cdot \norm{\Delta}_2}$.
  \end{compactitem}
\end{myLemma}
\begin{proof}
The proof follows the same arguments in~\citet[Page 381]{book'12:Matrix}. We first show that $B = A(I + A^{-1}\Delta)$ is an invertible matrix. Since $A$ is an invertible matrix by the assumption, it is sufficient to show that $I+A^{-1}\Delta$ is invertible. The statement is true since the spectral radius of $A^{-1}\Delta$ satisfies $\rho(A^{-1}\Delta)\leq \norm{A^{-1}\Delta}_2\leq 1/2$. The bounded spectral radius implies that $-1$ is not an eigenvalue of the matrix $A^{-1}\Delta$, and thus it is invertible. 

Then, we can upper bound the norm
\begin{align}
\label{eq:lemma-matrix-1}
    \norm{B^{-1}-A^{-1}}_2 = \norm{A^{-1}\Delta B^{-1}}_2 \leq \norm{A^{-1}\Delta}_2\norm{B^{-1}}_2,
\end{align}
where the term $\norm{B^{-1}}_2$ can be further bounded by
\begin{align*}
    \norm{B^{-1}}_2 \leq \norm{A^{-1}}_2 + \norm{B^{-1}-A^{-1}}_2\leq \norm{A^{-1}} +  \norm{A^{-1}\Delta}_2\norm{B^{-1}}_2.
\end{align*}
Rearranging the above inequality, we have
\begin{align}
\label{eq:lemma-matrix-2}
    \norm{B^{-1}}_2\leq \frac{\norm{A^{-1}}_2}{1 - \norm{A^{-1}\Delta}_2}.
\end{align}

Then, plugging~\eqref{eq:lemma-matrix-2} into~\eqref{eq:lemma-matrix-1} and rearranging the terms, we have 
\begin{align*}
    \norm{B^{-1}-A^{-1}}_2\leq \frac{\norm{A^{-1}\Delta}_2\norm{A^{-1}}_2}{1-\norm{A^{-1}\Delta}_2} \leq\frac{\norm{A^{-1}}^2_2\norm{\Delta}_2}{1-\norm{A^{-1}\Delta}_2}\leq 2{\norm{A^{-1}}^2_2\norm{\Delta}_2},
\end{align*}
which completes the proof.
\end{proof}

\begin{myLemma}[Theorem 19 of~\citet{NIPS'15:fast-rate-game}]
\label{lemma: Hedge-NIPS-15}
Let $\bm{\ell}_t\in\R^N$ be the loss vector taking $\ell_{t,i}\in\R$ as its $i$-th entry and $\m_t\in\R^{N}$ the hint vector taking $m_{t,i}$ as its $i$-th entry. The Hedge algorithm updating with $p_{t,i}\propto \exp\bigg(-\epsilon\big(\sum_{s=1}^{t-1} \ell_{s,i} + m_{t,i}\big)\bigg)$ satisfies
\begin{align*}
\sum_{t=1}^T \sum_{j=1}^N p_{t,j} \ell_{t,j} - \sum_{t=1}^T \ell_{t,i} \leq \frac{\ln N+2}{\epsilon} + \epsilon\sum_{t=1}^T\norm{\bm{\ell}_t-\m_t}^2_{\infty}
\end{align*}
for any $i\in[N]$, where $\epsilon>0$ is the step size.
\end{myLemma}

%% file: arxiv_v2.bbl
\begin{thebibliography}{83}
\providecommand{\natexlab}[1]{#1}
\providecommand{\url}[1]{\texttt{#1}}
\expandafter\ifx\csname urlstyle\endcsname\relax
  \providecommand{\doi}[1]{doi: #1}\else
  \providecommand{\doi}{doi: \begingroup \urlstyle{rm}\Url}\fi

\bibitem[Quionero-Candela et~al.(2009)Quionero-Candela, Sugiyama, Schwaighofer,
  and Lawrence]{edit:Quinonero-Candela+etal:2009}
Joaquin Quionero-Candela, Masashi Sugiyama, Anton Schwaighofer, and Neil~D.
  Lawrence.
\newblock \emph{Dataset Shift in Machine Learning}.
\newblock The MIT Press, 2009.

\bibitem[Sugiyama and Kawanabe(2012)]{book/mit/sugiyama2012machine}
Masashi Sugiyama and Motoaki Kawanabe.
\newblock \emph{Machine {L}earning in {N}on-stationary {E}nvironments:
  Introduction to {C}ovariate {S}hift {A}daptation}.
\newblock The MIT Press, 2012.

\bibitem[Dietterich(2017)]{TGD:robust-AI}
Thomas~G. Dietterich.
\newblock Steps toward robust artificial intelligence.
\newblock \emph{{AI} Magazine}, 38\penalty0 (3):\penalty0 3--24, 2017.

\bibitem[Bengio et~al.(2021)Bengio, Lecun, and Hinton]{CACM'21:DL-AI}
Yoshua Bengio, Yann Lecun, and Geoffrey Hinton.
\newblock Deep learning for {AI}.
\newblock \emph{Communication of {ACM}}, 64\penalty0 (7):\penalty0 58–65,
  2021.

\bibitem[Zhou(2022)]{nsr'22:Open-Survey}
Zhi-Hua Zhou.
\newblock {Open-environment machine learning}.
\newblock \emph{National Science Review}, 9\penalty0 (8):\penalty0 nwac123,
  2022.

\bibitem[G{\'{o}}mez{-}Villa et~al.(2017)G{\'{o}}mez{-}Villa, Salazar, and
  Vargas{-}Bonilla]{ECOI'17:wild}
Alexander G{\'{o}}mez{-}Villa, Augusto Salazar, and Jes{\'{u}}s~Francisco
  Vargas{-}Bonilla.
\newblock Towards automatic wild animal monitoring: Identification of animal
  species in camera-trap images using very deep convolutional neural networks.
\newblock \emph{Ecological Informatics}, 41:\penalty0 24--32, 2017.

\bibitem[Norouzzadeh et~al.(2018)Norouzzadeh, Nguyen, Kosmala, Swanson, Palmer,
  Packer, and Clune]{PNAS'18:Species}
Mohammad~Sadegh Norouzzadeh, Anh Nguyen, Margaret Kosmala, Alexandra Swanson,
  Meredith~S. Palmer, Craig Packer, and Jeff Clune.
\newblock Automatically identifying, counting, and describing wild animals in
  camera-trap images with deep learning.
\newblock \emph{Proceedings of the National Academy of Sciences}, 115\penalty0
  (25):\penalty0 E5716--E5725, 2018.

\bibitem[Saerens et~al.(2002)Saerens, Latinne, and
  Decaestecker]{DBLP:journals/neco/SaerensLD02}
Marco Saerens, Patrice Latinne, and Christine Decaestecker.
\newblock Adjusting the outputs of a classifier to new a priori probabilities:
  {A} simple procedure.
\newblock \emph{Neural Computation}, 14\penalty0 (1):\penalty0 21--41, 2002.

\bibitem[Zhang et~al.(2013)Zhang, Sch{\"{o}}lkopf, Muandet, and
  Wang]{conf/icml/ZhangSMW13}
Kun Zhang, Bernhard Sch{\"{o}}lkopf, Krikamol Muandet, and Zhikun Wang.
\newblock Domain adaptation under target and conditional shift.
\newblock In \emph{Proceedings of the 30th International Conference on Machine
  Learning {(ICML)}}, pages 819--827, 2013.

\bibitem[du~Plessis and Sugiyama(2014)]{DBLP:journals/nn/PlessisS14}
Marthinus~Christoffel du~Plessis and Masashi Sugiyama.
\newblock Semi-supervised learning of class balance under class-prior change by
  distribution matching.
\newblock \emph{Neural Networks}, 50:\penalty0 110--119, 2014.

\bibitem[Nguyen et~al.(2015)Nguyen, du~Plessis, and
  Sugiyama]{DBLP:conf/acml/NguyenPS15}
Tuan~Duong Nguyen, Marthinus~Christoffel du~Plessis, and Masashi Sugiyama.
\newblock Continuous target shift adaptation in supervised learning.
\newblock In \emph{Proceedings of the 7th Asian Conference on Machine Learning
  ({ACML})}, pages 285--300, 2015.

\bibitem[Lipton et~al.(2018)Lipton, Wang, and Smola]{conf/icml/LiptonWS18}
Zachary~C. Lipton, Yu{-}Xiang Wang, and Alexander~J. Smola.
\newblock Detecting and correcting for label shift with black box predictors.
\newblock In \emph{Proceedings of the 35th International Conference on Machine
  Learning (ICML)}, pages 3128--3136, 2018.

\bibitem[Azizzadenesheli et~al.(2019)Azizzadenesheli, Liu, Yang, and
  Anandkumar]{conf/iclr/Azizzadenesheli19}
Kamyar Azizzadenesheli, Anqi Liu, Fanny Yang, and Animashree Anandkumar.
\newblock Regularized learning for domain adaptation under label shifts.
\newblock In \emph{Proceedings of the 7th International Conference on Learning
  Representations (ICLR)}, 2019.

\bibitem[Garg et~al.(2020)Garg, Wu, Balakrishnan, and
  Lipton]{DBLP:conf/nips/GargWBL20}
Saurabh Garg, Yifan Wu, Sivaraman Balakrishnan, and Zachary~C. Lipton.
\newblock A unified view of label shift estimation.
\newblock In \emph{Advances in Neural Information Processing Systems 33
  (NeurIPS)}, pages 3290--3300, 2020.

\bibitem[Hazan(2016)]{book'16:Hazan-OCO}
Elad Hazan.
\newblock Introduction to {O}nline {C}onvex {O}ptimization.
\newblock \emph{Foundations and Trends in Optimization}, 2\penalty0
  (3-4):\penalty0 157--325, 2016.

\bibitem[Wu et~al.(2021{\natexlab{a}})Wu, Guo, Su, and
  Weinberger]{NIPS'21:Online-LS}
Ruihan Wu, Chuan Guo, Yi~Su, and Kilian~Q. Weinberger.
\newblock Online adaptation to label distribution shift.
\newblock In \emph{Advances in Neural Information Processing Systems 34
  (NeurIPS)}, pages 11340--11351, 2021{\natexlab{a}}.

\bibitem[Besbes et~al.(2015)Besbes, Gur, and Zeevi]{OR'15:dynamic-function-VT}
Omar Besbes, Yonatan Gur, and Assaf~J. Zeevi.
\newblock Non-stationary stochastic optimization.
\newblock \emph{Operations Research}, 63\penalty0 (5):\penalty0 1227--1244,
  2015.

\bibitem[Zhang et~al.(2018{\natexlab{a}})Zhang, Lu, and
  Zhou]{NIPS'18:Zhang-Ader}
Lijun Zhang, Shiyin Lu, and Zhi-Hua Zhou.
\newblock Adaptive online learning in dynamic environments.
\newblock In \emph{Advances in Neural Information Processing Systems 31
  (NeurIPS)}, pages 1330--1340, 2018{\natexlab{a}}.

\bibitem[Zhao et~al.(2020{\natexlab{a}})Zhao, Zhang, Zhang, and
  Zhou]{NIPS'20:sword}
Peng Zhao, Yu-Jie Zhang, Lijun Zhang, and Zhi-Hua Zhou.
\newblock Dynamic regret of convex and smooth functions.
\newblock In \emph{Advances in Neural Information Processing Systems 33
  (NeurIPS)}, pages 12510--12520, 2020{\natexlab{a}}.

\bibitem[Zinkevich(2003)]{ICML'03:zinkvich}
Martin Zinkevich.
\newblock Online convex programming and generalized infinitesimal gradient
  ascent.
\newblock In \emph{Proceedings of the 20th International Conference on Machine
  Learning (ICML)}, pages 928--936, 2003.

\bibitem[Cesa-Bianchi and Lugosi(2006)]{book/Cambridge/cesa2006prediction}
Nicol\`{o} Cesa-Bianchi and G{\'a}bor Lugosi.
\newblock \emph{Prediction, {L}earning, and {G}ames}.
\newblock Cambridge {U}niversity {P}ress, 2006.

\bibitem[Bachem et~al.(2017)Bachem, Lucic, and Krause]{arXiv'17:coreset}
Olivier Bachem, Mario Lucic, and Andreas Krause.
\newblock Practical coreset constructions for machine learning.
\newblock arXiv:1703.06476, 2017.

\bibitem[Mirzasoleiman et~al.(2020)Mirzasoleiman, Bilmes, and
  Leskovec]{ICML'20:coreset}
Baharan Mirzasoleiman, Jeff~A. Bilmes, and Jure Leskovec.
\newblock Coresets for data-efficient training of machine learning models.
\newblock In \emph{Proceedings of the 37th International Conference on Machine
  Learning (ICML)}, pages 6950--6960, 2020.

\bibitem[Wu et~al.(2021{\natexlab{b}})Wu, Xu, Liu, and Zhou]{TKDE'21:RKME}
Xi-Zhu Wu, Wenkai Xu, Song Liu, and Zhi-Hua Zhou.
\newblock Model reuse with reduced kernel mean embedding specification.
\newblock \emph{IEEE Transactions on Knowledge and Data Engineering},
  35:\penalty0 699--710, 2021{\natexlab{b}}.

\bibitem[Zhang et~al.(2021{\natexlab{a}})Zhang, Yan, Zhao, and
  Zhou]{AAAI'21:UnseenJob}
Yu-Jie Zhang, Yu-Hu Yan, Peng Zhao, and Zhi-Hua Zhou.
\newblock Towards enabling learnware to handle unseen jobs.
\newblock In \emph{Proceedings of the 35th AAAI Conference on Artificial
  Intelligence (AAAI)}, pages 10964--10972, 2021{\natexlab{a}}.

\bibitem[Muandet et~al.(2017)Muandet, Fukumizu, Sriperumbudur, and
  Sch{\"{o}}lkopf]{FTML'17:KME}
Krikamol Muandet, Kenji Fukumizu, Bharath~K. Sriperumbudur, and Bernhard
  Sch{\"{o}}lkopf.
\newblock Kernel mean embedding of distributions: {A} review and beyond.
\newblock \emph{Foundations and Trends in Machine Learning}, 10\penalty0
  (1-2):\penalty0 1--141, 2017.

\bibitem[Jadbabaie et~al.(2015)Jadbabaie, Rakhlin, Shahrampour, and
  Sridharan]{AISTATS'15:dynamic-optimistic}
Ali Jadbabaie, Alexander Rakhlin, Shahin Shahrampour, and Karthik Sridharan.
\newblock Online optimization : Competing with dynamic comparators.
\newblock In \emph{Proceedings of the 18th International Conference on
  Artificial Intelligence and Statistics (AISTATS)}, pages 398--406, 2015.

\bibitem[Zhang et~al.(2020{\natexlab{a}})Zhang, Zhao, and Zhou]{UAI'20:Simple}
Yu{-}Jie Zhang, Peng Zhao, and Zhi{-}Hua Zhou.
\newblock A simple online algorithm for competing with dynamic comparators.
\newblock In \emph{Proceedings of the 36th Conference on Uncertainty in
  Artificial Intelligence (UAI)}, pages 390--399, 2020{\natexlab{a}}.

\bibitem[Abernethy et~al.(2008{\natexlab{a}})Abernethy, Bartlett, Rakhlin, and
  Tewari]{COLT'08:lower-bound}
Jacob Abernethy, Peter~L Bartlett, Alexander Rakhlin, and Ambuj Tewari.
\newblock Optimal strategies and minimax lower bounds for online convex games.
\newblock In \emph{Proceedings of the 21st Annual Conference on Learning Theory
  (COLT)}, pages 415--423, 2008{\natexlab{a}}.

\bibitem[Cesa{-}Bianchi et~al.(1997)Cesa{-}Bianchi, Freund, Haussler, Helmbold,
  Schapire, and Warmuth]{JACM'97:doubling-trick}
Nicol{\`{o}} Cesa{-}Bianchi, Yoav Freund, David Haussler, David~P. Helmbold,
  Robert~E. Schapire, and Manfred~K. Warmuth.
\newblock How to use expert advice.
\newblock \emph{Journal of the ACM}, 44\penalty0 (3):\penalty0 427--485, 1997.

\bibitem[Auer et~al.(2002)Auer, Cesa{-}Bianchi, and
  Gentile]{JCSS'02:Auer-self-confident}
Peter Auer, Nicol{\`{o}} Cesa{-}Bianchi, and Claudio Gentile.
\newblock Adaptive and self-confident on-line learning algorithms.
\newblock \emph{Journal of Computer and System Sciences}, 64\penalty0
  (1):\penalty0 48--75, 2002.

\bibitem[Freund and Schapire(1997)]{JCSS'97:boosting}
Yoav Freund and Robert~E. Schapire.
\newblock A decision-theoretic generalization of on-line learning and an
  application to boosting.
\newblock \emph{Journal of Computer and System Sciences}, 55\penalty0
  (1):\penalty0 119--139, 1997.

\bibitem[Roughgarden(2020)]{book'20:beyond-worst-case}
Tim Roughgarden, editor.
\newblock \emph{Beyond the {W}orst-{C}ase {A}nalysis of {A}lgorithms}.
\newblock Cambridge University Press, 2020.

\bibitem[Zhao et~al.(2020{\natexlab{b}})Zhao, Cai, and Zhou]{MLJ'20:Condor}
Peng Zhao, Le-Wen Cai, and Zhi-Hua Zhou.
\newblock Handling concept drift via model reuse.
\newblock \emph{Machine Learning}, 109\penalty0 (3):\penalty0 533--568,
  2020{\natexlab{b}}.

\bibitem[Chiang et~al.(2012)Chiang, Yang, Lee, Mahdavi, Lu, Jin, and
  Zhu]{COLT'12:variation-Yang}
Chao-Kai Chiang, Tianbao Yang, Chia-Jung Lee, Mehrdad Mahdavi, Chi-Jen Lu, Rong
  Jin, and Shenghuo Zhu.
\newblock Online optimization with gradual variations.
\newblock In \emph{Proceedings of the 25th Conference On Learning Theory
  (COLT)}, pages 6.1--6.20, 2012.

\bibitem[Rakhlin and Sridharan(2013)]{conf/colt/RakhlinS13}
Alexander Rakhlin and Karthik Sridharan.
\newblock Online learning with predictable sequences.
\newblock In \emph{Proceedings of the 26th Conference On Learning Theory
  (COLT)}, pages 993--1019, 2013.

\bibitem[Kulis and Bartlett(2010)]{ICML'10:Implict-OL}
Brian Kulis and Peter~L. Bartlett.
\newblock Implicit online learning.
\newblock In \emph{Proceedings of the 27th International Conference on Machine
  Learning (ICML)}, pages 575--582, 2010.

\bibitem[Campolongo and Orabona(2020)]{NIPS'20:Implict-V_T}
Nicol{\`{o}} Campolongo and Francesco Orabona.
\newblock Temporal variability in implicit online learning.
\newblock In \emph{Advances in Neural Information Processing Systems 33
  (NeurIPS)}, pages 12377--12387, 2020.

\bibitem[Gjoreski et~al.(2018)Gjoreski, Ciliberto, Wang, Morales, Mekki,
  Valentin, and Roggen]{DBLP:journals/access/GjoreskiCWMMVR18}
Hristijan Gjoreski, Mathias Ciliberto, Lin Wang, Francisco
  Javier~Ord{\'{o}}{\~{n}}ez Morales, Sami Mekki, Stefan Valentin, and Daniel
  Roggen.
\newblock The university of {Sussex-Huawei} locomotion and transportation
  dataset for multimodal analytics with mobile devices.
\newblock \emph{{IEEE} Access}, 6:\penalty0 42592--42604, 2018.

\bibitem[Sanh et~al.(2019)Sanh, Debut, Chaumond, and
  Wolf]{journals/corr/abs-1910-01108}
Victor Sanh, Lysandre Debut, Julien Chaumond, and Thomas Wolf.
\newblock {DistilBERT}, a distilled version of {BERT:} smaller, faster, cheaper
  and lighter.
\newblock arXiv:1910.01108, 2019.

\bibitem[Helber et~al.(2018)Helber, Bischke, Dengel, and
  Borth]{DBLP:journals/staeors/HelberBDB19}
Patrick Helber, Benjamin Bischke, Andreas Dengel, and Damian Borth.
\newblock Introducing {EuroSAT}: A novel dataset and deep learning benchmark
  for land use and land cover classification.
\newblock In \emph{Proceedings of the IEEE International Geoscience and Remote
  Sensing Symposium ({IGARSS})}, pages 204--207, 2018.

\bibitem[He et~al.(2016)He, Zhang, Ren, and Sun]{DBLP:conf/cvpr/HeZRS16}
Kaiming He, Xiangyu Zhang, Shaoqing Ren, and Jian Sun.
\newblock Deep residual learning for image recognition.
\newblock In \emph{Proceedings of the {IEEE} Conference on Computer Vision and
  Pattern Recognition (CVPR)}, pages 770--778, 2016.

\bibitem[LeCun et~al.(1998)LeCun, Bottou, Bengio, and
  Haffner]{DBLP:journals/pieee/LeCunBBH98}
Yann LeCun, L{\'{e}}on Bottou, Yoshua Bengio, and Patrick Haffner.
\newblock Gradient-based learning applied to document recognition.
\newblock \emph{Proceedings of the {IEEE}}, 86\penalty0 (11):\penalty0
  2278--2324, 1998.

\bibitem[Xiao et~al.(2017)Xiao, Rasul, and
  Vollgraf]{DBLP:journals/corr/abs-1708-07747}
Han Xiao, Kashif Rasul, and Roland Vollgraf.
\newblock {Fashion-MNIST}: A novel image dataset for benchmarking machine
  learning algorithms.
\newblock arXiv:1708.07747, 2017.

\bibitem[Krizhevsky(2009)]{Krizhevsky09learningmultiple}
Alex Krizhevsky.
\newblock Learning multiple layers of features from tiny images.
\newblock Technical report, 2009.

\bibitem[Darlow et~al.(2018)Darlow, Crowley, Antoniou, and
  Storkey]{DBLP:journals/corr/abs-1810-03505}
Luke~Nicholas Darlow, Elliot~J. Crowley, Antreas Antoniou, and Amos~J. Storkey.
\newblock {CINIC-10} is not {ImageNet} or {CIFAR-10}.
\newblock arXiv:1810.03505, 2018.

\bibitem[Deng et~al.(2009)Deng, Dong, Socher, Li, Li, and
  Fei-Fei]{deng2009imagenet}
Jia Deng, Wei Dong, Richard Socher, Li-Jia Li, Kai Li, and Li~Fei-Fei.
\newblock {ImageNet}: A large-scale hierarchical image database.
\newblock In \emph{Proceedings of the IEEE Conference on Computer Vision and
  Pattern Recognition (CVPR)}, pages 248--255, 2009.

\bibitem[Shalev-Shwartz(2012)]{book'12:Shai-OCO}
Shai Shalev-Shwartz.
\newblock Online {L}earning and {O}nline {C}onvex {O}ptimization.
\newblock \emph{Foundations and Trends in Machine Learning}, 4\penalty0
  (2):\penalty0 107--194, 2012.

\bibitem[Abernethy et~al.(2008{\natexlab{b}})Abernethy, Bartlett, Rakhlin, and
  Tewari]{DBLP:conf/colt/AbernethyBRT08}
Jacob~D. Abernethy, Peter~L. Bartlett, Alexander Rakhlin, and Ambuj Tewari.
\newblock Optimal stragies and minimax lower bounds for online convex games.
\newblock In \emph{Proceedings of the 21st Conference On Learning Theory
  (COLT)}, pages 415--424, 2008{\natexlab{b}}.

\bibitem[Hazan et~al.(2007)Hazan, Agarwal, and Kale]{journals/ml/HazanAK07}
Elad Hazan, Amit Agarwal, and Satyen Kale.
\newblock Logarithmic regret algorithms for online convex optimization.
\newblock \emph{Machine Learning}, 69\penalty0 (2-3):\penalty0 169--192, 2007.

\bibitem[Hazan and Kale(2008)]{COLT'08:Hazan-variation}
Elad Hazan and Satyen Kale.
\newblock Extracting certainty from uncertainty: Regret bounded by variation in
  costs.
\newblock In \emph{Proceedings of the 21st Annual Conference on Learning Theory
  (COLT)}, pages 57--68, 2008.

\bibitem[Blanchard et~al.(2010)Blanchard, Lee, and
  Scott]{journals/jmlr/BlanchardLS10}
Gilles Blanchard, Gyemin Lee, and Clayton Scott.
\newblock Semi-supervised novelty detection.
\newblock \emph{Journal of Machine Learning Research}, 11:\penalty0 2973--3009,
  2010.

\bibitem[du~Plessis et~al.(2017)du~Plessis, Niu, and
  Sugiyama]{MLJ'17:class-prior}
Marthinus~Christoffel du~Plessis, Gang Niu, and Masashi Sugiyama.
\newblock Class-prior estimation for learning from positive and unlabeled data.
\newblock \emph{Machine Learning}, 106\penalty0 (4):\penalty0 463--492, 2017.

\bibitem[Ramaswamy et~al.(2016)Ramaswamy, Scott, and
  Tewari]{conf/icml/RamaswamyST16}
Harish~G. Ramaswamy, Clayton Scott, and Ambuj Tewari.
\newblock Mixture proportion estimation via kernel embeddings of distributions.
\newblock In \emph{Proceedings of the 33rd International Conference on Machine
  Learning ({ICML})}, pages 2052--2060, 2016.

\bibitem[Scott(2015)]{conf/aistats/Scott15}
Clayton Scott.
\newblock A rate of convergence for mixture proportion estimation, with
  application to learning from noisy labels.
\newblock In \emph{Proceedings of the 18th International Conference on
  Artificial Intelligence and Statistics (AISTATS)}, pages 838--846, 2015.

\bibitem[du~Plessis et~al.(2014)du~Plessis, Niu, and
  Sugiyama]{conf/nips/PlessisNS14}
Marthinus~Christoffel du~Plessis, Gang Niu, and Masashi Sugiyama.
\newblock Analysis of learning from positive and unlabeled data.
\newblock In \emph{Advances in Neural Information Processing Systems 27
  ({NeurIPS})}, pages 703--711, 2014.

\bibitem[du~Plessis et~al.(2015)du~Plessis, Niu, and
  Sugiyama]{conf/icml/PlessisNS15}
Marthinus~Christoffel du~Plessis, Gang Niu, and Masashi Sugiyama.
\newblock Convex formulation for learning from positive and unlabeled data.
\newblock In \emph{Proceedings of the 32nd International Conference on Machine
  Learning (ICML)}, pages 1386--1394, 2015.

\bibitem[Zhang et~al.(2020{\natexlab{b}})Zhang, Zhao, Ma, and
  Zhou]{NeurIPS'20:EULAC}
Yu-Jie Zhang, Peng Zhao, Lanjihong Ma, and Zhi-Hua Zhou.
\newblock An unbiased risk estimator for learning with augmented classes.
\newblock In \emph{Advances in Neural Information Processing Systems 33
  ({NeurIPS})}, pages 10247--10258, 2020{\natexlab{b}}.

\bibitem[Yang et~al.(2016)Yang, Zhang, Jin, and Yi]{ICML'16:Yang-smooth}
Tianbao Yang, Lijun Zhang, Rong Jin, and Jinfeng Yi.
\newblock Tracking slowly moving clairvoyant: Optimal dynamic regret of online
  learning with true and noisy gradient.
\newblock In \emph{Proceedings of the 33rd International Conference on Machine
  Learning (ICML)}, pages 449--457, 2016.

\bibitem[Zhang et~al.(2017)Zhang, Yang, Yi, Jin, and Zhou]{NIPS:2017:Zhang}
Lijun Zhang, Tianbao Yang, Jinfeng Yi, Rong Jin, and Zhi-Hua Zhou.
\newblock Improved dynamic regret for non-degenerate functions.
\newblock In \emph{Advance in Neural Information Processing Systems 30 (NIPS)},
  pages 732--741, 2017.

\bibitem[Zhang et~al.(2018{\natexlab{b}})Zhang, Yang, Jin, and
  Zhou]{ICML'18:zhang-dynamic-adaptive}
Lijun Zhang, Tianbao Yang, Rong Jin, and Zhi-Hua Zhou.
\newblock Dynamic regret of strongly adaptive methods.
\newblock In \emph{Proceedings of the 35th International Conference on Machine
  Learning (ICML)}, pages 5877--5886, 2018{\natexlab{b}}.

\bibitem[Baby and Wang(2019)]{conf/nips/BabyW19}
Dheeraj Baby and Yu{-}Xiang Wang.
\newblock Online forecasting of total-variation-bounded sequences.
\newblock In \emph{Advances in Neural Information Processing Systems 32
  (NeurIPS)}, pages 11069--11079, 2019.

\bibitem[Chen et~al.(2019)Chen, Wang, and Wang]{OR'19:V_T-pq}
Xi~Chen, Yining Wang, and Yu-Xiang Wang.
\newblock Non-stationary stochastic optimization under ${L}_{p, q}$-variation
  measures.
\newblock \emph{Operations Research}, 67\penalty0 (6):\penalty0 1752--1765,
  2019.

\bibitem[Zhao and Zhang(2021{\natexlab{a}})]{L4DC'21:SC-Smooth}
Peng Zhao and Lijun Zhang.
\newblock Improved analysis for dynamic regret of strongly convex and smooth
  functions.
\newblock In \emph{Proceedings of the 3rd Conference on Learning for Dynamics
  and Control (L4DC)}, pages 48--59, 2021{\natexlab{a}}.

\bibitem[Mokhtari et~al.(2016)Mokhtari, Shahrampour, Jadbabaie, and
  Ribeiro]{CDC'16:dynamic-sc}
Aryan Mokhtari, Shahin Shahrampour, Ali Jadbabaie, and Alejandro Ribeiro.
\newblock Online optimization in dynamic environments: Improved regret rates
  for strongly convex problems.
\newblock In \emph{Proceedings of the 55th IEEE Conference on Decision and
  Control (CDC)}, pages 7195--7201, 2016.

\bibitem[Zhang et~al.(2020{\natexlab{c}})Zhang, Lu, and Yang]{AISTATS'20:Zhang}
Lijun Zhang, Shiyin Lu, and Tianbao Yang.
\newblock Minimizing dynamic regret and adaptive regret simultaneously.
\newblock In \emph{Proceedings of the 23rd International Conference on
  Artificial Intelligence and Statistics (AISTATS)}, pages 309--319,
  2020{\natexlab{c}}.

\bibitem[Cutkosky(2020)]{ICML'20:Ashok}
Ashok Cutkosky.
\newblock Parameter-free, dynamic, and strongly-adaptive online learning.
\newblock In \emph{Proceedings of the 37th International Conference on Machine
  Learning (ICML)}, pages 2250--2259, 2020.

\bibitem[Zhao et~al.(2021{\natexlab{a}})Zhao, Zhang, Zhang, and
  Zhou]{arXiv'21:Sword++}
Peng Zhao, Yu{-}Jie Zhang, Lijun Zhang, and Zhi{-}Hua Zhou.
\newblock Adaptivity and non-stationarity: Problem-dependent dynamic regret for
  online convex optimization.
\newblock arXiv:2112.14368, 2021{\natexlab{a}}.

\bibitem[Zhao et~al.(2021{\natexlab{b}})Zhao, Wang, Zhang, and
  Zhou]{JMLR'21:BCO}
Peng Zhao, Guanghui Wang, Lijun Zhang, and Zhi-Hua Zhou.
\newblock Bandit convex optimization in non-stationary environments.
\newblock \emph{Journal of Machine Learning Research}, 22\penalty0
  (125):\penalty0 1 -- 45, 2021{\natexlab{b}}.

\bibitem[Baby and Wang(2021)]{conf/colt/BabyW21}
Dheeraj Baby and Yu{-}Xiang Wang.
\newblock Optimal dynamic regret in exp-concave online learning.
\newblock In \emph{Proceedings of the 34th Annual Conference on Learning Theory
  (COLT)}, pages 359--409, 2021.

\bibitem[Zhao et~al.(2022{\natexlab{a}})Zhao, Wang, and
  Zhou]{AISTATS'22:memory}
Peng Zhao, Yu-Xiang Wang, and Zhi-Hua Zhou.
\newblock Non-stationary online learning with memory and non-stochastic
  control.
\newblock In \emph{Proceedings of the 25th International Conference on
  Artificial Intelligence and Statistics (AISTATS)}, pages 2101--2133,
  2022{\natexlab{a}}.

\bibitem[Zhang et~al.(2021{\natexlab{b}})Zhang, Jiang, Lu, and
  Yang]{NeurIPS:2021:Zhang:A}
Lijun Zhang, Wei Jiang, Shiyin Lu, and Tianbao Yang.
\newblock Revisiting smoothed online learning.
\newblock In \emph{Advances in Neural Information Processing Systems 34
  (NeurIPS)}, pages 13599--13612, 2021{\natexlab{b}}.

\bibitem[Zhao et~al.(2022{\natexlab{b}})Zhao, Li, and Zhou]{ICML'22:mdp}
Peng Zhao, Long-Fei Li, and Zhi-Hua Zhou.
\newblock Dynamic regret of online markov decision processes.
\newblock In \emph{Proceedings of the 39th International Conference on Machine
  Learning (ICML)}, pages 26865--26894, 2022{\natexlab{b}}.

\bibitem[Zhang et~al.(2022)Zhang, Zhao, Luo, and Zhou]{ICML'22:TVgame}
Mengxiao Zhang, Peng Zhao, Haipeng Luo, and Zhi-Hua Zhou.
\newblock No-regret learning in time-varying zero-sum games.
\newblock In \emph{Proceedings of the 39th International Conference on Machine
  Learning (ICML))}, pages 26772--26808, 2022.

\bibitem[Jacobsen and Cutkosky(2022)]{colt'22:parameter-free-mirror-descent}
Andrew Jacobsen and Ashok Cutkosky.
\newblock Parameter-free mirror descent.
\newblock In \emph{Proceedings of the 35th Annual Conference on Learning Theory
  (COLT)}, pages 4160--4211, 2022.

\bibitem[Tropp(2012)]{FCM'12:Matrix-Concentration}
Joel~A. Tropp.
\newblock User-friendly tail bounds for sums of random matrices.
\newblock \emph{Foundations of Computational Mathematics}, 12\penalty0
  (4):\penalty0 389--434, 2012.

\bibitem[Hsu et~al.(2012)Hsu, Kakade, and Zhang]{JCSS'12:Hsu}
Daniel~J. Hsu, Sham~M. Kakade, and Tong Zhang.
\newblock A spectral algorithm for learning hidden markov models.
\newblock \emph{Journal of Computer and System Sciences}, 78\penalty0
  (5):\penalty0 1460--1480, 2012.

\bibitem[Zhang et~al.(2020{\natexlab{d}})Zhang, Zhao, and
  Jin]{AAAI'20:DynamicOMD}
Teng Zhang, Peng Zhao, and Hai Jin.
\newblock Optimal margin distribution learning in dynamic environments.
\newblock In \emph{Proceedings of the 34th {AAAI} Conference on Artificial
  Intelligence (AAAI)}, pages 6821--6828, 2020{\natexlab{d}}.

\bibitem[Zhou et~al.(2022)Zhou, Zhao, Zhang, Wang, Chang, Wang, and
  Zhu]{AISTATS'22:OnlineContinualAdaptation}
Shiji Zhou, Han Zhao, Shanghang Zhang, Lianzhe Wang, Heng Chang, Zhi Wang, and
  Wenwu Zhu.
\newblock Online continual adaptation with active self-training.
\newblock In \emph{Proceedings of the 25th International Conference on
  Artificial Intelligence and Statistics (AISTATS)}, pages 8852--8883, 2022.

\bibitem[Zhao and Zhang(2021{\natexlab{b}})]{L4DC'21:sc_smooth}
Peng Zhao and Lijun Zhang.
\newblock Improved analysis for dynamic regret of strongly convex and smooth
  functions.
\newblock In \emph{Proceedings of the 3rd Conference on Learning for Dynamics
  and Control (L4DC)}, pages 48--59, 2021{\natexlab{b}}.

\bibitem[Barbakh and Fyfe(2008)]{DBLP:journals/ijns/BarbakhF08}
Wesam Barbakh and Colin Fyfe.
\newblock Online clustering algorithms.
\newblock \emph{International Journal of Neural Systems}, 18\penalty0
  (3):\penalty0 185--194, 2008.

\bibitem[Horn and Johnson(2012)]{book'12:Matrix}
Roger~A. Horn and Charles~R. Johnson.
\newblock \emph{Matrix {A}nalysis}.
\newblock Cambridge University Press, second edition, 2012.

\bibitem[Syrgkanis et~al.(2015)Syrgkanis, Agarwal, Luo, and
  Schapire]{NIPS'15:fast-rate-game}
Vasilis Syrgkanis, Alekh Agarwal, Haipeng Luo, and Robert~E. Schapire.
\newblock Fast convergence of regularized learning in games.
\newblock In \emph{Advances in Neural Information Processing Systems 28
  (NIPS)}, pages 2989--2997, 2015.

\end{thebibliography}
